%% file: scoping_review_ml_power_protection.tex
\documentclass[a4paper,fleqn]{cas-sc}

\def\tsc#1{\csdef{#1}{\textsc{\lowercase{#1}}\xspace}}
\tsc{WGM}
\tsc{QE}
\usepackage{amsmath}          % Advanced math typesetting
\usepackage{array}            % Extended tabular environment
\usepackage{booktabs}         % For professional-looking tables
\usepackage{longtable}        % For tables that span multiple pages
\usepackage{tabularx}         % For tables with flexible column widths
\usepackage{siunitx}          % Consistent typesetting of SI units and numbers
\usepackage{enumitem}

\usepackage{tikz}
\usepackage{forest}
\usetikzlibrary{fit,arrows.meta,positioning,shadows.blur}

\usepackage{graphicx}
\graphicspath{{figs/}}

\usepackage[numbers]{natbib} % Numeric citation style (e.g., [1], [2])
\usepackage{acronym} % Define and manage acronyms without list generation

\usepackage[colorinlistoftodos,prependcaption,textsize=tiny]{todonotes}        % For adding TODO notes in the margin

\usepackage{subfiles}         % Load subfiles, useful for modular documents (load last)

\usepackage{xcolor}           % Enhanced color support (e.g., for tikz or hyperlinks)
\usepackage{color}

\def\tsc#1{\csdef{#1}{\textsc{\lowercase{#1}}\xspace}}
\tsc{WGM}
\tsc{QE}
\begin{document}
\input{acronyms}
\let\WriteBookmarks\relax
\def\floatpagepagefraction{1}
\def\textpagefraction{.001}

\shorttitle{Scoping Review on ML in Power System Protection}    

\shortauthors{J. Oelhaf et al.}  

\title [mode = title]{A Scoping Review of Machine Learning Applications in Power System Protection and Disturbance Management}

\author[1]{Julian Oelhaf}[orcid=0009-0008-8204-589X]
\cormark[1] % Corresponding author indication
\ead{julian.oelhaf@fau.de}
\ead[url]{https://lme.tf.fau.de}
\credit{Writing - original draft, Methodology, Investigation}

\author[2]{Georg Kordowich}[orcid=0000-0003-2225-7926]
\ead{georg.kordowich@fau.de}
\ead[url]{https://ees.tf.fau.de}
\credit{Writing - original draft, Methodology, Investigation}

\author[1]{Mehran Pashaei}[orcid=0009-0001-4755-0785]
\ead{mehran.pashaei@fau.de}
\credit{Investigation}

\author[3]{Christian Bergler}
\ead{c.bergler@oth-aw.de}
\ead[url]{https://oth-aw.de}
\credit{Writing - review \& editing, Supervision}

\author[1]{Andreas Maier}[orcid=0000-0002-9550-5284]
\ead{andreas.maier@fau.de}
\credit{Supervision, Project administration, Funding acquisition}

\author[2]{Johann Jäger}
\ead{johann.jaeger@fau.de}
\credit{Supervision, Project administration, Funding acquisition}

\author[1]{Siming Bayer}[orcid=0000-0003-2874-4805]
\ead{siming.bayer@fau.de}
\credit{Writing - review \& editing, Supervision, Project administration, Funding acquisition}

\cortext[1]{Corresponding author.}

\affiliation[1]{organization={Pattern Recognition Lab, Friedrich-Alexander-Universität Erlangen-Nürnberg},
            city={Erlangen},
            country={Germany}}

\affiliation[2]{organization={Institute of Electrical Energy Systems, Friedrich-Alexander-Universität Erlangen-Nürnberg},
            city={Erlangen},
            country={Germany}}

\affiliation[3]{organization={Department of Electrical Engineering, Media and Computer Science, Ostbayerische Technische Hochschule Amberg-Weiden},
            city={Amberg},
            country={Germany}}

\begin{abstract}
The integration of renewable and distributed energy resources has fundamentally reshaped modern power systems, challenging conventional protection schemes built around centralized generation. This scoping review synthesizes recent literature on machine learning (ML) applications in power system protection and disturbance management, following the PRISMA for Scoping Reviews framework. Based on over 100 publications, three key objectives are addressed: (i) assessing the scope of ML research in protection tasks; (ii) evaluating ML performance across diverse operational scenarios; and (iii) identifying methods suitable for evolving grid conditions.

Machine learning models often demonstrate high accuracy on simulated datasets; however, their performance under real-world conditions remains insufficiently validated. The existing literature is fragmented, with inconsistencies in methodological rigor, dataset quality, and evaluation metrics. This lack of standardization hampers the comparability of results and limits the generalizability of findings across different applications. To address these challenges, this review introduces a machine learning-oriented taxonomy for protection tasks, resolves key terminological inconsistencies, and advocates for standardized reporting practices. It further provides guidelines for comprehensive dataset documentation, methodological transparency, and consistent evaluation protocols, aiming to improve reproducibility and enhance the practical relevance of research outcomes.

Critical gaps remain, including the scarcity of real-world validation, insufficient robustness testing, and limited consideration of deployment feasibility. Future research should prioritize public benchmark datasets, realistic validation methods, and advanced ML architectures. These steps are essential to move ML-based protection from theoretical promise to practical deployment in increasingly dynamic and decentralized power systems.
\end{abstract}

%\item ML models show high accuracy on simulated data but lack real-world validation.
%\item Studies vary in task scope, dataset use, and evaluation, hindering comparisons.
%\item A unified taxonomy is proposed for protection tasks in power system research.
%\item Guidelines for dataset reporting and reproducibility support future benchmarks.

% Keywords, each keyword is seperated by \sep

\maketitle

\section{Introduction}
\label{sec:introduction}

The transformation of electric power systems is accelerating, driven by widespread integration of renewable and \ac{der}. This shift from traditional, centralized generation -- primarily coal, gas, and nuclear power -- towards decentralized, variable sources fundamentally reshapes electricity generation, transmission, distribution, and protection~\cite{walling_summary_2008, blaabjerg_distributed_2017}. The transition to \acp{ibr} and \acp{der} introduces significant operational changes, as these technologies differ fundamentally from conventional synchronous machines and are poorly supported by legacy grid infrastructures~\cite{potrc_sustainable_2021}. This shift challenges traditional protection systems, as \ac{ibr}-dominated grids face issues such as voltage instability, harmonic distortion, degraded power quality, reactive power imbalances, and synchronization difficulties~\cite{kataray_integration_2023}. These effects, once localized, now occur at scale with greater spatiotemporal variability, further diminishing the reliability of conventional protection schemes~\cite{blackburn_protective_2014}. Further complications arise from increasingly meshed distribution networks~\cite{biller_protection_2022}, emerging hybrid AC-DC architectures~\cite{prommetta_protection_2020}, and dynamic, bidirectional power flows. Traditional threshold-based protection methods, optimized for unidirectional fault currents, struggle to maintain selectivity, sensitivity, and rapid response under these conditions~\cite{protection_and_automation_b5_protection_2015, central_power_research_institute_protection_2017}.

\subsection{Motivation and Background}

Consequently, there is growing interest in leveraging data-driven approaches, particularly \ac{ai} and \ac{ml}, to enhance or replace conventional protection logic. Unlike fixed-threshold systems, \ac{ml} methods can learn nonlinear decision boundaries, adapt to changing conditions, and extract fault-relevant features directly from high-frequency waveform data~\cite{vaish_machine_2021}. These models generalize from data without relying on explicit heuristic rules. Recently, a sub-domain of \ac{ml}, \ac{dl} has achieved significant breakthroughs across diverse domains, proving effective at tackling complex, high-dimensional tasks.

Notable achievements in computer vision (ResNet~\cite{he_deep_2016}), natural language processing (BERT~\cite{devlin_bert_2019}, GPT~\cite{brown_language_2020}), strategic decision-making (AlphaGo~\cite{silver_mastering_2016}), and medical diagnostics~\cite{esteva_dermatologist-level_2017} have exemplified \ac{dl}'s capacity to automatically learn hierarchical structures and generalize effectively from unstructured inputs. Motivated by these successes, power systems research increasingly adopts \ac{ml} approaches, benefiting from the proliferation of high-resolution sensor data, computational advancements, and the need for adaptive protection schemes. Recent IEEE guidance underscores the practical promise of \ac{ai} for power system protection, highlighting scenarios where such methods surpass conventional techniques in speed, accuracy, and adaptability~\cite{abder_elandaloussi_practical_2023}.

\begin{figure}
    \centering
    \includegraphics[width=\textwidth]{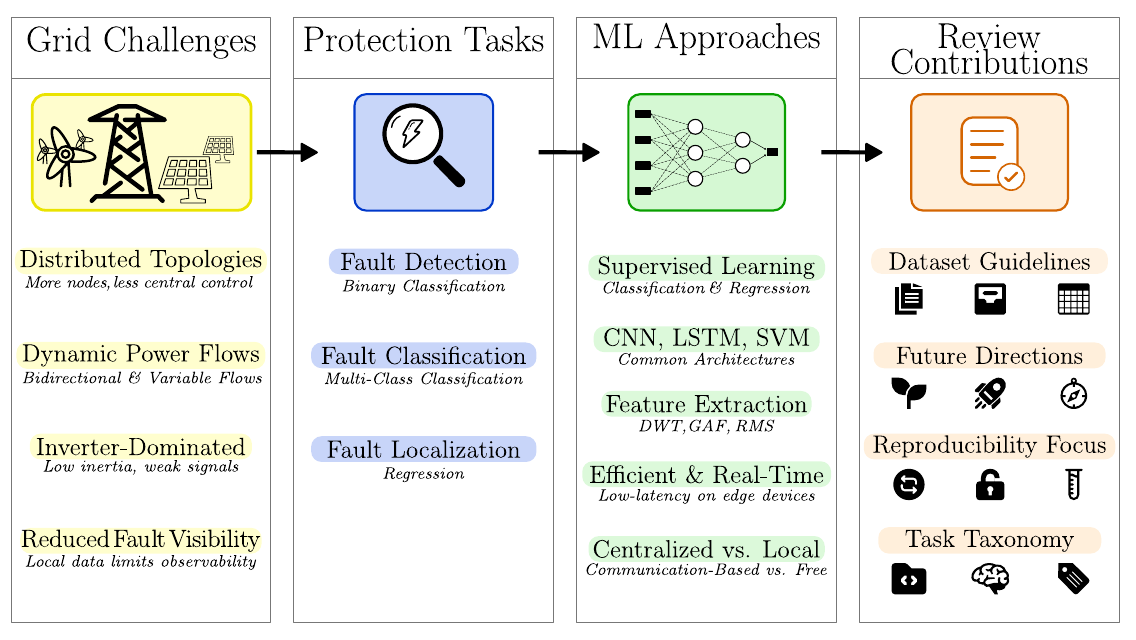}
    \caption{This overview figure illustrates the analytical flow of the review. The study begins with the emerging challenges of modern power grids, which motivate the need for intelligent protection systems. These challenges are mapped onto core protection tasks, which are examined through the lens of machine learning - focusing on common approaches, architectures, and deployment considerations. The review concludes with structured contributions that synthesize insights, promote reproducibility, and guide future research.}
    \label{fig:overview_figure_protection_review}
\end{figure}

As a result, the body of literature exploring \ac{ml}-based protection strategies has expanded significantly, encompassing diverse models, datasets, and evaluation strategies. However, methodological inconsistencies, varying performance metrics, and inadequate consideration of deployment scenarios remain prevalent. Thus, a structured synthesis is needed to comprehensively evaluate current research, identify promising trends, and reveal gaps that impede practical application.
To aid in this process, Figure~\ref{fig:overview_figure_protection_review} provides an overview of the review structure, highlighting key protection tasks, prevailing methodological patterns, and the main contributions of this work.

\subsection{Terminology and Scope}
\label{sec:terminology}

Inconsistent terminology complicates the review of \ac{ml} applications in power system protection. Central to this field is the concept of a \emph{fault}, denoting abnormal conditions such as insulation breakdown or unintended conductive paths (e.g., line-to-ground, line-to-line, or three-phase) that produce high currents, thermal and mechanical stress, and potential equipment damage. Protection schemes are designed to rapidly detect, classify, and localize faults so that the affected component can be isolated -- typically through breaker tripping -- to preserve system integrity and prevent cascading failures. These fault-related tasks remain the backbone of protection engineering and represent a primary focus of ML-based studies.

By contrast, \emph{disturbances} denote abnormal operating states that do not necessarily require immediate disconnection, such as islanding, voltage or frequency deviations, oscillations, harmonics, and other power quality issues. We use the term \emph{disturbance management} to describe the monitoring, detection, and mitigation of such events. While fault protection ensures safety by isolating permanent failures, disturbance management sustains stability and power quality. For \ac{ml}, disturbances are particularly challenging because their signatures are weaker, more distributed, and data-sparse compared to faults.

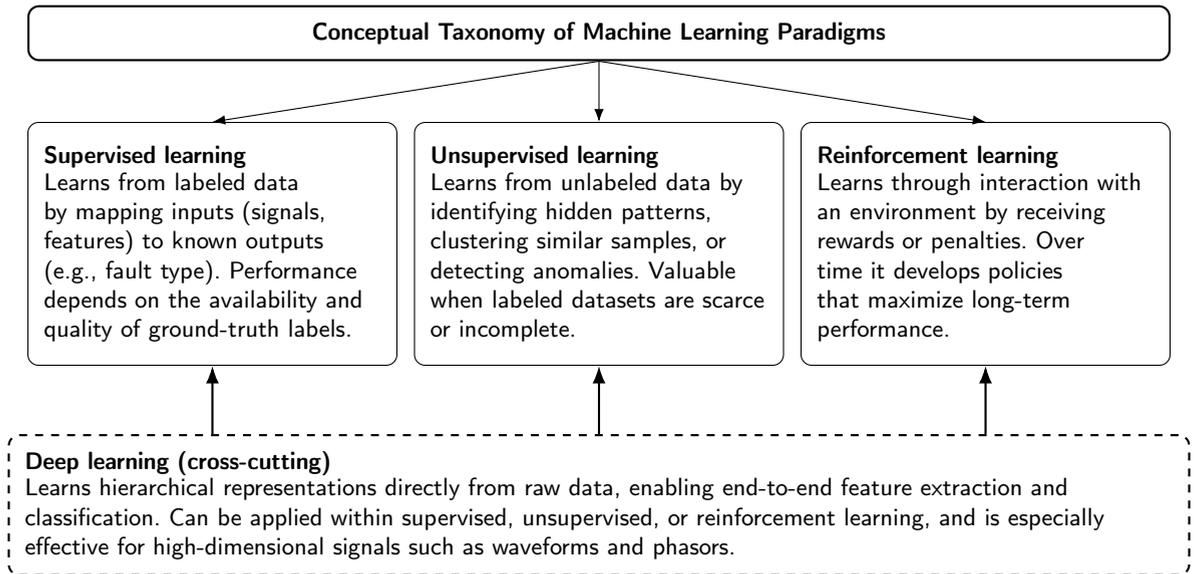
\begin{figure}
\centering
\begin{tikzpicture}[font=\small, >=Latex]

% ---------- styles ----------
\tikzset{
  box/.style={draw, rounded corners, align=left, inner xsep=6pt, inner ysep=5pt, fill=white, minimum height=32mm},
  banner/.style={draw, rounded corners, thick, align=center, inner xsep=8pt, inner ysep=6pt, fill=white},
  dlbox/.style={draw, rounded corners, align=left, inner xsep=6pt, inner ysep=6pt, dashed, thick, fill=white}
}

% ---------- banner ----------
\node[banner, text width=0.88\linewidth] (banner)
  {\textbf{Conceptual Taxonomy of Machine Learning Paradigms}};

% ---------- paradigm boxes (concept-focused, harmonized wording) ----------
\node[box, text width=0.27\linewidth, below=8mm of banner.south west, anchor=north west] (sup) {%
\textbf{Supervised learning}\\[-1pt]
Learns from labeled data by mapping inputs (signals, features) to known outputs (e.g., fault type). Performance depends on the availability and quality of ground-truth labels.
};

\node[box, text width=0.27\linewidth, below=8mm of banner.south, anchor=north] (unsup) {%
\textbf{Unsupervised learning}\\[-1pt]
Learns from unlabeled data by identifying hidden patterns, clustering similar samples, or detecting anomalies. Valuable when labeled datasets are scarce or incomplete.
};

\node[box, text width=0.27\linewidth, below=8mm of banner.south east, anchor=north east] (rl) {%
\textbf{Reinforcement learning}\\[-1pt]
Learns through interaction with an environment by receiving rewards or penalties. Over time it develops policies that maximize long-term performance.
};

% connectors from banner
\draw[->] (banner.south) -- (sup.north);
\draw[->] (banner.south) -- (unsup.north);
\draw[->] (banner.south) -- (rl.north);

% ---------- deep learning cross-cutting panel ----------
\node[dlbox, text width=0.92\linewidth, below=9mm of unsup.south] (dl) {%
\textbf{Deep learning (cross-cutting)}\\[-1pt]
Learns hierarchical representations directly from raw data, enabling end-to-end feature extraction and classification. Can be applied within supervised, unsupervised, or reinforcement learning, and is especially effective for high-dimensional signals such as waveforms and phasors.
};

% symmetric arrows from DL panel to each paradigm
\draw[->, thick] (dl.north -| sup) -- (sup.south);
\draw[->, thick] (dl.north -| unsup) -- (unsup.south);
\draw[->, thick] (dl.north -| rl) -- (rl.south);

\end{tikzpicture}
\caption{Conceptual taxonomy of machine learning paradigms in power system protection and disturbance management. 
Supervised learning learns from labeled data, unsupervised learning learns from unlabeled data, and reinforcement learning learns through interaction with an environment. 
Deep learning provides a cross-cutting family that learns hierarchical representations directly from raw data and enhances all three paradigms.}
\label{fig:ml_taxonomy}
\end{figure}

The reviewed methods are organized by \emph{learning paradigm}. \emph{Supervised learning} dominates, using labeled data for tasks such as fault classification or disturbance detection. \emph{Unsupervised learning} addresses clustering, anomaly detection, and dimensionality reduction when labels are unavailable. \emph{Reinforcement learning} remains exploratory, targeting adaptive and coordinated protection. Finally, \emph{deep learning} (e.g., convolutional, recurrent, and graph networks) enables hierarchical feature extraction across all paradigms (Figure~\ref{fig:ml_taxonomy}).

\subsection{Related Works}

Several recent reviews address various facets of protection in power systems, albeit typically focusing narrowly on specific contexts or tasks. Perez-Molina et al.~\cite{perez-molina_review_2021} and Sahebkar Farkhani et al.~\cite{sahebkar_farkhani_comprehensive_2024} specifically examine protection methods for multi-terminal \ac{hvdc} grids, emphasizing hardware implementations and fault-clearing strategies. Similarly, Chandio et al.~\cite{chandio_control_2023} extensively discuss control and protection methods specifically within \ac{mmc}-based \ac{hvdc} systems, whereas Mishra et al.~\cite{mishra_signal_2024} highlight signal processing and \ac{ai}-integrated methods for \ac{hvdc} networks. Panahi et al.~\cite{panahi_advances_2021} provide a broad review of fault localization methods for transmission networks, categorizing approaches such as traveling-wave techniques and impedance calculations. In contrast, De La Cruz et al.~\cite{de_la_cruz_fault_2023} investigate the localization of faults in distribution smart grids, emphasizing communication challenges and infrastructural considerations.
Bindi et al.~\cite{bindi_comprehensive_2023} explore both diagnostic and prognostic methods for medium and high-voltage lines, focusing primarily on predictive maintenance and algorithmic approaches. Alhamrouni et al.~\cite{alhamrouni_comprehensive_2024} broadly review \ac{ai}'s role across stability, control, and protection, albeit without detailed methodological comparisons across studies. Zaben et al.~\cite{zaben_machine_2024} systematically analyze \ac{ml}-based fault diagnosis, but restrict their coverage to AC microgrid environments. Meanwhile, broader perspectives such as Xie et al.~\cite{xie_massively_2023} emphasize general challenges and opportunities of digitized and \ac{ai}-driven grids, rather than detailed protection-specific methodologies. Additionally, Dai et al.~\cite{yang_review_2022} specifically survey \ac{ai} applications in relay protection but provide only a brief overview without comprehensive methodological synthesis. In contrast, our review uses a framework to systematically analyze \ac{ml} methods specifically for power system protection and disturbance management, focusing on identifying where these methods are applied and how they are implemented across different protection tasks, datasets, and grid types.

\begin{figure}
    \centering
    \includegraphics[width=\textwidth]{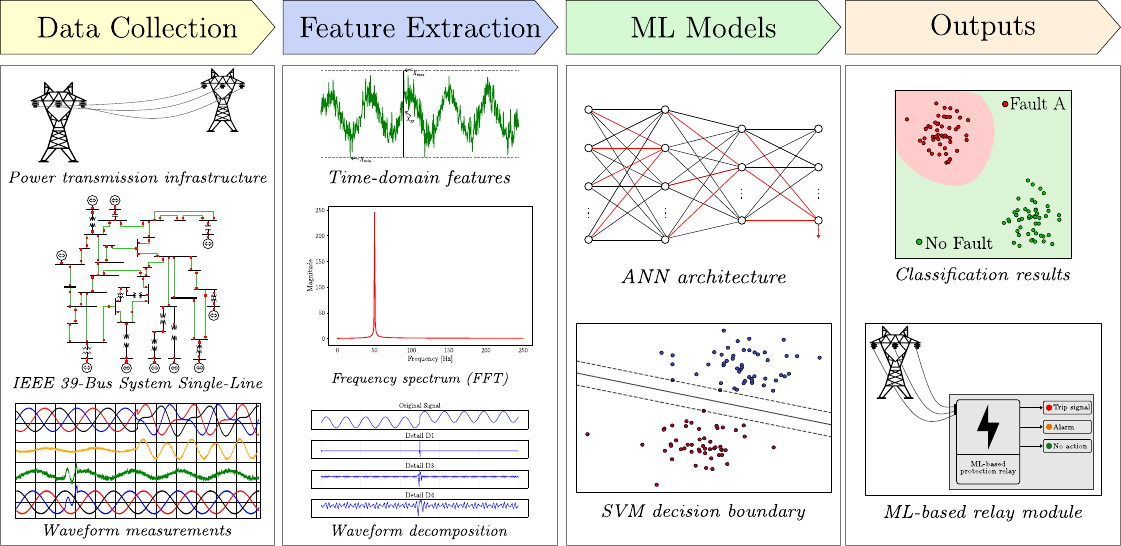}
    \caption{Classical \ac{ml} pipeline applied to power system protection and disturbance management. The first stage, \textbf{Data Collection}, entails acquiring voltage and current measurements from transmission infrastructure and system-level monitoring devices (e.g., protective relays, PMUs, and SCADA systems). The \textbf{Feature Extraction} stage focuses on transforming raw signal data into informative representations, including time-domain amplitude features, frequency spectra (e.g., via Fourier analysis), and multi-resolution decompositions (e.g., wavelet transforms), to emphasize fault-related characteristics. In the \textbf{\ac{ml} Models} stage, the extracted features are employed to train and evaluate classifiers such as \acp{ann} and \acp{svm}, enabling accurate discrimination between normal and faulty system states. Finally, the \textbf{Outputs} stage illustrates the model's decisions, such as identifying specific fault types (e.g., Fault A) versus no-fault conditions, and initiating system-level actions including trip signals, alarms, or holding operations. This end-to-end pipeline is designed to enhance protection reliability, improve situational awareness, and support faster, more adaptive disturbance management in modern, decentralized power systems. \textit{Figure recreated and adapted after~\cite{yang_machine_2021}.}}
    \label{fig:framework_of_applying_ml_to_power_system_protection}
\end{figure}

\subsection{Contributions of This Work}

In this paper, we present a scoping review guided by the \ac{prisma-scr} framework~\cite{tricco_prisma_2018, page2021prisma}. We systematically analyze over 100 studies addressing \ac{ml} applications in power system protection and disturbance management.
As modern power grids grow increasingly complex, driven by renewable integration, decentralized generation, and digitalized infrastructure, conventional protection methods face mounting limitations in ensuring system resilience and responsiveness. In response, \ac{ml}-based approaches have emerged as promising alternatives, offering adaptability to evolving grid conditions and the ability to extract actionable insights from heterogeneous data. However, the diversity of tasks, model architectures, and evaluation protocols across studies complicates comparison and impedes the development of generalizable conclusions.
An overview of a classical \ac{ml} pipeline tailored for power system protection and disturbance management is illustrated in Figure~\ref{fig:framework_of_applying_ml_to_power_system_protection}.
Scoping reviews are particularly suitable for such rapidly evolving and interdisciplinary domains, as they enable a structured and transparent synthesis of research activity, highlight underexplored areas, and support alignment between academic developments and practical deployment needs. Our review provides a comprehensive foundation to better understand the current landscape, guide future research, and inform the design of more robust, scalable, and effective protection schemes.

\subsection{Paper Organization}

The remainder of this paper is structured as follows: Section~\ref{sec:methods} details our review methodology, including the search strategy, eligibility criteria, and synthesis approach. Sections~\ref{sec:fault_detection} to~\ref{sec:fault_location} discuss findings categorized by protection task. Section~\ref{sec:analysis_future_directions} explores overarching trends, limitations, and future research opportunities. Section~\ref{sec:recommendations} provides recommendations to enhance methodological rigor and reproducibility. Section~\ref{sec:conclusion} concludes by highlighting key insights and implications for future work. For clarity, the key abbreviations used throughout this paper are summarized in Table~\ref{tab:abbreviations}.

\input{tables/tab_abbreviations}

\section{Methods}\label{sec:methods}

This study adopts a structured scoping review methodology, guided by the \ac{prisma-scr} framework~\cite{tricco_prisma_2018,page2021prisma}. The application of this standardized approach enhances transparency, reproducibility, and methodological rigor throughout the review process. The following subsections detail the procedures employed for data collection, screening, and analysis. A comprehensive \ac{prisma-scr} checklist is provided in Supplementary Table~\ref{tab:checklist}.

\emph{Eligibility Criteria}: This scoping review includes original research that contributes directly to the application of \ac{ml} in power system protection and disturbance management. Eligible studies address tasks such as \acl{fd}, \acl{fc}, \acl{fl}, protection coordination, or disturbance analysis using \ac{ml} methods. To ensure scientific rigor, only peer-reviewed journal articles and conference papers published in English between January 2020 and December 2024 were included. Review articles, meta-analyses, and other secondary literature were excluded to maintain a focus on original methodological contributions. These criteria were designed to capture recent and relevant advancements in \ac{ml}-based protection and disturbance management.

\emph{Information Sources}: A systematic search was conducted in the Scopus and IEEE Xplore databases using a standardized set of keywords aligned with the review objectives. The complete search queries are provided in Supplementary Table~\ref{tab:database_queries}. All searches were carried out on January 24, 2025.

The \emph{Search Strategy} was developed to capture literature at the intersection of power system protection and \ac{ml} techniques. We constructed structured queries by combining keyword groups across three thematic areas: protection-related concepts (e.g., relay protection, protection coordination), \ac{ml} methodologies (e.g., deep learning, reinforcement learning), and power system contexts (e.g., smart grids, \ac{hvdc}, AC-DC systems). The complete search expressions used for each database are presented in Supplementary Table~\ref{tab:database_queries}.
In addition to the systematic database queries, a manual search was conducted to identify relevant studies not captured through automated search, including recent publications and references cited in key articles.

\begin{figure}
    \centering
    \includegraphics[width=0.8\textwidth]{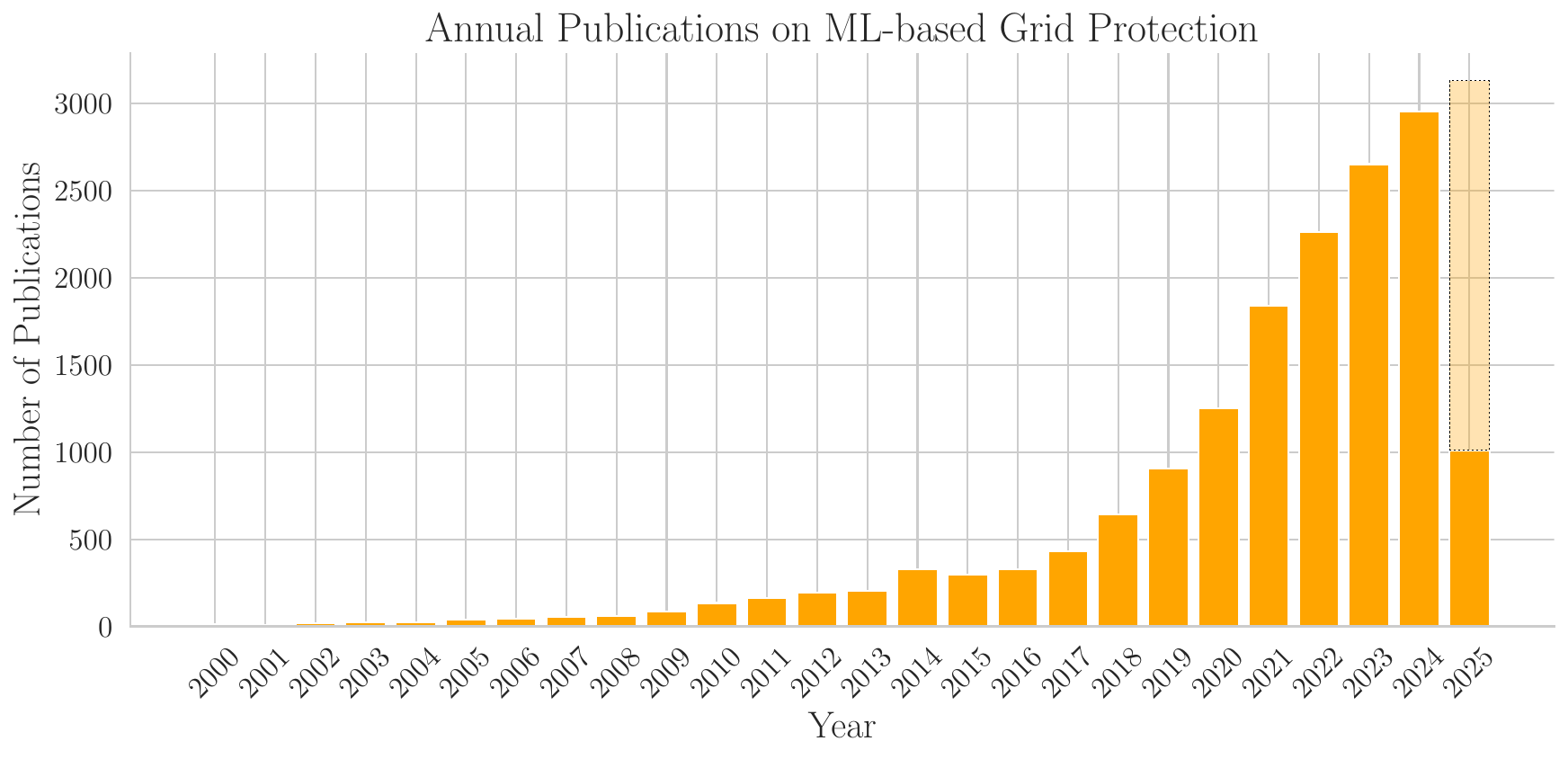}
    \caption{Trend in annual publications related to machine learning-based power system protection, based on results from Google Scholar using the same keyword combinations defined in the search strategy (see Table~\ref{tab:database_queries}). The plot illustrates a substantial increase in publication activity, particularly from 2018 onward, highlighting the growing research interest in this interdisciplinary field. The value for 2025 is extrapolated based on partial-year data.}

    \label{fig:publication_trend}
\end{figure}

To visualize the growth in this research area, a publication trend analysis was performed using Google Scholar, applying the same keyword combinations defined in the search strategy (see Table~\ref{tab:database_queries}). As illustrated in Figure~\ref{fig:publication_trend}, the number of publications on ML-based grid protection has increased considerably since 2018, reflecting the rising academic interest in this domain.

\emph{Inclusion and Exclusion Criteria}: To be included, studies had to apply \ac{ml} or \ac{ai} techniques to tasks directly related to power system protection and disturbance management. Relevant applications include fault detection, classification, localization, disturbance detection, and protection coordination. Eligible studies were required to clearly describe the dataset(s) used -- particularly those involving protection or disturbance scenarios -- and to specify the employed \ac{ml}/\ac{ai} methods along with a rationale for their selection. Moreover, studies had to report results supported by performance metrics, validation methods, or comparisons against baseline or state-of-the-art approaches.

Studies were excluded if they did not focus on protection or disturbance management, or did not employ \ac{ml}/\ac{ai} methods. Work of a purely theoretical nature, lacking simulations, case studies, or experimental validation, was also excluded. In addition, studies addressing cybersecurity issues such as intrusion detection or false data injection were excluded, as these fall outside the scope of physical protection and disturbance response. Forecasting and other operational tasks unrelated to protection or disturbance mitigation were likewise not considered.

\begin{figure}
\centering
\includegraphics[width=0.9\textwidth]{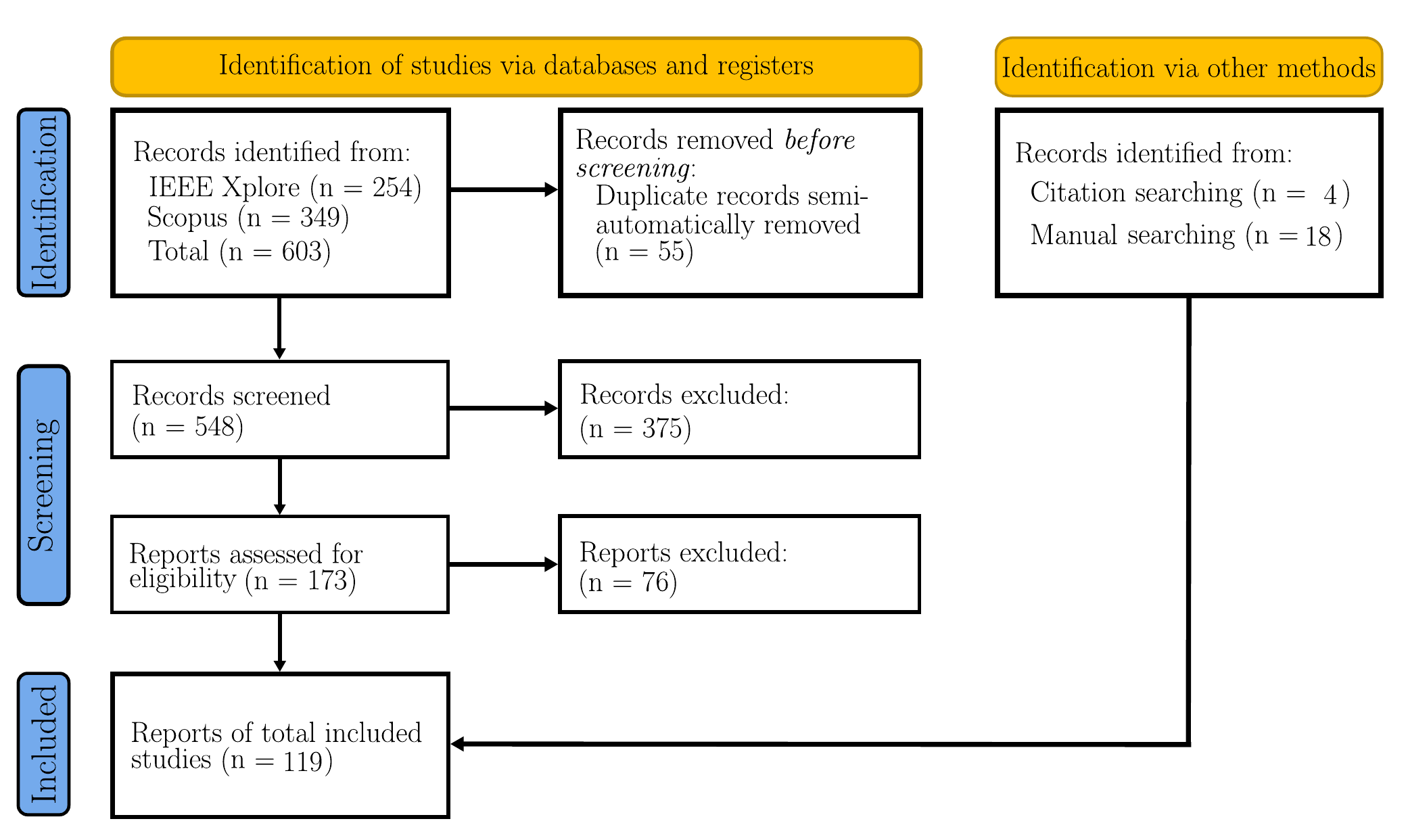}
\caption{PRISMA-ScR flow diagram illustrating the structured study selection process used in this review. A total of 603 records were retrieved from Scopus and IEEE Xplore and imported into \textit{Rayyan} for screening. After removing 55 duplicates, 548 unique studies were screened by title and abstract, resulting in the exclusion of 375 records. Full-text assessment led to the inclusion of 97 articles, with an additional 22 studies identified through manual search, totaling 119 included studies. The arrows in the diagram trace the sequential progression through identification, deduplication, screening, and eligibility assessment, clearly indicating how many records were retained or excluded at each step to ensure transparency and reproducibility.}
\label{fig:prisma_flowchart}
\end{figure}

\emph{Included Studies}: After applying these criteria, a total of \(119\) studies were included in the review, out of an initial pool of \(603\) studies.
The selection process followed a structured and transparent workflow aligned with the \ac{prisma-scr} framework. Initially, records retrieved from Scopus and IEEE Xplore were imported into \textit{Rayyan}~\cite{ouzzani2016rayyan}, a web-based tool that facilitates collaborative screening and duplicate removal. After eliminating duplicates, a two-stage screening process was conducted: first, titles and abstracts were reviewed to remove clearly irrelevant studies; second, the remaining articles underwent full-text assessment to determine final eligibility based on the inclusion and exclusion criteria. In addition to the systematic search, a manual screening of key references and recent publications was performed to capture relevant studies not retrieved by the database queries. The complete study selection process, including the number of records retained or excluded at each stage, is visualized in Figure~\ref{fig:prisma_flowchart}.

\emph{Synthesis of Results}: The selected studies were analyzed across key dimensions, including dataset type (e.g., simulation or real-world), targeted grid components (e.g., transmission or distribution), applied \ac{ml} models, and addressed protection tasks.
For clarity, studies were grouped into three protection-related categories: \ac{fd}, \ac{fc}, and \ac{fl}, reflecting ascending task complexity. When studies covered multiple tasks, they were classified under the most advanced, \ac{fl} over \ac{fc}, and \ac{fc} over \ac{fd}, to avoid redundancy and highlight their highest contribution.
Findings are presented through narrative summaries and comparative tables. The narratives focus on methods and application domains, while the tables summarize datasets, \ac{ml} techniques, and evaluation strategies for each task: fault detection (Table~\ref{tab:fault_detection}), fault classification (Table~\ref{tab:fault_classification}), and fault localization (Table~\ref{tab:fault_location}). Together, they provide a structured overview of current trends and challenges in \ac{ml}-based power system protection and disturbance management.

\section{Fault Detection in Power System Protection}\label{sec:fault_detection}
\subfile{sections/fault_detection}

\section{Fault Classification in Power System Protection}\label{sec:fault_classification}

\subfile{sections/fault_classification}

\section{Fault Localization in Power System Protection}\label{sec:fault_location}

\subfile{sections/fault_localization}

\section{Analysis and Future Research Directions}
\label{sec:analysis_future_directions}

\begin{figure}
    \centering
    \includegraphics[width=\textwidth]{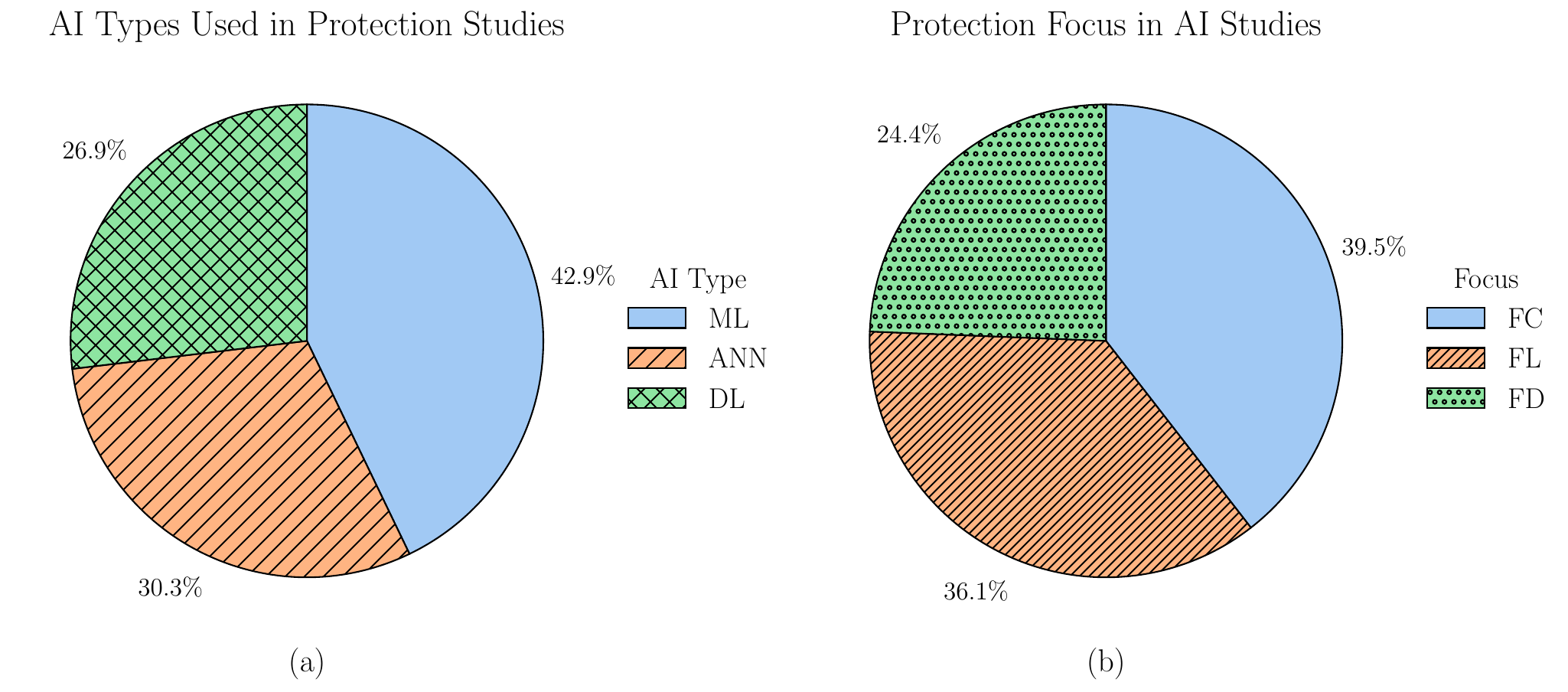}
    \caption{Distribution of AI types \textbf{(a)} and protection focus areas \textbf{(b)} in 119 selected studies on machine learning-based power system protection and disturbance management. The prevalence of \ac{ann} and \ac{dl} methods in (a) reflects current modeling trends, while (b) shows a strong emphasis on \ac{fc} and \ac{fl} as the dominant protection tasks.}

    \label{fig:combined-pie}
\end{figure}

This section synthesizes key patterns and limitations from the reviewed literature on \ac{ml}-based power system protection and disturbance management. Building on the task-specific findings in Sections~\ref{sec:fault_detection} to~\ref{sec:fault_location}, we highlight overarching trends in Section~\ref{sec:current_trends}, followed by recurring limitations in Section~\ref{sec:identified_gaps}. These insights inform targeted research priorities outlined in Section~\ref{sec:future_directions} to enhance reproducibility, robustness, and real-world viability of future \ac{ml}-based protection schemes.
As shown in Figure~\ref{fig:combined-pie}, the majority of reviewed studies employ machine learning-based approaches -- particularly \ac{ml} and \ac{ann} -- and focus primarily on either \ac{fc} or \ac{fl}, reflecting both a dominant modeling preference and a research emphasis on more complex protection tasks. Although papers were classified by their highest addressed task, many studies on \ac{fl} also perform \ac{fc} and \ac{fd}, and similarly, most \ac{fc}-focused works include \ac{fd} as a foundational step. In addition, the \ac{fd} category subsumes roughly 60\,\% of islanding detection papers, which were grouped under \ac{fd} due to their conceptual similarity.

\subsection{Current Trends in ML-Based Protection}
\label{sec:current_trends}

Recent advances in \ac{ml}-based protection reflect growing efforts to address the complexity, decentralization, and data challenges of modern power systems. This review analyzes studies along four dimensions -- protection task, data, methodology, and results -- focusing on factors such as communication assumptions, measurement types, model complexity, robustness, and real-time feasibility.

Protection tasks, \ac{fd}, \ac{fc}, and \ac{fl}, are typically framed as supervised learning problems using models like \ac{svm}, \ac{cnn}, and shallow \acp{nn} (especially \acp{mlp}), trained on synthetic data from Simulink, EMTDC, or PowerFactory. While simulations enable scalable testing, they lack real-world variability. Only a few studies incorporate real fault data or \ac{hil} validation.

Measurement approaches differ by grid type: transmission studies often use phasor-domain features (e.g., RMS, symmetrical components), while \ac{hvdc}, hybrid, and microgrid studies favor point-on-wave data to capture sub-cycle transients. Time–frequency transforms such as \ac{dwt}, \ac{fft}, and short-time fourier transform are common, though inconsistently applied.

Model complexity is generally low to balance latency and interpretability. While some works explore hybrids (e.g., \ac{cnn}–\ac{lstm}), simpler models dominate. \ac{rl} is occasionally used for adaptive control; transformer-based methods are rare. About one-quarter of studies assume communication infrastructure but rarely assess latency or reliability, with communication-free models preferred for simplicity.

Evaluation in the reviewed studies is predominantly accuracy-centric, with limited attention to deployment factors. Robustness to noise, data loss, or topology changes is rarely assessed, runtime and hardware requirements are seldom reported, and validation with field data or hardware-in-the-loop setups remains scarce. In most cases, the reported metric ``accuracy'' refers to \emph{classification accuracy}, defined as
\begin{equation}
    \text{Accuracy} = \frac{\text{TP} + \text{TN}}{\text{TP} + \text{TN} + \text{FP} + \text{FN}},
\end{equation}
where $\text{TP}$, $\text{TN}$, $\text{FP}$, and $\text{FN}$ denote true positives, true negatives, false positives, and false negatives, respectively. Although accuracy is straightforward and widely used, it becomes misleading in imbalanced datasets where the majority class dominates. In such cases, a model may achieve high accuracy by consistently predicting the majority class while failing to capture minority fault types. To address this, task-specific measures such as precision (positive predictive value), recall (sensitivity), and their harmonic mean, the F1-score, provide a more nuanced assessment of classification performance. An overview of these metrics, including their definitions, strengths, and limitations, is summarized in Table~\ref{tab:ml_metrics}.

\begin{table}[tb]
\centering
\caption{Overview of common evaluation metrics in ML-based protection and disturbance management studies.}
\label{tab:ml_metrics}
\begin{tabularx}{\textwidth}{@{}lXX@{}}
\toprule
\textbf{Metric} & \textbf{Definition} & \textbf{Strengths and Limitations} \\
\midrule
Accuracy & Ratio of correctly predicted samples to total samples. 
& Intuitive and widely used, but may obscure performance under class imbalance. \\
Precision & Fraction of correctly predicted positives among all predicted positives. 
& Reduces false alarms, but ignores missed detections. \\
Recall (Sensitivity) & Fraction of correctly predicted positives among all actual positives. 
& Captures detection ability, but high recall may increase false positives. \\
F1-Score & Harmonic mean of precision and recall. 
& Balances false positives and false negatives; more robust than accuracy on imbalanced data. \\
\bottomrule
\end{tabularx}
\end{table}

\subsection{Identified Gaps and Limitations}
\label{sec:identified_gaps}

Despite significant progress in applying \ac{ml} to power system protection, several limitations continue to impede reproducibility, comparability, and deployment. A primary challenge is the predominant reliance on simulation data. While simulations offer scalability, they cannot capture real-world variability. Only a small subset of studies incorporate real fault recordings, \ac{hil} setups, or field measurements, leaving generalizability largely untested.

Terminological inconsistencies further obstruct progress. Terms like fault detection, classification, diagnosis, and localization are often used interchangeably, despite referring to distinct tasks with different input requirements and evaluation metrics. These ambiguities, often reflecting differing task definitions in \ac{ee} and \ac{ml}, complicate comparison and reproducibility. The taxonomy in Sec.~\ref{sec:taxonomy} seeks to standardize definitions and promote interdisciplinary clarity.

A lack of open, standardized datasets hinders benchmarking. Even when simulated or proprietary data are used, essential metadata such as sampling rates, voltage levels, and fault types are frequently omitted. While concerns around confidentiality are valid, insufficient documentation limits transparency and reuse. To address this, we propose a dataset documentation checklist in Sec.~\ref{sec:dataset}, applicable to both public and private data sources.

Evaluation practices are similarly inconsistent. While accuracy is widely reported, task-specific metrics -- such as F1-score, detection latency, or localization error (see Sec.~\ref{sec:fault_location}) -- are often omitted. Few studies assess robustness under noise, missing data, or varying operating conditions. This lack of scenario-aware evaluation prevents fair comparison and limits confidence in real-world applicability.

Real-time feasibility is also underexplored. Most studies neglect latency, memory, and throughput constraints, despite the strict timing requirements of protection systems. Without runtime profiling, it remains unclear whether models can run on digital relays, edge devices, or microcontrollers. Future work should prioritize hardware-aware design and consider efficiency alongside accuracy.

\subsection{Guiding Future Research Directions}
\label{sec:future_directions}

To overcome the limitations outlined above, future research should focus on creating publicly accessible benchmark datasets with well-defined metadata. This includes clearly specifying voltage levels, sampling rates, grid topologies, and fault scenarios. Consistent task definitions, a standardized fault taxonomy (e.g., distinguishing DC pole-to-pole, pole-to-ground, and AC short-circuits), and unified terminology are also essential for enabling reproducible and comparable research. Without such information, meaningful comparison across studies remains impossible. At present, no commonly accepted taxonomy exists, and fault types are often described in inconsistent or ambiguous ways across publications. Addressing this gap is therefore a prerequisite for developing robust benchmarks and drawing generalizable conclusions from ML-based protection studies.

Evaluation metrics must capture not only accuracy, but also robustness to noise, missing data, and temporal misalignment. Runtime behavior and compatibility with field hardware should be included as standard performance indicators. For \ac{fl} tasks in particular, spatial accuracy should primarily be reported as a percentage of the total line length to enable fair comparison across different scenarios; physical units (e.g., meters) may be included additionally to support practical interpretation. At the same time, many current datasets are ``too clean,'' often omitting common-mode disturbances, background noise, or switching events. Expanding datasets to incorporate these effects is necessary for testing model specificity and resilience under realistic operating conditions.

Bridging the gap between academic models and operational use cases will require validation under realistic conditions, including pilot deployments and collaborations with utilities. Although \ac{hil} simulations are useful for demonstrating real-time performance and hardware integration, they are limited in their ability to assess model robustness under real-world disturbances due to their reliance on simulation-based models. However, \ac{hil} becomes significantly more valuable when combined with real fault recordings, such as COMTRADE files, enabling researchers to verify model reliability under authentic disturbances.

In this context, digital twins emerge as a particularly promising enabler. By providing high-fidelity, real-time virtual representations of power grids, they bridge the gap between simulation-only validation and real-world deployment, and they facilitate the generation of standardized benchmark datasets with ground-truth labels.

Research should also extend beyond transmission grids to include microgrids, \ac{hvdc}, and hybrid AC-DC systems. Transfer learning and domain adaptation may help build models that generalize across these diverse operating environments. Beyond this, feature extraction methods such as wavelet and time-frequency transforms deserve systematic benchmarking, as they remain widely applied in transient-rich domains (e.g., HVDC) yet lack rigorous comparative evaluation.

Looking ahead, \ac{ml}-based protection systems will need to satisfy increasingly demanding real-world constraints. Speed remains a fundamental requirement: protection decisions must be made within milliseconds to ensure timely fault clearing. This calls for low-latency models that can operate reliably on embedded devices such as digital relays or substation controllers. In this context, model efficiency becomes as important as accuracy. Lightweight architectures and model compression techniques -- including pruning, quantization, and knowledge distillation via teacher-student frameworks -- will be key to achieving real-time performance under resource constraints. Moreover, robustness under varying levels of \ac{der} penetration should be explicitly tested, as increasing renewable integration changes system dynamics and fault signatures.

Another essential area of research concerns robustness to communication issues. As protection strategies become more distributed, they increasingly depend  on data exchange between devices. Delays, packet loss, and link failures must be expected and managed. Models should either explicitly account for communication uncertainty or implement fallback mechanisms that maintain safety under degraded conditions. Hybrid input strategies that combine slow but resilient phasor data with high-resolution waveform signals could further enhance classification accuracy and robustness in distributed and hybrid AC/DC settings.

A related challenge is input robustness. \Ac{ml}-based protection systems must handle noisy or atypical inputs, including scenarios that unintentionally resemble adversarial examples. Although targeted attacks such as data poisoning or model tampering are theoretically possible, they are less relevant in secured environments where direct manipulation of sensor data is often more effective. Research should prioritize adversarial robustness, secure deployment, and update mechanisms that support auditability and traceability.

Explainability remains a critical challenge, as protection decisions are inherently safety-critical and must be trusted by human operators. Black-box \ac{ml} models are difficult to adopt in this domain unless their behavior can be interpreted in a transparent and reliable manner. Methods from the broader field of explainable AI (XAI) -- such as SHAP or LIME -- offer post-hoc explanations that can enhance operator trust, but have so far been rarely applied in power system protection. At the same time, it is important to distinguish these methods from approaches that embed prior knowledge or physical constraints into the model (e.g., known-operator networks or physics-informed neural networks), which pursue interpretability through structure rather than explanation. Both directions remain largely unexplored in the reviewed literature. Ultimately, progress in post-hoc explanation, confidence calibration, and inherently interpretable architectures must go hand in hand with broader access to real-world, labeled fault data. Enabling data sharing, improving dataset documentation, and developing standardized synthetic-to-real validation pipelines are crucial steps toward building trustworthy and deployable models.

Lastly, modern \ac{ml} techniques remain underexplored. Transformer architectures~\cite{ding_novel_2022}, \acp{gnn}~\cite{chen_interaction-aware_2023}, and \acp{pinn}~\cite{zideh_physics-informed_2024} show promise in structured domains and may be valuable in protection tasks, for example by modeling spatiotemporal dependencies or physical constraints. An emerging direction is \ac{kol}, which embeds known operators, such as grid equations or relay logic, into trainable models~\cite{maier_learning_2019,maier_known_2022}. This hybrid strategy improves interpretability and robustness while reducing model complexity, and has the potential to enhance fault localization accuracy by embedding physical grid constraints directly into the learning process. Although widely adopted in medical imaging, \ac{kol} remains largely unexplored in power system protection.

\section{Recommendations for Structured and Reproducible Research}
\label{sec:recommendations}

Building on the trends, limitations, and opportunities identified in Sections~\ref{sec:current_trends} and~\ref{sec:identified_gaps}, this section presents practical recommendations to improve the transparency, comparability, and reproducibility of \ac{ml}-based research in power system protection.
Section~\ref{sec:guidelines_open_research} presents general guidelines for open and structured research, data sharing, evaluation practices, and model documentation. Together with the unified taxonomy in Section~\ref{sec:taxonomy} and the dataset documentation standards in Section~\ref{sec:dataset}, these recommendations support standardized benchmarking and reproducible comparison across studies.

\subsection{Guidelines for Transparent and Reproducible ML Research in Power System Protection}
\label{sec:guidelines_open_research}
Despite significant progress in \ac{ml}-based protection, the field still suffers from limited reproducibility and inconsistent evaluation practices. Variability in performance metrics, undocumented model configurations, and inaccessible datasets or codebases hinder meaningful comparison and slow down translation to practice. To address these issues, we outline core guidelines grounded in open-source principles, emphasizing transparency, comparability, and reproducibility across the full research pipeline.

Studies should report more than aggregate accuracy. Task-specific metrics -- such as precision, recall, F1-score, and class-wise performance -- are essential for meaningful comparison, especially in imbalanced or multi-class settings. Yet, we found that 23.5\,\% of studies did not report any performance metric beyond general statements of accuracy, limiting interpretability and hindering fair comparison across methods. For \ac{fd}, detection latency and false negative rates are particularly relevant, as they directly impact operational reliability. Robustness should be evaluated under realistic conditions, yet only a minority of studies conduct such analyses. Necessary ablation studies include evaluations under varying fault types, fault resistances, fault locations, load fluctuations, \ac{der} outages, and different network topologies. It is also important to check whether models can distinguish faults from normal but disruptive events such as inrush currents, line switching, or energization.

Notable examples such as~\cite{prasad_tiwari_enhancing_2021} demonstrate the importance of systematically assessing performance across these dimensions. Noise robustness has been examined in~\cite{najafzadeh_fault_2024, taheri_fault-location_2021, taheri_novel_2022, yousaf_intelligent_2022, ren_fault_2024}, and generalization to unseen fault types in~\cite{diefenthaler_artificial_2023}, while missing data remains largely unexplored in the reviewed literature. An example analysis is provided in our prior work~\cite{oelhaf_impact_2025}. Similarly, runtime efficiency~\cite{afrasiabi_fault_2022, tong_semi_2022, gnanamalar_cnnsvm_2023}, memory usage, and deployment constraints are rarely reported. While these factors may not limit small or moderately sized models, they become relevant for large-scale deployments, resource-constrained environments, or real-time protection logic. Their omission hinders comparability and makes practical applicability harder to assess.
These omissions hinder comparability and practical relevance.

Transparency in experimental design is essential. Key aspects -- such as model architecture (e.g., centralized vs. decentralized), input representation (phasor, waveform, frequency domain), output targets, and evaluation protocols -- should be clearly stated. Training details (e.g., network layers, activation functions, optimizer settings, batch size, and epochs) and hyperparameter configurations must be reported. These are often omitted from the main text and instead shared via appendices or external repositories like GitHub, which is effective if properly documented and maintained.

Adopting open-source practices is vital for community progress. While only 1.7\,\% of the reviewed studies publicly released their code or models, we recommend that future work consistently share source code, preprocessing scripts, trained models, and evaluation tools. If proprietary or sensitive data cannot be shared, detailed documentation of data formats, preprocessing pipelines, and experimental protocols remains essential. These practices form the foundation for building rigorous, scalable, and practically relevant \ac{ml} solutions in power system protection.

\subsection{Proposed Taxonomy}
\label{sec:taxonomy}

Terminological inconsistencies remain a major barrier in \ac{ml}-based power system protection research, largely due to its interdisciplinary nature. To improve communication, reproducibility, and benchmarking, we propose a unified taxonomy of protection tasks, aligning conventional protection concepts with common \ac{ml} formulations.

\emph{Data window} refers to the segment of voltage and current waveforms or phasor measurements used for analysis, typically spanning \qtyrange{1}{100}{\milli\second} to comply with relay response times~\cite{blackburn_protective_2014,horowitz_power_2023}. While training data are often aligned to fault inception, real-time models operate on continuously sampled signals.
Discrete measurement points sampled within this window form the multivariate time-series input to \ac{ml} models. The sampling rate (number of measurements per second), number of samples, and number of channels together determine the input dimensionality and influence the model’s ability to capture transient behavior in real-time.

\emph{Fault detection} determines whether a fault has occurred and is consistently framed as a binary classification task. This mirrors conventional protection logic (e.g., overcurrent and distance relays)~\cite{blackburn_protective_2014,horowitz_power_2023} and is widely adopted in recent literature~\cite{yousaf_novel_2023,cano_integrating_2024,oelhaf_systematic_2025,najafzadeh_fault_2024,poudel_zonal_2022}.

\emph{Islanding detection} is the task of identifying when \ac{dg} or a \ac{mg} is disconnected from the main utility grid but continues to energize a local subsection of the grid.
Conventionally addressed by passive, active, or hybrid techniques, the problem in \ac{ml}-based protection is typically formulated as a binary classification, where islanding must be distinguished from non-islanding disturbances such as faults, switching, or load variations~\cite{ali_hierarchical_2021,hussain_islanding_2023,nsaif_island_2023}.

\emph{Fault classification} identifies the type of fault, such as single-phase-to-ground, line-to-line, or three-phase faults, and is typically formulated as a multi-class classification task. High-impedance faults often receive special attention due to their detection difficulty~\cite{tsotsopoulou_protection_2023,cano_integrating_2024,najafzadeh_fault_2024,ren_fault_2024,ghaderi_high_2017}.

\emph{Fault Area Classification}, instead of \emph{Fault Zone Classification}, is suggested to avoid confusion with conventional distance protection zones, as it assigns faults to protection areas, only sometimes consistent with the zone reach of distance protection relays~\cite{blackburn_protective_2014}. Though less common, this task is gaining importance in advanced distribution protection settings~\cite{poudel_zonal_2022,banihashemi_novel_2023}.

\emph{Fault line identification} involves determining the specific line, feeder, or substation where the fault occurred. It is increasingly relevant in centralized schemes and \ac{wams}-based architectures, where distinguishing between multiple monitored elements is essential. This task is typically formulated as a multi-class classification~\cite{oelhaf_systematic_2025,zhu_fault-line_2018,he_single-phase_2022}. A related but under-defined task is \emph{fault bus identification}, which extends the concept to pinpoint the faulted bus in an interconnected grid.

\emph{Fault localization} estimates the fault’s location along a transmission line, usually expressed as distance (in kilometers or percentage of line length). In \ac{ml}, this task is naturally modeled as a regression problem, with traditional impedance- and traveling-wave-based methods forming the foundation~\cite{blackburn_protective_2014,najafzadeh_fault_2024,ren_fault_2024}.

\emph{Centralized} or \emph{decentralized} architectures must be clearly specified when describing an \ac{ml}-based scheme. Centralized schemes rely on synchronized data from multiple substations (grid-level centralized) or can also be centralized at the substation level, where data from different devices within a single substation are aggregated. In contrast, decentralized schemes use only local measurements. This distinction has significant implications for communication infrastructure, latency, and reliability~\cite{blackburn_protective_2014}.

This taxonomy promotes consistent definitions across studies, enabling clearer problem formulation and improving the comparability and reproducibility of \ac{ml}-based protection research.

\textbf{Extended Task Definitions and Distinctions}: Beyond detection, classification, and localization, related terms are often used inconsistently across the literature. \emph{Fault diagnosis} typically refers to a composite task that combines detection, classification, and localization. \emph{Fault analysis}, in contrast, is used more broadly to describe the overall evaluation of power system behavior under fault conditions, including detection, classification, localization, and parameter estimation.
\emph{Fault isolation} describes the process of disconnecting the faulted section from the grid to prevent cascading failures, \emph{restoration} involves reconfiguring the network to reroute power after a fault has been cleared, and \emph{protection coordination} (or relay coordination) ensures the correct tripping sequence among multiple protection devices, often relying on teleprotection schemes in decentralized architectures to enable fast and selective operation.

\emph{Fault direction discrimination} refers to determining whether a fault lies in the forward or reverse direction relative to a relay's location, which is critical for systems with bidirectional power flow or in meshed networks. \emph{Faulty phase selection} aims to identify which specific phase(s) are involved in the fault. This is particularly important for unbalanced or asymmetrical faults and is often a prerequisite for accurate classification and localization~\cite{yadav_overview_2014}. It is strongly related to \ac{fc}, as knowledge of the affected phases often implies the fault type.

Related tasks include \emph{disturbance classification}, which involves identifying and categorizing non-fault events (e.g., switching, load fluctuations), and \emph{anomaly detection}, which targets deviations from expected behavior using often unsupervised or self-supervised methods. \emph{Adaptive protection} dynamically updates relay settings in response to changing grid conditions. Finally, \emph{fault identification} refers to estimating specific fault parameters such as impedance, resistance, or location index.

\subsection{Proposed Dataset Guidelines}
\label{sec:dataset}

High-quality data is essential for advancing \ac{ml}-based protection research. However, obtaining real-world fault recordings remains highly challenging. Legal and regulatory barriers often prevent data sharing due to strict confidentiality agreements and critical infrastructure protection laws. Even anonymized datasets may fall under data privacy regulations such as the GDPR. Organizational and logistical constraints further limit access: many utilities lack the mandate or infrastructure to archive high-frequency data in a structured, accessible format. Where datasets do exist, they are frequently stored in proprietary systems with inconsistent or undocumented metadata, complicating retrieval and interpretation. Technical failures during fault events -- such as communication loss, buffer overflows, or device limitations -- can result in corrupted or incomplete recordings. Moreover, spatial gaps in sensor deployment reduce observational completeness and introduce sampling bias.

Despite these challenges, a few commendable efforts have contributed to improving data availability. Bera et al.~\cite{bera_hybrid_2024} and Alhanaf et al.~\cite{alhanaf_intelligent_2023} published their dataset alongside the study.
Public datasets were used in~\cite{kumar_improved_2024} and~\cite{dhingra_machine_2023}.
Other works note that data are available upon request from the corresponding author, including~\cite{uddin_protection_2022, mehdi_squaring_2023, cano_integrating_2024, gnanamalar_cnnsvm_2023, ren_fault_2024, kumari_enhancing_2024}, though this approach offers limited transparency and reproducibility.
A notable example is provided by Galkin et al.~\cite{galkin_toolset_2023}, who released both the dataset and code, setting a strong precedent for open-source contributions in this field.

Out of the 119 studies reviewed, only 1.7\,\% published their own datasets, and another 1.7\,\% relied on publicly available ones. An additional 10.9\,\% indicated that their data are available upon request, while 2.5\,\% explicitly stated that their datasets are confidential. The remaining 82.5\,\% did not provide access to any data. Notably, 21.0\,\% employed standardized topologies -- such as the IEEE 6-bus, 34-bus, or 123-bus systems -- which may support structural reproducibility. However, meaningful replication is only possible if all simulation parameters, signal characteristics, and configuration details are fully documented. These figures highlight an urgent need for more transparent data-sharing practices and the development of standardized, openly available benchmark datasets to foster reproducibility and comparability in \ac{ml}-based power system protection research.

In addition to access constraints, other challenges affect dataset quality. Faults are rare and geographically localized events, often resulting in sparse and highly imbalanced datasets. Certain fault types may not be observed at all during the recording period. Ground-truth labeling is particularly difficult and may require fusing multiple data sources or expert interpretation, making large-scale annotation costly and error-prone. As a result, most studies rely on simulated data, where fault conditions can be precisely defined. However, only 16.0\,\% of the reviewed studies used real-world data, and simulation fidelity varies widely. Documentation practices are also inconsistent: 34.5\,\% of studies do not specify the voltage level, 52.1\,\% omit the sampling frequency -- both essential for interpreting signal characteristics -- and 7.6\,\% fail to state the simulation tool or version. These gaps limit reproducibility, hinder comparability across studies, and complicate efforts to benchmark \ac{ml}-based protection methods.

To address these issues, we introduce two reference tables. Table~\ref{tab:reporting_standards} specifies essential metadata that should accompany any dataset to ensure clarity and consistency. Table~\ref{tab:benchmark_recommendations} outlines key design principles for constructing publicly available benchmark datasets that reflect diverse operating conditions and support standardized evaluation of \ac{ml}-based protection methods. We encourage journals, conferences, and peer reviewers to adopt these recommendations as minimum standards for dataset publication in this domain.

\input{tables/tab_reporting_standards}

\input{tables/tab_public_power_system_benchmark_datasets}

Well-documented datasets are essential for transparent and reproducible research. These guidelines aim to support standardized reporting and future benchmarks that reflect realistic scenarios and enable fair comparison across methods.

\section{Summary \& Conclusion}
\label{sec:conclusion}

This paper presents a comprehensive review of \ac{ml} applications in power system protection, with a focus on fault detection, classification, and localization. The review identifies increasing methodological diversity, a strong reliance on supervised learning, and enduring challenges in task formulation, dataset design, and evaluation methodology. Widely adopted models include \acp{mlp}, \acp{svm}, and \acp{cnn}, selected for their balance between accuracy and computational efficiency. Only a limited number of recent studies have explored more sophisticated architectures -- such as hybrid \ac{cnn}-\ac{lstm} networks, attention mechanisms (e.g., \acp{vit}), and multi-agent reinforcement learning -- to better capture spatiotemporal dependencies in grid dynamics.

Simulated data, predominantly generated using MATLAB/Simulink or PSCAD, remains the primary basis for model training and evaluation. Most studies focus on \ac{hv} transmission systems (110\,kV and above), typically with radial or weakly meshed topologies. \ac{lv} grids, microgrids, and hybrid AC-DC systems remain underrepresented. Dataset sizes range from hundreds to millions of samples, with sampling frequencies from 1-100\,kHz. However, key metadata -- such as signal representation (e.g., EMT, RMS, phasor), fault parameters, and data partitioning strategies -- are frequently omitted, limiting reproducibility and comparability.
Although many models report near-perfect accuracy, most evaluations are limited to noise-free, fully observable, and latency-free conditions. Robustness to measurement noise, missing data, communication delays, and other real-world constraints is rarely tested. Runtime analysis, hardware requirements, and fault recovery performance are often neglected, hindering practical deployment.

Three overarching limitations emerge: (i) near-perfect accuracies on synthetic datasets highlight a lack of realism in evaluation protocols; (ii) the absence of standardized datasets, reporting formats, and grid configurations impedes reproducibility and benchmarking; and (iii) validation practices rarely address deployment-critical aspects such as communication failures, noisy signals, or delayed responses.

We emphasize the importance of evaluation protocols that prioritize robustness, interpretability, and open-source practices. To complement this, we propose: (i) a task-oriented taxonomy that aligns protection objectives with standard \ac{ml} formulations; and (ii) dataset reporting guidelines to improve transparency and reproducibility.
Future research should prioritize benchmark datasets, hybrid validation approaches (e.g., \ac{hil} testing), and comprehensive coverage of underrepresented grid types. In addition to model accuracy, metrics such as runtime efficiency, scalability, and resilience to cyber and communication failures must be considered. Promising but underutilized techniques -- including \acp{gnn}, \acp{pinn}, and known operator learning -- offer opportunities to embed physical knowledge and exploit topological structure. Transfer learning, knowledge distillation, and domain adaptation may further enhance model generalization under varying grid conditions.

Advancing \ac{ml}-based protection requires shared standards, realistic validation, and reproducible methodologies. With these in place, the community can move toward reliable, interpretable, and field-ready protection solutions for future power systems.

\printcredits

\section*{Declaration of competing interest}
\noindent
The authors declare that they have no known competing financial interests or personal relationships that could have appeared to influence the work reported in this paper.

\section*{Acknowledgment}
\noindent
This project was funded by the Deutsche Forschungsgemeinschaft (DFG, German Research Foundation) - 535389056.

\bibliography{refs}

\newpage
\bio{}
\noindent
\small \textbf{Julian Oelhaf} is a PhD student at the Pattern Recognition Lab, Friedrich-Alexander-Universität Erlangen-Nürnberg (FAU), under the supervision of Dr.-Ing. Siming Bayer and Prof. Dr.-Ing. habil. Andreas Maier. He began his doctoral studies in 2024 after completing a Master’s degree in Computer Science at FAU. His research focuses on \ac{ml} applications for \ac{fd} and protection coordination.
\\[0.5cm] % Adds space after each biography

\noindent
\textbf{Georg Kordowich} studied Electrical Energy Technologies at the Friedrich-Alexander-Universität Erlangen-Nürnberg (FAU). After completing his Master’s Degree in 2021 he started working as a PhD Student at the Institute of Electrical Energy Systems at FAU under the supervision of Prof. Dr. Johann Jäger. 
His main research focus is the application of \ac{ml} for \ac{fd} and protection coordination. \\[0.5cm] % Adds space after each biography

\noindent
\textbf{Mehran Pashaei} is a Master's student in Medical Engineering at Friedrich-Alexander-Universität Erlangen-Nürnberg (FAU), specializing in medical image and data processing. He holds a Bachelor's degree in Electrical and Electronics Engineering from Ege University, Turkey. His interests include robotics and the application of \ac{ml} in healthcare, with prior experience in AI-driven navigation for autonomous robots.
\\[0.5cm]

\noindent
\textbf{Christian Bergler} received his M.Eng. degree in Information Technology and Automation from Ostbayerische Technische Hochschule (OTH) Amberg-Weiden, Germany, in 2016. He earned his doctorate (Dr.-Ing.) in Computer Science from Friedrich-Alexander-Universität Erlangen-Nürnberg (FAU), Germany, in September 2023, graduating with highest distinction (summa cum laude). He is currently a Professor for Artificial Intelligence at OTH Amberg-Weiden, where he teaches and conducts research in deep learning and applied machine learning. In addition, he serves as CEO and co-founder of OrasTEC GmbH, a company specializing in AI-driven software solutions. His research is dedicated to the development of advanced deep learning methodologies, with a particular focus on architecturally and conceptually novel algorithmic frameworks tailored to a diverse set of data-centric and domain-specific applications.
\\[0.5cm]

\noindent
\textbf{Andreas Maier} (M’05–SM’20) received a M.Sc. in computer science from the Friedrich-Alexander-Universität Erlangen-Nürnberg (FAU), Erlangen, Germany, in 2005, and the Ph.D. degree in computer science from the same institution in 2009. He is currently a Professor and Head of the Pattern Recognition Lab at the University of Erlangen–Nuremberg, where he has been a faculty member since 2015. His research interests include medical imaging, image and audio processing, digital humanities, and interpretable \ac{ml}. Prof. Maier has developed significant tools in medical signal processing, such as the PEAKS online speech intelligibility assessment tool, and has contributed to over 4,000 patient and control subject analyses. He has received numerous awards, including an ERC Synergy Grant for “4D Nanoscope”.\\[0.5cm] % Adds space after each biography

\noindent
\textbf{Johann Jäger} received the Dipl.-Ing. and Dr.-Ing. degrees in electrical engineering from Friedrich-Alexander-Universität Erlangen-Nürnberg (FAU), Erlangen, Germany, in 1990 and 1996, respectively. From 1996, he was with Power System Planning Department, Siemens, Erlangen, Germany. He was working on different fields of FACTS devices, network planning and protection systems in worldwide projects. Since 2004, he is in charge of a Full Professorship for Electrical Power Systems with the FAU University Erlangen-Nürnberg. He is author of numerous publications and books as well as inventor of several patents.\\[0.5cm]

\noindent  
\textbf{Siming Bayer} received her Dr.-Ing. in Computer Science from Friedrich-Alexander-Universität Erlangen-Nürnberg (FAU) in 2022, where she developed \ac{ml} algorithms for vascular structure registration in collaboration with Siemens Healthineers. She also holds an M.Sc. in Biomedical Engineering from FAU. 
She is currently a Research Scientist at FAU, leading the ''Data Processing for Utility Infrastructure'' group at the Pattern Recognition Lab, and a Strategic Research \& Collaboration Manager at Siemens Healthineers, focusing on cross-business research in digital health. Her research interests include \ac{ml} for medical imaging, industrial data processing, and innovation management in healthcare and utility infrastructures.\\[0.5cm]
\endbio


\appendix
\renewcommand{\thetable}{S.\arabic{table}}
\setcounter{table}{0} % Reset the table counter for the appendix

\newpage
\section{Supplementary information}

\subsection{Review Tables}
\input{tables/tab_fault_detection}
\input{tables/tab_islanding_detection}
\input{tables/tab_fault_classification}
\input{tables/tab_fault_location}

\newpage
% PRISMA-SCR CHECKLIST 
\subsection{PRISMA-ScR Checklist}\label{sec:checklist}
\input{tables/tab_checklist}

% TABLE DATABASE SEARCH QUERIES
\subsection{Database search queries}\label{sec:queries}
\input{tables/tab_database_query}

\end{document}

%% file: acronyms.tex
% --- Machine Learning Models & Methods ---
\newacro{ann}[ANN]{artificial neural network}
\newacro{anfis}[ANFIS]{adaptive neuro-fuzzy inference system}
\newacro{bclr}[BCLR]{binary class logistic regression}
\newacro{bml}[BML]{Boltzmann machine learning}
\newacro{cnn}[CNN]{convolutional neural network}
\newacro{ddpg}[DDPG]{deep deterministic policy gradient}
\newacro{dl}[DL]{deep learning}
\newacro{dnn}[DNN]{deep neural network}
\newacro{dt}[DT]{decision tree}
\newacro{ebt}[EBT]{extremely randomized trees}
\newacro{ebtm}[EBTM]{Ensemble Bagged Trees Model}
\newacro{elm}[ELM]{extreme learning machine}
\newacro{edgevit}[EdgeViT]{Edge Vision Transformer}
\newacro{ffnn}[FFNN]{feedforward neural network}
\newacro{gbt}[GBT]{Gradient Boosting Trees}
\newacro{gmdh}[GMDH]{Group Method of Data Handling}
\newacro{gnn}[GNN]{graph neural network}
\newacro{gru}[GRU]{gated recurrent unit}
\newacro{knn}[KNN]{k-nearest neighbor}
\newacro{kol}[KOL]{known operator learning}
\newacro{lstm}[LSTM]{Long Short-Term Memory}
\newacro{ml}[ML]{machine learning}
\newacro{mlelm}[ML-ELM]{multi-layer extreme learning machine}
\newacro{mlp}[MLP]{multilayer perceptron}
\newacro{mlpnn}[MLPNN]{multilayer perceptron neural network}
\newacro{nb}[NB]{Naive Bayes}
\newacro{nc}[NC]{nearest centroid}
\newacro{nn}[NN]{neural network}
\newacro{pinn}[PINN]{Physics-Informed Neural Network}
\newacro{qda}[QDA]{quadratic discriminant analysis}
\newacro{rbfnn}[RBFNN]{radial basis function neural network}
\newacro{rf}[RF]{random forest}
\newacro{rl}[RL]{reinforcement learning}
\newacro{rnn}[RNN]{recurrent neural network}
\newacro{rusboost}[RUSBoost]{random under-sampling boosting}
\newacro{sae}[SAE]{stacked autoencoder}
\newacro{sarsa}[SARSA]{State-Action-Reward-State-Action}
\newacro{svm}[SVM]{support vector machine}
\newacro{s-dnn}[S-DNN]{Statistical Feature-based Deep Neural Network}
\newacro{tst}[TST]{Time Series Transformer}
\newacro{vit}[ViT]{Vision Transformer}

% --- Signal Processing & Feature Extraction ---
\newacro{cwt}[CWT]{continuous wavelet transform}
\newacro{dost}[DOST]{discrete Orthonormal Stockwell transform}
\newacro{dwt}[DWT]{discrete wavelet transform}
\newacro{ewt}[EWT]{empirical wavelet transform}
\newacro{fdst}[FDST]{fast discrete S-transform}
\newacro{fft}[FFT]{Fast Fourier Transform}
\newacro{gaf}[GAF]{Gramian angular field}
\newacro{mtf}[MTF]{Markov transition field}
\newacro{rp}[RP]{recurrence plot}
\newacro{wt}[WT]{wavelet transform}

% --- Electrical & Power Engineering Terms ---
\newacro{adn}[ADN]{active distribution network}
\newacro{afd}[AFDD]{adaptive fault diagnosis}
\newacro{csc}[CSC]{current source converter}
\newacro{dbpr}[DBPR]{differential-based protection relay}
\newacro{dc}[DC]{direct current}
\newacro{der}[DER]{distributed energy resources}
\newacro{dg}[DG]{distributed generation}
\newacro{dfig}[DFIG]{doubly fed induction generator}
\newacro{dp}[DP]{distance protection}
\newacro{ef}[EF]{earth fault}
\newacro{fc}[FC]{fault classification}
\newacro{fd}[FD]{fault detection}
\newacro{fl}[FL]{fault localization}
\newacro{fli}[FLI]{fault line identification}
\newacro{hif}[HIF]{high-impedance fault}
\newacro{hil}[HIL]{hardware-in-the-loop}
\newacro{hv}[HV]{high voltage}
\newacro{hvdc}[HVDC]{high-voltage direct current}
\newacro{ibr}[IBR]{inverter-based resources}
\newacro{iidg}[IIDG]{inverter-interfaced distributed generator}
\newacro{loe}[LOE]{loss of excitation}
\newacro{lv}[LV]{low voltage}
\newacro{mg}[MG]{microgrid}
\newacro{mmc}[MMC]{modular multilevel converter}
\newacro{mt-hvdc}[MT-HVDC]{multi-terminal high-voltage direct current}
\newacro{mtdc}[MTDC]{multi-terminal direct current}
\newacro{mv}[MV]{medium voltage}
\newacro{ndz}[NDZ]{non-detection zone}
\newacro{pll}[PLL]{phase-locked loop}
\newacro{pmu}[PMU]{phasor measurement unit}
\newacro{psc}[PSC]{power swing condition}
\newacro{pv}[PV]{photovoltaic generators}
\newacro{res}[RES]{renewable energy sources}
\newacro{rtds}[RTDS]{real-time digital simulator}
\newacro{sc}[SC]{short circuit}
\newacro{stvs}[STVS]{short-term voltage stability}
\newacro{vmd}[VMD]{variational mode decomposition}
\newacro{vsc}[VSC]{voltage source converter}
\newacro{wams}[WAMS]{wide-area measurement systems}

% --- General / Optimization / Other ---
\newacro{ai}[AI]{artificial intelligence}
\newacro{bo}[BO]{Bayesian optimization}
\newacro{ee}[EE]{electrical engineering}
\newacro{emo}[EMO]{efficient model}
\newacro{pca}[PCA]{principal component analysis}
\newacro{prisma-scr}[PRISMA-ScR]{PRISMA for Scoping Reviews}
\newacro{pso}[PSO]{particle swarm optimization}

%% file: tables/tab_abbreviations.tex
\begin{table}[htbp]
\centering
\caption{List of Abbreviations}
\label{tab:abbreviations}
\begin{tabular}{ll @{\hskip 1.5cm} ll}
\toprule
\textbf{Abbreviation} & \textbf{Definition} & \textbf{Abbreviation} & \textbf{Definition} \\
\midrule
\multicolumn{4}{l}{\textbf{General / Methods}} \\
\acs{ai}   & \Acl{ai}   & \acs{ml}   & \Acl{ml} \\
\acs{dl}   & \Acl{dl}   & \acs{pca}  & \Acl{pca} \\
\midrule
\multicolumn{4}{l}{\textbf{Signal Processing}} \\
\acs{fft}  & \Acl{fft}  & \acs{wt}   & \Acl{wt} \\
\acs{cwt}  & \Acl{cwt}  & \acs{dwt}  & \Acl{dwt} \\
\midrule
\multicolumn{4}{l}{\textbf{Power \& Energy Systems}} \\
\acs{der}  & \Acl{der}  & \acs{res}  & \Acl{res} \\
\acs{mg}   & \Acl{mg}   & \acs{hv}   & \Acl{hv} \\
\acs{lv}   & \Acl{lv}   & \acs{mv}   & \Acl{mv} \\
\acs{hvdc} & \Acl{hvdc} & \acs{mtdc} & \Acl{mtdc} \\
\acs{mmc}  & \Acl{mmc}  & \acs{vsc}  & \Acl{vsc} \\
\acs{pmu}  & \Acl{pmu}  & \acs{wams} & \Acl{wams} \\
\midrule
\multicolumn{4}{l}{\textbf{Protection Tasks}} \\
\acs{fd}   & \Acl{fd}   & \acs{fc}   & \Acl{fc} \\
\acs{fl}   & \Acl{fl}   & \acs{fli}  & \Acl{fli} \\
\acs{dp}   & \Acl{dp}   & \acs{sc}   & \Acl{sc} \\
\bottomrule
\end{tabular}
\end{table}

%% file: sections/fault_detection.tex
\Acl{fd} is a fundamental component of power system protection, serving as the initial step in most protection schemes to promptly detect abnormal conditions such as short circuits or equipment failures. As power systems evolve -- with growing integration of renewable energy, digital substations, and data-driven architectures -- the scope and implementation of \ac{fd} are also advancing. This section reviews its functional role, conventional methods, and recent research trends, with a focus on \ac{ml}-based approaches.

\subsection{Conventional Approaches and Emerging Challenges}
\label{sec:fd_conventional_challenges}

Traditionally, \ac{fd} has been performed locally within substations using protective relays, often relying soley on current transformers~\cite{blackburn_protective_2014}. These hardwired relays operate autonomously to isolate faults in real time. With digital substations and IEC~61850~\cite{international_electrotechnical_commission_iec_2024, hunt_substation_2019}, measurement and protection logic are decoupled, enabling synchronized data to be processed externally~\cite{aminifar_synchrophasor_2014}. As a result, \ac{fd} can now occur at various architectural levels: via edge devices~\cite{huo_research_2020}, local substation processors~\cite{hunt_substation_2019}, or centralized \ac{wams}~\cite{fan_wide_2019}, allowing \ac{ml}-based methods to be adapted to different deployment scenarios.

While \ac{fd} remains essential for promptly detecting conditions like short circuits or equipment failures, traditional schemes face growing limitations. They must operate quasi-instantaneously~\cite{chen_electrical_2005}, yet rely on fixed thresholds (e.g., over-current, impedance, differential relays) originally designed for high fault currents~\cite{protection_and_automation_b5_protection_2015}. In modern grids with inverter-based \ac{dg}, fault currents are often lower and less predictable~\cite{nsaif_fault_2021}, reducing detection reliability, especially under high-impedance faults or dynamic topologies~\cite{vaish_machine_2021}. Renewable-dominated systems further complicate protection due to bidirectional power flows and varying short-circuit levels~\cite{chen_electrical_2005}, often leading to incorrect or missed operations. Moreover, fixed-scenario designs limit adaptability to evolving grid conditions~\cite{protection_and_automation_b5_protection_2015}.

These challenges have spurred interest in data-driven \ac{fd} methods that analyze current and voltage waveforms using signal processing and \ac{ml} techniques~\cite{aleem_methodologies_2015}. Models such as \acp{wt}, \acp{svm}, and \acp{dnn} have shown strong performance under realistic conditions~\cite{nsaif_fault_2021}. This review identified \num{11} studies focusing exclusively on \ac{ml}-based \ac{fd}, covering distribution networks, transmission systems, and hybrid AC/DC configurations. The following sections explore these contributions, beginning with distribution-level approaches.

\subsection{Machine Learning Approaches to Fault Detection}

% Distribution networks
Several studies have proposed \ac{fd} methods for distribution networks, addressing challenges such as \ac{hif}s, variable \ac{dg}, and limited observability. Bakkar et al.~\cite{bakkar_artificial_2022} introduced a dual \ac{nn}-based controller validated on simulation and lab-scale data. Çelik et al.~\cite{celik_deep_2023} used a deep \ac{gmdh}-\ac{ann} for \ac{hif} detection in an \ac{adn}, showing high accuracy on real 20\,kV data. Diefenthaler et al.~\cite{diefenthaler_artificial_2023} trained a \ac{cnn} on RMS current images from a real \ac{mv} grid, while Nieminen et al.~\cite{nieminen_identifying_2023} applied a \ac{cnn} to COMTRADE data for \ac{sc} and \ac{ef} detection. Other approaches include one-class classification for varying \ac{der} scenarios~\cite{lin_one-class_2020} and an \ac{ann} for \ac{fd} in radial \ac{lv} systems~\cite{moloi_power_2021}. Kordowich et al.~\cite{kordowich_hybrid_2022} proposed a deep Q-learning-based scheme that combines centralized and decentralized agents for adaptive detection, while Wu et al. introduced a nested \ac{rl} framework for relay coordination~\cite{wu_deep_2022}.

While fewer studies address transmission line \ac{fd}, the demand for ultra-fast and accurate detection in \ac{hv} systems remains critical. Chen et al.~\cite{chen_electrical_2005} emphasized the need for sub-cycle detection to prevent cascading failures. Kumar et al.~\cite{kumar_improved_2024} developed a hybrid \ac{svm}-\ac{cnn} model trained on real voltage signals from the ENET Centre. 

Hybrid AC/DC grids pose additional challenges due to power electronics and diverse fault signatures. Yousaf et al.~\cite{yousaf_novel_2023} used a feedforward \ac{ann} combined with multi-resolution analysis for \ac{fd} in a \qty{320}{\kilo\volt}, four-terminal \ac{mt-hvdc} system, achieving high accuracy. Galkin et al.~\cite{galkin_toolset_2023} applied a \ac{cnn} to identify sympathetic inrush phenomena in transformer data generated in Simulink, aiming to prevent false tripping by distinguishing non-fault transient from actual faults. Rezaei et al.~\cite{rezaei_novel_2020} proposed a \ac{fd} algorithm for \ac{dfig}-based wind farms under varying operating conditions.

The reviewed \ac{fd} approaches exhibit diverse methodological and experimental designs. Most rely on simulated data -- via tools like PowerFactory, Simulink, and OpenDSS -- typically using IEEE test systems or real grid models. Some incorporate real measurements, such as voltage recordings from the ENET Centre~\cite{kumar_improved_2024} or COMTRADE files from the Finnish grid~\cite{nieminen_identifying_2023}, enhancing realism. Others combine simulation with lab validation or hardware-in-the-loop testing~\cite{bakkar_artificial_2022,rezaei_novel_2020}.

Dataset sizes vary widely, from fewer than \num{200} real events~\cite{nieminen_identifying_2023} to over \num{800000} high-frequency samples~\cite{kumar_improved_2024}. Several studies assess robustness under varying \ac{dg} levels, noise, fault types, and topologies. Many emphasize fast response times (e.g., sub-\qty{20}{\milli\second}) and deployment aspects such as selective tripping, communication delays, and false-trip resilience, reflecting growing system maturity.

The studies span distribution and transmission networks, transformers, wind farms, and DC grids, applying a broad range of \ac{ml} methods. Common techniques include \acp{cnn}, \acp{dnn}, and \acp{svm}, along with hybrid models like \ac{svm}-\ac{cnn} and \ac{rl}-\ac{lstm} frameworks~\cite{wu_deep_2022}. Several also explore less conventional approaches, such as one-class classifiers~\cite{lin_one-class_2020}, teacher-student strategies~\cite{kordowich_hybrid_2022}, and multi-resolution analysis with random search~\cite{yousaf_multisegmented_2023}. While \ac{fd} remains a foundational task, many studies extend their focus to \ac{fc}, aiming to distinguish between different fault types and improve situational awareness.

\subsubsection*{Islanding Detection}
In addition to short circuits, high-impedance faults, and broken conductors, protection and disturbance management systems must also detect inadvertent islanding. Within the context of this review, inadvertend islanding is treated as an abnormal event and thus a critical disturbance to be addressed by protection systems. Inadvertent islanding refers to a condition where a section of a distribution grid becomes electrically isolated from the main supply but remains energized by \acp{der}. Such scenarios are a major concern for distribution system operators, as they can damage equipment and endanger maintenance crews. Therefore, \acp{der} are commonly required to be disconnected rapidly in this case of islanding.

This scoping review identified and compared a substantial body of work on ML-based islanding detection, summarized in Table~\ref{tab:islanding_detection}. While earlier reviews often gave this topic limited attention, the growing penetration of renewable generation has motivated extensive research on fast and reliable detection, with both feature-driven and end-to-end \ac{ml} approaches now widely explored.

Feature-driven methods extract information from voltage and current signals before classification. Examples include harmonic sequence components with \ac{lstm}, intrinsic mode or wavelet decompositions with \ac{ann}/\ac{dnn}, and RMS- or THD-based features with \ac{svm}~\cite{abdelsalam_islanding_2020,admasie_intelligent_2020,baghaee_anti-islanding_2020,baghaee_support_2020,choudhury_islanding_2021,najar_intelligent_2020,nsaif_island_2023}. Such approaches report high accuracy, fast detection (often below 10\,ms), and reduced \acp{ndz}.
End-to-end deep learning has emerged as an alternative, directly processing raw or time-frequency transformed signals via \acp{cnn}, \acp{lstm}, or hybrid models. Reported accurarcies typically exceed 95\,\%, with some studies demonstrating perfect recall under noisy conditions and strong robustness to topology changes~\cite{allan_new_2021,reddy_deep_2021,hussain_intelligent_2022,david_islanding_2024,ozcanli_novel_2022,ozcanli_islanding_2022}.

A further trend addresses deployment and integration. Ali et al.~\cite{ali_hierarchical_2021} embedded an \ac{ann}-based detector into a hierarchical IoT framework, achieving IEEE~1547 compliance. Real-time and hardware-in-the-loop implementations demonstrate feasibility at scale, including classification speeds above 170,000\,obs/s and dependable detection within 5--15\,ms~\cite{ezzat_microgrids_2021,saifi_intelligent_2023}.

A recurring focus across all studies is minimization of the \ac{ndz}, defined as the range of operating conditions where an islanding event occurs but remains undetected by the protection scheme. Some approaches consistently report near-zero \ac{ndz} using \ac{cnn}, \ac{dnn}, or \ac{rnn}/\ac{lstm}-based classifiers, while others achieve sub-$5$~ms detection with \acp{ndz} below $4\%$~\cite{hussain_intelligent_2022,hussain_islanding_2023,hussain_communication-less_2024,nsaif_island_2023}. Overall, \ac{ml}-based methods achieve high accuracy, fast response, and improved robustness, but most rely on simulation data with limited field validation. Future work should address domain adaptation, interoperability, and integration with digital substation architectures to ensure reliable operation under practical grid conditions.

%% file: sections/fault_classification.tex
Faults are abnormal electrical conditions that interrupt the normal operation of power systems. Short circuits, either between phases or from a phase to ground, are especially critical, as the resulting high current flow can cause thermal and mechanical stress, potentially damaging equipment and compromising system stability \cite{lester_high_1974}. Classifying faults accurately is a fundamental step in designing protection schemes and improving response time during disturbances~\cite{hewitson_practical_2005}. In this section, we examine both conventional classification strategies and recent \ac{ml}-based approaches across various grid types and protection contexts.

\subsection{Conventional Approaches and Emerging Challenges}
\label{sec:fc_conventional_challenges}

Power system faults typically fall into three categories: symmetrical, unsymmetrical, and open-circuit. Symmetrical faults, such as three-phase short circuits, are rare but severe, affecting all phases equally. Unsymmetrical faults -- including single line-to-ground (L-G), line-to-line (L-L), and double line-to-ground (L-L-G) -- occur more frequently and cause phase imbalances. Open-circuit faults involve conductor disconnection, leading to unbalanced currents and equipment stress. Each fault type exhibits distinct electrical characteristics, requiring tailored detection and mitigation strategies.

Conventional protective relays implicitly distinguish faults through specialized designs rather than explicit classification. Distance relays measure impedance trajectories to identify fault locations and indirectly infer fault types. Differential relays detect internal faults via current imbalances, while overcurrent relays trip based on exceeded current thresholds~\cite{blackburn_protective_2014, horowitz_power_2023}. Classical analytical methods like symmetrical components and traveling wave analysis also support fault discrimination but lack explicit classification capability~\cite{wilches-bernal_survey_2021}.

Recent \ac{ml} methods explicitly classify faults using waveform patterns captured from high-frequency signals. Techniques such as \acp{cnn}, \acp{rnn} (e.g., \acsp{lstm}, \acsp{gru}), and ensemble methods directly assign fault labels, while accounting for diverse grid topologies and noisy conditions~\cite{vaish_machine_2021, hassan_fault_2024, alilouch_intelligent_2022}. Preprocessing techniques like wavelet transforms further improve feature extraction and model accuracy.

However, \ac{ml}-based classification introduces significant challenges. Models typically rely on simulated data, limiting generalizability to real-world operational conditions. Strict real-time constraints impose trade-offs between model complexity and inference latency~\cite{abder_elandaloussi_practical_2023}. Additionally, the black-box nature of \ac{ml} models complicates interpretability, trust, and deployment in safety-critical environments. Ultimately, explicit \ac{fc} via \ac{ml} offers promising avenues toward adaptive, proactive, and intelligent protection strategies, positioning it as an essential component of future grid reliability and resilience.

\subsection{Machine Learning Approaches to Fault Classification}

Recent studies have applied \ac{ml} to \ac{fd} and \ac{fc} in transmission systems using simulation tools like Simulink, PSCAD, and PSS/E. Classifiers range from \acp{ann}, \acp{dt}, and \ac{svm} to probabilistic and ensemble models. Many works use single-line Simulink models with methods such as \ac{lstm}, double-channel extreme learning machines, or \ac{qda}, focusing on time-domain features from voltage and current signals~\cite{devi_detection_2023, haq_improved_2021, dhingra_machine_2023, kumar_deep_2022,fahim_self_2020}.

More complex grid scenarios include IEEE 39-bus simulations to benchmark classifiers such as \ac{mlelm}, stacked autoencoders, \ac{mlp}, \ac{knn}, \ac{svm}, and \ac{nb}~\cite{harish_comparative_2023, massaoudi_short-term_2023}. Studies targeting TCSC-compensated lines use one-cycle post-fault data or fast S-transform features with ensemble classifiers~\cite{mohanty_decision_2020, ramana_intelligent_2023}. Uddin et al.~\cite{uddin_protection_2022} evaluate \ac{svm}-based detection on an IEC benchmark under varying conditions. Fahim et al.~\cite{fahim_self_2020} present a self-attention \ac{cnn} with time-series imaging and wavelet preprocessing for robust \ac{fd} and \ac{fc}. 

Other contributions include Moloi et al.~\cite{moloi_wavelet-neural_2020}, who combine \ac{dwt} features with a \ac{pso}-optimized \ac{ann}, and Muntasir et al.~\cite{muntasir_fast_2023}, who compare multiple classifiers on data from a real \qty{50}{\mega\watt} plant. Myint et al.~\cite{myint_artificial_2024} use traveling-wave features and a Bayesian-trained \ac{ann} for \ac{fc} and phase identification in traction systems. Guama et al.~\cite{guama_simple_2021} uniquely incorporate real COMTRADE data from a \qty{230}{\kilo\volt} substation, combined with PowerFactory simulations. Their two-stage self-organizing map-based model uses zero-sequence components to detect ground faults and modal currents for phase identification, generalizing well to unseen events. Han et al.~\cite{han_faulted-phase_2021} propose a gradient similarity-based classifier for faulted-phase identification in high-voltage grids, outperforming ANN and component-based approaches under noisy and high-impedance conditions.

While most \ac{fc} studies focus on AC transmission systems, emerging \ac{hvdc} and hybrid HVAC/\ac{hvdc} grids introduce new challenges. Reviewed works span two- and four-terminal, bipolar, and \ac{mmc}-based \ac{hvdc} systems with voltages from \qty{100}{\kilo\volt} to \qty{800}{\kilo\volt}, simulated primarily in Simulink or EMTDC, with some incorporating real data or \ac{rtds}-based validation. Topologies include \ac{vsc}, line-commutated converter, multi-terminal meshes, and IEEE 39-bus extensions.

Several studies employ deep learning models on either raw or image-transformed waveforms. Li et al.~\cite{li_data-driven_2024} use a Con-Att-BiLSTM with adversarial training on \num{110000} alarm records from a \(\pm\)\qty{800}{\kilo\volt} U\ac{hvdc} system, while Liang et al.~\cite{liang_current_2023} apply a CA-\ac{cnn} to current trajectory images in hybrid grids. Others use signal-derived features for classification, including time- or frequency-domain attributes processed by \acp{ann}, \ac{bml}, or logistic regression models~\cite{mehdi_squaring_2023, merlin_frequency_2022, muzzammel_low_2022}.

Multi-stage and ensemble approaches are also common. Pragati et al.~\cite{pragati_bayesian_2023, pragati_fault_2024} use Bayesian-optimized decision trees, Stockwell transforms, GRU-\acp{rnn}, and Teager–Kaiser operators in sequential \ac{fd}/\ac{fc} pipelines. Comparative evaluations of \ac{lstm}, \ac{rf}, and \ac{ann} models using voltage-difference features are provided by Swetapadma et al.~\cite{swetapadma_novel_2022}, while Tsotsopoulou et al.~\cite{tsotsopoulou_protection_2023} propose a communication-free XGBoost classifier using only local data.
Time-frequency analysis remains central in the work of Wong, Xiang, and Yousaf. Wong et al.~\cite{wong_hybrid_2024} extract traveling wave features for binary \ac{fd}/\ac{fc}, Xiang et al.~\cite{xiang_ann-based_2020} use one-level \ac{dwt} with a 5-class \ac{ann}, and Yousaf et al.~\cite{yousaf_novel_2023, yousaf_deep_2023} combine \ac{dwt}-based features with \ac{bo}-optimized \ac{ann} and later propose a communication-independent \ac{lstm} for \ac{fc} in a four-terminal \qty{320}{\kilo\volt} \ac{mt-hvdc} mesh.

Together, these studies highlight a rich landscape of \ac{ml}-based protection in \ac{hvdc} and hybrid grids, combining deep learning, signal-driven features, and optimized classifiers, with growing emphasis on real-time performance and minimal communication needs.

% \subsection{Machine Learning Approaches for Fault Classification in Distribution, Generation, Substations, and Microgrids}
% Balan, Wang. Transformer = 2  

Recent studies have extended \ac{ml}-based \ac{fc} to distribution networks, substations, and generation systems. Balan et al.~\cite{balan_detection_2023} use supervised learning on three years of real-world IoT transformer data -- including voltage, current, power, oil level, and temperature -- for fault diagnosis without simulation. Wang et al.~\cite{wang_detection_2024} apply decision trees to empirical substation data to detect abnormal topologies in three-phase transformers.

In distribution networks, simulated studies explore tasks such as islanding detection in \ac{pv}-integrated grids~\cite{buscariolli_new_2023}, \ac{hif} detection in radial feeders~\cite{moloi_high_2022}, and microgrid protection with \ac{iidg}~\cite{chhetija_fault_2024}. Preprocessing includes \ac{dwt}~\cite{moloi_high_2022}, modified Gabor wavelets~\cite{wang_use_2020}, and transient-voltage features~\cite{chhetija_fault_2024}. Classifiers span \ac{svm}~\cite{moloi_high_2022,jones_machine_2021}, ensemble trees~\cite{chhetija_fault_2024}, and compact \acp{cnn}~\cite{wang_use_2020}, trained on signals sampled at \qtyrange{6.4}{20}{\kilo\hertz}.

\Ac{ml} has also been applied to \ac{fc} in power generation systems. Ashtiani et al.~\cite{ashtiani_supervised_2024} develop a current-only relay for wind turbines using time-domain features and \ac{dt} models. Ramadoss et al.~\cite{ramadoss_machine_2023} detect excitation failure in synchronous generators using statistical indicators and binary \ac{svm} classification of SMIB data. Sadiq et al.~\cite{sadiq_machine_2024} classify faults in Zeta converters within DC microgrids using multi-class models under simulated degradation.

%  Alsaba, Aryan, Cano, Mana, Ojetola, Patnaik, Prasad Tiwari, Kabeel. Microgrid = 8 

Microgrid-focused studies primarily rely on MATLAB/Simulink simulations, with occasional use of EMTDC. Aryan et al.~\cite{aryan_fault_2022} employ an \ac{rbfnn} trained on simulated RMS signals for \ac{fd} and \ac{fc}. A few works incorporate real fault recordings, including a \qty{750}{\volt} DC microgrid at ETL-KAFB~\cite{ojetola_testing_2022} and additional data from Kirtland AFB~\cite{patnaik_modwt-xgboost_2021}. Common test systems include IEEE 5-, 9-, and 13-bus networks, the CIGRÉ low-voltage benchmark, and custom topologies spanning \qtyrange{0.35}{13}{\kilo\volt}. Signal decomposition techniques such as \ac{dwt} and multi-resolution analysis are widely used for feature extraction~\cite{mana_detection_2023, cano_integrating_2024}, while classifiers range from conventional models (\ac{svm}, \ac{dt}, \ac{nb}) to advanced architectures, including \acp{cnn}, hybrid \ac{cnn}-GRU~\cite{alsaba_efficient_2023}, and both centralized and communication-free \ac{ann}-based schemes~\cite{prasad_tiwari_enhancing_2021, kabeel_centralized_2022}. Some studies assess generalization by training on simulations and testing on real data~\cite{ojetola_testing_2022}.

Across all domains -- transmission, hybrid HVAC/\ac{hvdc} systems, distribution networks, generation units, and microgrids -- \ac{ml}-based \ac{fc} methods exhibit substantial diversity in system configurations, data sources, and classifier designs. While most rely on simulation tools like MATLAB/Simulink and PSCAD, several incorporate real-world measurements from substations, converters, and utility grids. Feature extraction spans time, frequency, and time-frequency domains, using techniques such as \ac{dwt}, Stockwell transforms, and energy-based operators. Classifiers range from \ac{svm}, \ac{dt}, and \ac{nb} to deep architectures like \ac{cnn}, \ac{lstm}, and hybrid ensembles. Recent trends emphasize robustness to noise, communication independence, cross-grid generalization, and real-time feasibility-reflecting a growing maturity of \ac{ml}-based protection adapted to the complexity and decentralization of modern power systems.

Given this growing complexity, conventional \ac{fl} methods often struggle with speed, scalability, and resilience under real-world constraints. As a result, recent research has shifted toward \ac{ml}-based \ac{fl}, leveraging data-driven models built on electrical signals, topology, and fault history to achieve faster, more accurate, and more adaptive localization-particularly in operational contexts where traditional techniques fall short.

%% file: sections/fault_localization.tex
Fault localization identifies the precise point along a power line where a disturbance has occurred (e.g., at 12.5\% of the line length), enabling selective tripping and faster service restoration. Conventional methods rely on distance relays that divide lines into zones, triggering immediate response in the primary zone and delayed actions in others to preserve coordination.

However, the rise of \acp{ibr}, which supply limited fault current, reduces the effectiveness of impedance-based techniques. In response, high-resolution measurements and modern communication protocols have enabled more adaptable, data-driven and \ac{ml}-based \ac{fl} methods. This section reviews conventional approaches, their limitations, and recent advances in \ac{ml}-based \ac{fl}.

\subsection{Conventional Approaches and Emerging Challenges}
\label{sec:fl_conventional_challenges}

\Ac{fl} is a core function in power system protection, enabling selective tripping during faults and supporting faster post-event diagnosis. Accurate localization reduces outage durations, particularly in complex or remote networks, but its effectiveness depends on grid topology and data availability.

In distribution systems, non-uniform line parameters, lateral taps, and mixed conductor types hinder precision~\cite{blackburn_protective_2014}. Still, detailed topology and impedance data can often narrow fault zones sufficiently. Transmission networks, with their more uniform structure and enhanced observability facilitated by a higher number of measurements, are better suited for automated \ac{fl}, especially where manual inspection is costly or slow.

Recent standards like IEC~61850~\cite{international_electrotechnical_commission_iec_2024} and high-speed communication protocols~\cite{suhail_hussain_novel_2016} have expanded \ac{fl}'s applicability by enabling time-synchronized, system-wide monitoring~\cite{blackburn_protective_2014}.

Conventional \ac{fl} methods are typically classified into impedance-based and traveling-wave-based approaches. Impedance-based techniques are computationally efficient and easy to implement, but rely on simplifying assumptions that often do not hold in real-world conditions. Traveling-wave methods, using high-frequency transients and GPS-synchronized measurements, offer higher accuracy, especially in high-impedance or multi-line scenarios~\cite{lopes_fault_2014,jensen_online_2014,lin_universal_2012,gazzana_integrated_2014,dashti_fault_2014,morales_identification_2014}.

\Ac{wams} further improves \ac{fl} by enabling system-wide data integration~\cite{brahma_fault_2004,jia_new_2013,izykowski_accurate_2010,elkalashy_unsynchronized_2016}, but traditional methods still face challenges such as blind zones, low observability, and hardware failures~\cite{elhaffar_optimized_2012,korkali_traveling-wave-based_2012,korkali_optimal_2013}.

Emerging tasks like \ac{fli}~\cite{oelhaf_systematic_2025} and zonal fault classification~\cite{uddin_hybrid_2022} require scalable, data-rich algorithms capable of handling heterogeneous conditions. These were previously constrained by centralized architectures and limited communication but are now increasingly viable due to decentralized designs and fast data exchange. In this context, \ac{ml}-based approaches are gaining traction. The next section reviews recent contributions that leverage signal-derived and topological features to improve fault classification where conventional methods fall short.

\subsection{Machine Learning Approaches to Fault Localization}

As discussed in Sec.~\ref{sec:fl_conventional_challenges}, the higher measurement density and well-defined protection zones of transmission networks make them well-suited for \ac{fl}, especially at \ac{hv} and UHV levels, where rapid and accurate localization is critical for maintaining system stability, despite potentially complex topologies. Of the \num{43} reviewed studies on \ac{fl}, \num{19} focus on transmission-level scenarios. However, only one includes hardware-in-the-loop validation, underscoring the persistent gap between simulation and real-world application.
The following examples highlight selected studies that illustrate recent advances in \ac{fl} using \ac{ml}, particularly in terms of accuracy, integration with legacy systems, and noise robustness.

Tran~\cite{tran_integration_2023} proposed a hybrid scheme combining conventional distance protection with an \ac{ml}-based correction layer. An \ac{mlp}, trained on \num{2136} simulated samples from a \qty{118.5}{\kilo\meter} AC-185/29 line in Vietnam, processed \qty{60}{\milli\second} windows of wavelet, raw, and frequency-domain features. Replay via an Omicron CMC-356 and Siemens 7SA522 relay demonstrated that the \ac{mlp} reduced average \ac{fl} error from \qty{0.92}{\percent} to \qty{0.42}{\percent}, improving localization by \qty{0.6}{\kilo\meter} and showcasing the value of \ac{ml}-augmented legacy systems.

Raj et al.~\cite{n_transmission_2021} developed a lightweight \ac{mlp} model for \ac{fd}, \ac{fc}, and \ac{fl} on a \qty{132}{\kilo\volt}, \qty{100}{\kilo\meter} line using RMS voltage and current features from \ac{fft}. While \ac{fd} and \ac{fl} performance was strong, \ac{fc} lagged-highlighting the trade-offs in simplified architectures.

Najafzadeh et al.~\cite{najafzadeh_fault_2024} applied WHO-optimized \ac{rf}, \ac{dt}, and adaptive neuro-fuzzy inference system (ANFIS) models with fuzzy thresholding on high-resolution \ac{pmu} data under low SNR conditions. Their ensemble achieved sub-\qty{200}{\meter} localization error, demonstrating strong noise resilience and precise tuning for \ac{hv} environments.

Together, these studies reflect the flexibility of \ac{ml}-based \ac{fl} in transmission systems-ranging from real-world hardware integration to lightweight simulation models and noise-robust ensembles.

Beyond targeted contributions, a broader body of work explores \ac{fl} across diverse system configurations. Studies range from large-scale IEEE 68- and 145-bus networks using \ac{wams} data and recurrent models~\cite{afrasiabi_fault_2022}, to single-line \qty{400}{\kilo\volt}, \qty{100}{\kilo\meter} simulations with hybrid \ac{cnn}-\ac{lstm} architectures~\cite{alilouch_intelligent_2022}. Conventional approaches based on \acp{ann} and \ac{rf} remain common~\cite{alrashdan_artificial_2023, bera_hybrid_2024}, while lightweight models like \ac{elm}~\cite{cholla_end--end_2023} target resource-constrained environments. Several studies address practical issues such as incomplete relay coverage~\cite{meyer_system-based_2020}, limited measurement availability~\cite{n_transmission_2021}, and integration of synchronized phasor data~\cite{najafzadeh_fault_2024}, reflecting a growing focus on robustness and real-world feasibility. Hassan et al.~\cite{hassan_fault_2024} apply an optimizable ELM for \ac{fc} and \ac{fl} on IEEE 9-bus simulation data.

A subset of work addresses \ac{fl} in wind-integrated systems, where variable generation introduces additional uncertainty.
Prabhakar~\cite{prabhakar_distance_2023} uses a hybrid \ac{svm}-\ac{ann} on a modified IEEE 9-bus system with wind, while Uddin et al.~\cite{uddin_hybrid_2022} classify fault zones in \ac{dfig}-based farms using \ac{dt}, \ac{svm}, and \ac{pca}.
Taheri et al.~\cite{taheri_fault-location_2021, taheri_novel_2022} propose \ac{lstm}-based models for \ac{fl} and impedance estimation, demonstrating robustness to noise and high fault resistance. Rizk-Allah et al.~\cite{rizk-allah_dynamic_2023} improve accuracy by tuning neural network hyperparameters using \ac{pso}, while Garika et al.~\cite{garika_condition_2022} apply bior1.5 wavelets and \acp{ann} in IoT-enabled wind–\ac{svm} systems. 

While these studies demonstrate the adaptability of \ac{ml}-based \ac{fl}, most rely on simulated data, small or non-standardized datasets, and lack field validation-limiting comparability and practical deployment. These limitations are even more pronounced in \ac{hvdc} systems, where fault behavior differs fundamentally and research remains in its early stages.

%\subsubsection*{Fault Localization in HVDC-Systems}
% hvdc: Farshad, Gnanamalar, Imani, Muzzammel, Muzzammel, Ren, Tiwari, Tong 
\ac{ml}-based \ac{fl} in \ac{hvdc} systems is gaining traction due to the fast DC fault dynamics and the limitations of conventional methods in handling transient behavior and directional ambiguity. Recent studies cover point-to-point and multi-terminal \ac{csc}-, \ac{vsc}-, and \ac{mmc}-based topologies, typically operating between \qty{100}{\kilo\volt} and \qty{500}{\kilo\volt}. Yousaf et al.~\cite{yousaf_intelligent_2022} addressed \ac{fl} in a four-terminal \ac{mmc}-\ac{hvdc} grid using \ac{dwt}-based features and a Bayesian-optimized \ac{ffnn}, trained under varying fault resistances and noise.

Most studies use EMTDC or Simulink for simulation and apply different \ac{ml} models for \ac{fd}, \ac{fc}, and \ac{fl}. Classical methods rely on wavelet or statistical features-such as \ac{pca}, maximal overlap discrete \ac{wt}, or moving averages-combined with \ac{svm} or general regression \ac{nn} classifiers~\cite{muzzammel_support_2020, muzzammel_wavelet_2023, imani_novel_2024, farshad_intelligent_2021}. More recent work explores hybrid and deep learning models, including semi-\ac{ai} and \ac{cnn}-\ac{svm} setups, achieving detection under \qty{5}{\milli\second} and handling high-resistance faults~\cite{tong_semi_2022, gnanamalar_cnnsvm_2023}.

Tiwari~\cite{tiwari_deep_2023} extended \ac{cnn}-based \ac{fl} to wind-integrated \ac{hvdc} systems, outperforming standard methods across different fault cases. Ren et al.~\cite{ren_fault_2024} proposed a vision transformer (\ac{vit}) model that turns voltage signals into image-like formats, \ac{gaf}, \ac{rp}, \ac{mtf}, \ac{cwt}, and processes them using small \ac{vit} networks. Trained on EMTDC data from a four-terminal \qty{500}{\kilo\volt} grid, the model reached a mean error of \qty{0.441}{\kilo\meter} and stayed accurate even with noise, high resistance, and low sampling rates.

Together, these studies show increasing variety in methods and a stronger focus on reliable performance for \ac{ml}-based \ac{fl} in \ac{hvdc} grids. As the field develops, similar work is now appearing in distribution and microgrid systems, where irregular layouts and limited measurements create new challenges.

% \subsubsection*{Fault Localization in Distribution Grids and Microgrids}
% Microgrids: Jayasinghe, Kumari, Miraj, Ponukumati, Qusayer, Srivastava, Mampilly
% Distribution:Rong, Zhang, Kandasamy
% Generaton: Alhanaf, Zhang

Limited measurements and communication infrastructure in distribution networks and microgrids complicate \ac{fl}, therefore \ac{ml}-based methods are increasingly explored. Convolutional models are common, including hybrid \ac{cnn}-\ac{lstm} for ring networks with variable \ac{dg}, and pure \ac{cnn} classifiers for compact grids~\cite{alhanaf_fault_2024, rong_convolutional_2023}. Some works incorporate \ac{dg} effects using \ac{ann}, 1D-\ac{cnn}, or \ac{vmd}-\ac{cnn} models~\cite{alhanaf_intelligent_2023, zhang_fault_2022}. Kandasamy et al.~\cite{kandasamy_intelligent_2024} used a \ac{dost}-based deep model for \ac{fd}, \ac{fc}, and \ac{fl}, while Zhang et al.~\cite{zhang_fault_2023} proposed a spatio-temporal protection scheme using multi-agent reinforcement learning. Sapountzoglou et al.~\cite{sapountzoglou_fault_2020} use gradient boosting trees for \ac{fd}, \ac{fc}, and \ac{fl} in a simulated low-voltage smart grid.

In microgrids, characterized by low voltage and frequent mode transitions, most studies rely on Simulink simulations with \ac{dg} and \ac{res}. Jayasinghe et al.~\cite{jayasinghe_classification_2024} and Kumari et al.~\cite{kumari_enhancing_2024} used \ac{ann}-based models with strong \ac{fc} and \ac{fl} results. \Ac{ann}s remain dominant, used in four of six reviewed studies~\cite{jayasinghe_classification_2024, kumari_enhancing_2024, ponukumati_pattern_2021, qusayer_communication_2024}, while alternatives such as \ac{svm}~\cite{miraj_improved_2023}, and ensembles methods like \ac{rf}~\cite{srivastava_random_2023}, which are based on \acp{dt}, offer gains in interpretability and training speed.

Mampilly and Sheeba~\cite{mampilly_empirical_2024} combined \ac{ewt}-based decomposition with a hybrid \ac{cnn}-bi\ac{lstm} model optimized via pelican algorithm in an IEEE 13-bus \ac{res} microgrid. Goli et al.~\cite{goli_neuro-wavelet_2022} used wavelet-\ac{ann} methods for \ac{fd} and \ac{fl} in a hybrid \ac{pv}-wind setup.

Feature engineering techniques-such as \ac{dwt}~\cite{ponukumati_pattern_2021} or task-specific filtering~\cite{srivastava_random_2023}-are commonly used to improve model robustness. Despite promising results, many studies lack detail on key dataset characteristics. Sampling frequency (\( f_s \)), which critically affects signal resolution, is often unreported. Sample sizes range from under \num{500} to over \num{2.4} million~\cite{kumari_enhancing_2024, ponukumati_pattern_2021}, with sampling rates from \qty{1}{\kilo\hertz} to \qty{260}{\kilo\hertz}. Simulated topologies span transmission lines, distribution feeders, multi-terminal \ac{hvdc} systems, and microgrids.

Some works move toward broader validation-evaluating cross-platform consistency between Simulink and EMTDC~\cite{jayasinghe_classification_2024} or incorporating IEC~61850-based communication models~\cite{qusayer_communication_2024}, but overall reporting remains inconsistent. This hampers the assessment of realism and reproducibility, especially for \ac{fl}, which depends on subtle spatial and temporal patterns. Moreover, many studies report only aggregate accuracy, omitting localization-specific metrics such as mean absolute or distance error.

While microgrid-focused work demonstrates the feasibility of \ac{ml}-based \ac{fl} under constrained conditions, progress across grid types would benefit from more standardized reporting, clearer evaluation protocols, and stronger links between simulation and real-world deployment.

%% file: tables/tab_reporting_standards.tex
\begin{table}[tpb]
\centering
\caption{Minimum Reporting Standards for ML-Based Protection Studies}
\label{tab:reporting_standards}
\begin{tabularx}{\textwidth}{@{}lXl@{}}
\toprule
\textbf{Attribute} & \textbf{Description} & \textbf{Example} \\
\midrule
Voltage level and grid type & Specify whether the data pertains to MV, HV, or \ac{hvdc} networks. & \num{90}\,kV AC-transmission line \\
Topology & Include number of buses and, if available, a single-line diagram. & IEEE 39-bus system \\
Simulation tool and version & Clearly state the tool used and its version. & PSCAD v4.6.1 \\
Sampling frequency and signal domain & Report sampling rate and signal domain (e.g., EMT, RMS/phasor)
& \num{1.4}\,kHz, EMT domain \\
Time window and alignment & Define the time range and its alignment with the fault event. & \qty{-100}{ms} to \qty{300}{ms} \\
Input channels used & List all input signals used in the study. & $V_{abc}$, $I_{abc}$, frequency \\
Fault specifications & Describe the fault types, locations, and resistances. & SLG @ Bus 3, $R_f = \num{10}\,\Omega$ \\
\bottomrule
\end{tabularx}
\end{table}

%% file: tables/tab_public_power_system_benchmark_datasets.tex
\begin{table}[tbp]
\centering
\caption{Essential Aspects and Examples for Public Power System Benchmark Datasets}
\label{tab:benchmark_recommendations}
\begin{tabularx}{\textwidth}{@{}lXX@{}}
\toprule
\textbf{Aspect} & \textbf{Recommendation} & \textbf{Examples} \\
\midrule
Topology diversity &
Include multiple grid types and voltage levels to support generalization. &
IEEE 13-bus (MV, radial); IEEE 123-bus (LV, urban); CIGRÉ \ac{hvdc} test system (HVDC) \\

Fault scenario coverage &
Simulate varied fault types, resistances, and locations to represent realistic fault conditions. &
SLG, LLG, 3LG faults at buses and lines; $R_f = 0$--$100\,\Omega$; multiple inception angles \\

Operating condition variation &
Reflect daily and seasonal load changes, generation profiles, and DER penetration, and typical non-fault events. &
Light, nominal, peak load; PV and wind integration; islanding events; inrush currents; line switching; load steps \\

Signal representation &
Provide raw or derived signal domains for accurate modeling and analysis. &
EMT waveforms at 10\,kHz; phasors at 50\,Hz; $V_{abc}$, $I_{abc}$, frequency, ROCOF \\

Ground-truth annotation &
Label each instance for reproducibility and evaluation of downstream tasks. &
Fault type = LLG; location = Bus 4; impedance = 20\,$\Omega$; event start = 2.3\,s \\

Documentation and accessibility &
Provide structured metadata, licensing, and version control to enable long-term reuse. &
JSON/YAML config files; Simulink/PSCAD inputs; Zenodo DOI; CC-BY license; GitHub repository \\
\bottomrule
\end{tabularx}
\end{table}

%% file: tables/tab_fault_detection.tex
\begin{longtable}{@{}p{2.4cm}p{14.4cm}@{}}
\caption{Comparison of Recently Proposed ML-Based Approaches for Fault Detection.}
\label{tab:fault_detection} \\
\toprule
\textbf{Author} & \textbf{Description} \\
\midrule
\endfirsthead

\toprule
\textbf{Author} & \textbf{Description} \\
\midrule
\endhead

\bottomrule
\endfoot

Bakkar et al.\newline
\cite{bakkar_artificial_2022}\newline
(2022)
& \textbf{Focus:} \ac{ai}-based protection for distribution systems with high \ac{mg} penetration
\newline
\textbf{Data:} Simulation in PowerFactory and lab data from a scaled smart grid; \num{5000} fault cases with varying \ac{dg}, fault types, and resistances; \(\mathrm{f}_\mathrm{s}\) unspecified; signal type: phasor domain \(V\)/\(I\)
\newline
\textbf{Method:} Two \ac{ann}-based controllers (central and backup), trained with Levenberg–Marquardt algorithm
\newline
\textbf{Result:} Fast tripping \qty{10}\,ms; robust to varying \ac{dg} penetration, fault types, fault locations, transformer configurations, and fault resistances; tolerant to communication issues
\\ \addlinespace

Çelik et al.
\newline
\cite{celik_deep_2023}
\newline
(2023)
& \textbf{Focus:} \ac{fd} under disturbances and \ac{hif}s in \ac{adn}s
\newline
\textbf{Data:} Simulation in PowerFactory and Simulink, based on a real \qty{20}\,kV network; \num{1200} samples; \(\mathrm{f}_\mathrm{s}\) unspecified; signal type: phasor domain \(I\)
\newline
\textbf{Method:} Deep \ac{gmdh} \ac{ann}
\newline
\textbf{Result:} \ac{fd} within \qty{20}\,ms; MAPE of \(3.51\%\)
\\ \addlinespace

Diefenthaler et al.\newline
\cite{diefenthaler_artificial_2023}\newline
(2023)
& \textbf{Focus:} \ac{hif} detection in \ac{mv} \ac{dg}s using image-based \ac{ml}
\newline
\textbf{Data:} Simulink simulation of a real \ac{mv} grid in the Basque Country; \num{100} labeled RMS current images; \(\mathrm{f}_\mathrm{s}\) unspecified; signal type: phasor domain \(V\)/\(I\)
\newline
\textbf{Method:} \ac{cnn} on RMS current images
\newline
\textbf{Result:} \qty{96}\,\% accuracy
\\ \addlinespace

Galkin et al.\newline \cite{galkin_toolset_2023} \newline
(2023)
& \textbf{Focus:} \ac{fd} of transformer excitation phenomena
\newline
\textbf{Data:} Simulink model of a real substation in Älvkarleby, Sweden; \num{1280} samples at \qty{2}\,kHz; \(\mathrm{f}_\mathrm{s}\) unspecified; signal type: waveform \(V\)/\(I\)
\newline
\textbf{Method:} \ac{cnn}
\newline
\textbf{Result:} \qty{92}\,\% precision
\\ \addlinespace

Kordowich et al.\newline
\cite{kordowich_hybrid_2022} 
\newline
(2022)
& \textbf{Focus:} \ac{fd} protection for transmission lines using centralized and decentralized agents
\newline
\textbf{Data:} \num{30000} simulated short circuits in PowerFactory; \(\mathrm{f}_\mathrm{s}\) unspecified; signal type: phasor domain \(V\)/\(I\)
\newline
\textbf{Method:} Deep recurrent Q-learning with \ac{gru}-based \ac{ann} and teacher-student training
\newline
\textbf{Result:} \qty{100}\,\% fault clearance; selective tripping with delays between \qty{0.03}\,s and \qty{0.1}\,s, \qty{0}\,\% false trips by student agents
\\ \addlinespace

Kumar et al.\newline
\cite{kumar_improved_2024} \newline
(2024)
& \textbf{Focus:} \ac{fd} for transmission lines
\newline
\textbf{Data:} Real voltage signals of \num{800000} samples over \qty{20}\,ms from \qty{50}\,Hz power lines, labeled for partial discharge detection; \(\mathrm{f}_\mathrm{s} = \qty{40}\,\text{MHz}\); simulation software unspecified; signal type: waveform \(V\)/\(I\)
\newline
\textbf{Method:} \ac{svm}-\ac{cnn}
\newline
\textbf{Result:} Accuracy of \qty{95.7}\,\%
\\ \addlinespace

Lin et al.\newline
\cite{lin_one-class_2020} 
\newline
(2020)
& \textbf{Focus:} \ac{fd} in distribution systems with varying penetration levels of \ac{der}
\newline
\textbf{Data:} Simulation using GridLAB-D on IEEE 123-bus system; \num{297500} samples; \(\mathrm{f}_\mathrm{s}\) unspecified; signal type: phasor domain \(I\), \(P\), \(Q\)
\newline
\textbf{Method:} SVDD, iSVDD, HISVDD
\newline
\textbf{Result:} SVDD reduces mal-operations to \qty{1.2}\,\% with \qty{0}\,\% miss-trips under high \ac{der} penetration
\\ \addlinespace

Moloi et al.\newline
\cite{moloi_power_2021} \newline
(2021)
& \textbf{Focus:} \ac{fd} in radial \ac{lv} distribution systems
\newline
\textbf{Data:} Simulink simulations on 3-phase, 16-bus radial distribution network; sample size \& \(\mathrm{f}_\mathrm{s}\) unspecified; signal type: waveform \(I\)
\newline
\textbf{Method:} \ac{ann} trained on statistical features extracted from fault current signals
\newline
\textbf{Result:} \ac{fc} accuracy \qty{97.1}\,\%; faulted phase identification accuracy \qty{95.8}\,\%
\\ \addlinespace

Nieminen et al.\newline
\cite{nieminen_identifying_2023}\newline
(2023)
& \textbf{Focus:} \ac{fd} of \ac{sc} and \ac{ef} events in distribution networks
\newline
\textbf{Data:} \qty{181} real COMTRADE event recordings (73 normal, 89 \ac{ef}, 19 \ac{sc}) from the Finnish grid; simulation software \& \(\mathrm{f}_\mathrm{s}\) unspecified; signal type: waveform \(I\)
\newline
\textbf{Method:} \ac{cnn}
\newline
\textbf{Result:} \qty{100}\,\% accuracy for \ac{sc} and \qty{97.9}\,\% for \ac{ef}
\\ \addlinespace

Rezaei et al.\newline
\cite{rezaei_novel_2020}\newline
(2020)
& \textbf{Focus:} \Ac{dbpr} scheme enhanced by \ac{ml} to protect the intertie zone of large-scale centralized \ac{dfig}-based wind farms
\newline
\textbf{Data:} Simulink simulation and real-time testing on FPGA; internal/external faults under varying wind conditions; sample size \& \(\mathrm{f}_\mathrm{s}\) unspecified; signal type: phasor domain \(V\)/\(I\)
\newline
\textbf{Method:} \ac{svm} classifier integrated with \ac{dbpr}
\newline
\textbf{Result:} Accuracy of \qty{97.5}\,\%; no false trips during external faults or wind fluctuations; validated in simulation and FPGA tests
\\ \addlinespace

Wu et al.\newline
\cite{wu_deep_2022} 
\newline
(2022)
& \textbf{Focus:} \ac{fd} and protective relay coordination in radial distribution system
\newline
\textbf{Data:} Simulation of the IEEE 34-bus test system using OpenDSS; \num{5000} samples; \(\mathrm{f}_\mathrm{s}\) unspecified; signal type: phasor domain \(V\)/\(I\)
\newline
\textbf{Method:} Nested \ac{rl} framework enhanced with \ac{lstm}
\newline
\textbf{Result:} \ac{rl}-based relays showed \qty{0.38}\,\% missed detections and \qty{1.28}\,\% coordination failures.
\\ \addlinespace

Yousaf et al. \newline
\cite{yousaf_multisegmented_2023}
\newline
(2023)
& \textbf{Focus:} Single-ended \ac{fd} in \ac{mt-hvdc} grids 
\newline \textbf{Data:} Simulations of \num{3477} DC transmission line faults under varying fault impedances; simulation software \& \(\mathrm{f}_\mathrm{s}\) unspecified; signal type: DC \(V\)/\(I\)
\newline \textbf{Method:} Stationary wavelet transform for feature extraction; Random Search-optimized \ac{ann}
\newline \textbf{Result:} Accuracy \qty{98.13}\,\% (\qtyrange{20}{45}\,dB noise); minimum trip time \qty{0.9}\,ms
\\ \addlinespace

\end{longtable}

%% file: tables/tab_islanding_detection.tex
\begin{longtable}{@{}p{2.4cm}p{14.4cm}@{}}
\caption{Comparison of Recently Proposed ML-Based Approaches for Islanding Detection.} \label{tab:islanding_detection} \\
\toprule
\textbf{Author} & \textbf{Description} \\
\midrule
\endfirsthead

\toprule
\textbf{Author} & \textbf{Description} \\
\midrule
\endhead

\bottomrule
\endfoot

Abdelsalam et al.\newline
\cite{abdelsalam_islanding_2020}\newline
(2020)
& \textbf{Focus:} Islanding detection in inverter-based \acp{mg}
\newline
\textbf{Data:} Simulink simulations on a custom \qty{0.4}{kV} \ac{mg} with inverter-based wind \acp{dg}; laboratory validation on wind- and PV-based \ac{dg} testbed; number of samples unspecified; \(f_s\) unspecified; signal type: waveform \(V\)/\(I\)
\newline
\textbf{Method:} DFT-based extraction of 2nd-harmonic seq. comps. (V/I: pos., neg., zero); classification via \ac{lstm}
\newline
\textbf{Result:} Accuracy 99.5\,\%; dependability 99.3\,\%; avg. detection time 6\,ms
\\ \addlinespace

Admasie et al.\newline
\cite{admasie_intelligent_2020}\newline
(2020)
& \textbf{Focus:} Islanding detection in grid-connected \acp{mg} with multiple \acp{dg}
\newline
\textbf{Data:} Simulink simulations on a custom \qty{25}{kV} \ac{mg} with inverter- and synchronous-based \acp{dg}; sample size not specified; \(f_s=3.84\,\text{kHz}\); signal type: waveform \(V\)
\newline
\textbf{Method:} IMF-based statistical feature extraction; classification via \ac{ann}
\newline
\textbf{Result:} Accuracy 99.5\,\%, precision 99.5\,\%
\\ \addlinespace

Ali et al.\newline
\cite{ali_hierarchical_2021}\newline
(2021)
& \textbf{Focus:} Islanding detection in hierarchical \acp{mg} with IoT-based cloud \ac{ml}
\newline
\textbf{Data:} Simulink on modified IEEE-13 bus \ac{mg} (4.16\,kV) with four inverter-based \acp{dg}; real-time lab validation; samples and \(f_s\) not specified; signal type: phasor domain \(V\)/\(I\)
\newline
\textbf{Method:} \ac{ann} classifier integrated into 3-layer IoT control framework
\newline
\textbf{Result:} Detection time <2\,s; NDZ = none; IEEE 1547 compliant; THD 2.36\,\% (with IoT)
\\ \addlinespace

Allan and Morsi\newline
\cite{allan_new_2021}\newline
(2021)
& \textbf{Focus:} Passive islanding detection in PV-based grid-connected systems
\newline
\textbf{Data:} Simulink simulations on a custom \qty{120}{kV} PV system; \num{5770} samples; \(f_s=3.84\,\text{kHz}\); signal type: waveform \(V\)/\(I\)
\newline
\textbf{Method:} \ac{cwt}-based time–frequency representation; classification via \ac{cnn}
\newline
\textbf{Result:} Accuracy 97.8–98.6\,\%; precision 95\,\%; recall 99\,\%; F1-score 97\,\%
\\ \addlinespace

Baghaee et al.\newline
\cite{baghaee_anti-islanding_2020}\newline
(2020)
& \textbf{Focus:} Anti-islanding protection of PV-\acp{mg} with PHEV charging
\newline
\textbf{Data:} Simulink model of a \qty{0.4}{kV}, \qty{20}{kW} PV plant (Bosnia); experimental PCC data via IED; sample size not specified; 8-sample (40\,ms) stacks; signal type: phasor domain \(V\)/\(I\)
\newline
\textbf{Method:} \ac{svm} with Gaussian kernel; 5\,ms FIFO sampling; discrimination of islanding vs. PHEV charging; vs. other AI methods
\newline
\textbf{Result:} Accuracy 100\,\%; DT $\approx$40\,ms; NDZ $\approx$9.5\,\%; UL1741 compliant
\\ \addlinespace

Baghaee et al.\newline
\cite{baghaee_support_2020}\newline
(2020)
& \textbf{Focus:} Islanding and grid fault detection in PV-based active DNs
\newline
\textbf{Data:} Simulink with real PV plant parameters; sample size not specified; \(f_s=1\,\text{kHz}\); signal type: waveform \(V\)
\newline
\textbf{Method:} \ac{svm} with kernel dimension grouping; discrimination of islanding vs. grid faults; vs. passive/active methods
\newline
\textbf{Result:} Accuracy 100\,\%; reduced NDZ and DT; robust against grid faults
\\ \addlinespace

Choudhury et al.\newline
\cite{choudhury_islanding_2021}\newline
(2021)
& \textbf{Focus:} Islanding detection of PV-based \acp{mg}
\newline
\textbf{Data:} Simulink simulations on a PV-based \ac{mg} (0.5\,kV); \num{1035} samples; \(f_s\) unspecified; signal type: waveform \(V\)
\newline
\textbf{Method:} Empirical mode decomposition for feature extraction; classification via \ac{dt}
\newline
\textbf{Result:} Accuracy 97.4\,\% (noiseless); 91\,\% (SNR = 20\,dB)
\\ \addlinespace

David et al.\newline
\cite{david_islanding_2024}\newline
(2024)
& \textbf{Focus:} End-to-end islanding detection in PV-based \acp{mg} under noisy conditions
\newline
\textbf{Data:} Simulink simulations on a modified IEEE 13-bus \ac{mg}; \num{2000} voltage samples; \(f_s=200\,\text{kHz}\); signal type: waveform \(V\)
\newline
\textbf{Method:} \ac{tst}, \ac{lstm}, and InceptionTime models
\newline
\textbf{Result:} Accuracy 98.3\,\%; precision 96.6\,\%; recall 100\,\%
\\ \addlinespace

Ezzat et al.\newline
\cite{ezzat_microgrids_2021}\newline
(2021)
& \textbf{Focus:} Islanding detection in \acp{mg}
\newline
\textbf{Data:} Simulink with lab validation on a custom \qty{0.4}{kV} \ac{mg} including PV/wind \acp{dg}; events: breaker opening, 3P/1P faults, P/Q mismatch, simultaneous; sample size not specified; \(f_s=6.4\,\text{kHz}\); signal type: waveform \(V\)/\(I\)
\newline
\textbf{Method:} Fourier transform features; \ac{knn}; vs. DT, SVM, LR, ensemble
\newline
\textbf{Result:} Accuracy 99.2–99.8\,\% (20–40\,dB noise); dependability 100\,\%; security 99.7\,\%; detection time 5–15\,ms
\\ \addlinespace

Hussain et al.\newline
\cite{hussain_intelligent_2022}\newline
(2022)
& \textbf{Focus:} Islanding detection in multi-\ac{dg} systems with zero NDZ
\newline
\textbf{Data:} Simulink on IEC 61850-7-420 system (\qty{25}{kV}); \num{3159} samples; \(f_s=3.84\,\text{kHz}\); events: islanding + non-islanding (faults/switching); signal type: waveform \(V\)
\newline
\textbf{Method:} STFT features; deep \ac{cnn}; 5-fold CV
\newline
\textbf{Result:} Accuracy 100\,\% (noiseless), 97–99.5\,\% (30–40\,dB); zero NDZ
\\ \addlinespace

Hussain et al.\newline
\cite{hussain_islanding_2023}\newline
(2023)
& \textbf{Focus:} Islanding detection and non-islanding disturbance classification in multi-\ac{dg} systems
\newline
\textbf{Data:} Simulink on IEC 61850-7-420 system; \num{1740} samples; \(f_s=3.84\,\text{kHz}\); events: islanding, faults, load/capacitor switching; ; signal type: waveform \(V\)/\(I\)
\newline
\textbf{Method:} \ac{s-dnn} with statistical features from PCC V; vs. DT, MLP, SVM, \ac{lstm}
\newline
\textbf{Result:} Accuracy 99.8\,\%; precision 99.7\,\%; recall 100\,\%; F1 99.9\,\%; robust at 30–40\,dB
\\ \addlinespace

Hussain et al.\newline
\cite{hussain_communication-less_2024}\newline
(2024)
& \textbf{Focus:} Communication-less islanding detection in hybrid \ac{dg} systems
\newline
\textbf{Data:} Simulink on IEC 61850-7-420 hybrid system with sync. + inverter \acp{dg}; \num{3159} samples; events: islanding, faults, load/capacitor switching; signal type: waveform \(V\)
\newline
\textbf{Method:} Time-domain features (13) from PCC V; wrapper selection; \ac{rnn}/\ac{lstm}; vs. ANN, SVM
\newline
\textbf{Result:} Accuracy 99.7\,\%; F1 99.7\,\%; selectivity 99.4\,\%; sensitivity 100\,\%; robust at 30–40\,dB; DT 72\,ms
\\ \addlinespace

Najar et al.\newline
\cite{najar_intelligent_2020}\newline
(2020)
& \textbf{Focus:} Islanding detection in grid-connected \acp{mg} with PV, wind, and load variability
\newline
\textbf{Data:} PowerFactory simulations on a custom \ac{mg}; \num{11264} samples; \(f_s=3.2\,\text{kHz}\); signal type: \(V\)/\(I\) sequence components
\newline
\textbf{Method:} \ac{dwt}-based feature extraction; classification via \ac{dnn} (TensorFlow)
\newline
\textbf{Result:} Accuracy 99.9\,\%, dependability 99.6\,\%
\\ \addlinespace

Nsaif et al.\newline
\cite{nsaif_island_2023}\newline
(2023)
& \textbf{Focus:} Islanding detection for PV-based \acp{dg} in distribution grids
\newline
\textbf{Data:} Simulink simulations on a modified Malaysian distribution grid; \num{2048} samples; \(f_s=400\,\text{kHz}\); signal type: \(V\) sequence components + power signals at PCC
\newline
\textbf{Method:} Variational mode decomposition for IMF extraction; classification via Ensemble Bagged Trees Model
\newline
\textbf{Result:} Accuracy 100\,\%; NDZ <4\,\%; detection time 4.8\,ms
\\ \addlinespace

{\"O}zcanli and Baysal\newline
\cite{ozcanli_novel_2022}\newline
(2022)
& \textbf{Focus:} Islanding detection in modified \acp{mg}
\newline
\textbf{Data:} Simulink simulations on modified AEP/CERTS \ac{mg}; \num{2800} samples; \(f_s=1\,\text{kHz}\); signal type: \(V\)/\(I\) harmonic distortion at PCC
\newline
\textbf{Method:} Multi-\ac{lstm} classifier; compared with other intelligent IDMs
\newline
\textbf{Result:} Accuracy 99.3\,\% (avg.); 97.9\,\% for ±0.5\,\% power mismatch; detection time 50\,ms
\\ \addlinespace

{\"O}zcanli and Baysal\newline
\cite{ozcanli_islanding_2022}\newline
(2022)
& \textbf{Focus:} Islanding detection in modified \acp{mg}
\newline
\textbf{Data:} Simulink simulations on modified CERTS \ac{mg}; \num{4000} cases; \(f_s=1\,\text{kHz}\); signal type: \(V\)/\(I\) harmonics at PCC
\newline
\textbf{Method:} 1D-\ac{cnn} and hybrid \ac{cnn}-\ac{lstm}; compared with 2D-\ac{cnn}
\newline
\textbf{Result:} Accuracy up to 100\,\% (avg. 99.7\,\%); robust to low NDZ and topology changes
\\ \addlinespace

Reddy et al.\newline
\cite{reddy_deep_2021}\newline
(2021)
& \textbf{Focus:} Islanding detection in integrated \ac{dg} systems
\newline
\textbf{Data:} Simulink simulations on an \ac{dg} system; \num{258} samples; \(f_s\) unspecified; signal type: waveform \(V\)/\(I\)
\newline
\textbf{Method:} CWT-based scalogram generation; classification via \ac{cnn}; compared with ANN and fuzzy logic
\newline
\textbf{Result:} Accuracy 95.4\,\%
\\ \addlinespace

Saifi et al.\newline
\cite{saifi_intelligent_2023}\newline
(2023)
& \textbf{Focus:} Islanding classification in hybrid \acp{mg}
\newline
\textbf{Data:} Typhoon HIL simulations on a hybrid \ac{mg} (\qty{13.8}{kV}) with PV, wind, diesel, and battery; \num{31251} samples; \(f_s\) unspecified; signal type: waveform \(V\)/\(I\)
\newline
\textbf{Method:} \ac{mlp} classifier trained on islanding and grid fault scenarios
\newline
\textbf{Result:} Accuracy 99.5\,\%; classification speed 170k obs/s
\\ \addlinespace

\end{longtable}

%% file: tables/tab_fault_classification.tex
\begin{longtable}{@{}p{2.4cm}p{14.4cm}@{}}
\caption{Comparison of Recently Proposed ML-Based Approaches for Fault Classification}
\label{tab:fault_classification} \\
\toprule
\textbf{Author} & \textbf{Description} \\
\midrule
\endfirsthead

\toprule
\textbf{Author} & \textbf{Description} \\
\midrule
\endhead

\bottomrule
\endfoot

Alsaba et al.\newline
\cite{alsaba_efficient_2023}
\newline
(2022)
& \textbf{Focus:} \ac{fd} and \ac{fc} in \qty{20}\,kV \ac{mg}
\newline
\textbf{Data:} Simulink simulation on CIGRE 14-bus; \num{644} samples at \qty{1.6}\,kHz; signal type: waveform \(I\)
\newline
\textbf{Method:} \ac{cnn}-\ac{gru}
\newline
\textbf{Result:} Accuracy of \qty{87.72}\,\%
\\ \addlinespace

Aryan et al.\newline
\cite{aryan_fault_2022}
\newline
(2022)
& \textbf{Focus:} \ac{fd} and \ac{fc} in a simulated \ac{mg}
\newline \textbf{Data:} Simulink simulation of RMS signals from \qty{4.16}\,kV \ac{mg}; sample size \& \(\mathrm{f}_\mathrm{s}\) unspecified; signal type: phasor domain \(V\)/\(I\)
\newline
\textbf{Method:} \ac{rbfnn}
\newline
\textbf{Result:} Accuracy of \qty{82}\,\%
\\ \addlinespace

Ashtiani et al.\newline
\cite{ashtiani_supervised_2024}
\newline
(2024)
& \textbf{Focus:} \ac{fd} and \ac{fc} for current-only \ac{ml}-based relays
\newline
\textbf{Data:} Simulink simulation of wind turbine distributed generators; \num{1200} samples; \(\mathrm{f}_\mathrm{s}\) unspecified; signal type: waveform \(I\)
\newline
\textbf{Method:} \ac{knn}, \ac{rf}, \ac{dt}
\newline
\textbf{Result:} Accuracy of \qty{93}\,\% (\ac{knn}), \qty{92}\,\% (\ac{rf}), \qty{90}\,\% (\ac{dt})
\\ \addlinespace

Balan et al.\newline
\cite{balan_detection_2023}
\newline
(2023)
& \textbf{Focus:} \ac{fd} and fault diagnosis in transformers using sensor data and \ac{ml}
\newline
\textbf{Data:} \num{4381} real-world samples collected via IoT sensors at \qty{15}\,min intervals; simulation software \& \(\mathrm{f}_\mathrm{s}\) unspecified; signal type: phasor domain \(V\)/\(I\)
\newline
\textbf{Method:} \ac{dt}, \ac{rf}, XGBoost, AdaBoost, Logistic Regression, \ac{knn}, \ac{svm}
\newline
\textbf{Result:} Best performance: \qty{98.79}\,\% (DT), \qty{98.36}\,\% (RF); others >\qty{94}\,\%
\\ \addlinespace

Buscariolli et al.\newline
\cite{buscariolli_new_2023}
\newline
(2023)
& \textbf{Focus:} Islanding detection in distribution systems with \ac{pv}-based \ac{dg}
\newline
\textbf{Data:} Simulink simulation of UFABC network; \num{583} scenarios; sampling rate up to \qty{128}\,\text{samples/cycle}; signal type: waveform \(V\)/\(I\)
\newline
\textbf{Method:} \Ac{mlp}, \ac{ann}
\newline
\textbf{Result:} \qty{100}\,\% accuracy; detection time between \qty{0.06}\,s and \qty{0.08}\,s
\\ \addlinespace

Cano et al.\newline
\cite{cano_integrating_2024} \newline
(2024)
& \textbf{Focus:} \ac{fd} and \ac{fc} in renewable \ac{mg}s with PV, HKT, and variable loads under varying fault resistance
\newline
\textbf{Data:} Simulink simulations on IEEE-5 bus microgrid testbed; \num{144} fault cases; \( \mathrm{f}_\mathrm{s} \approx \qty{2.5}\,\text{Hz} \); signal type: waveform \(V\)/\(I\)
\newline
\textbf{Method:} \ac{dwt} feature extraction; classification via \ac{rbfnn}; compared with general regression \ac{nn}, probabilistic neural network, \ac{ffnn}, \ac{rnn}, nonlinear autoregressive with external input, \ac{cnn}, \ac{svm}, and adaptive neuro-fuzzy inference system
\newline
\textbf{Result:} \ac{rbfnn} achieved MSE \(= 1.31\times10^{-31}\), \(R^2 = 0.99998\); trained in \qty{1.63}\,s, converged in 100 epochs; robust across 12 fault types and ground resistances up to \qty{100}\,ohm
\\ \addlinespace

Chhetija et al.\newline
\cite{chhetija_fault_2024}
\newline
(2024)
& \textbf{Focus:} \ac{fd} and \ac{fc} in islanded microgrids with \ac{iidg}s, using local voltage-based features
\newline \textbf{Data:} Simulink simulation of modified CIGRE LV microgrid; \num{16038} cases, with fault variations across load levels, locations, and inception angles; \(\mathrm{f}_\mathrm{s}\); signal type: waveform \(V\)/\(I\)
\newline \textbf{Method:} Modified voltage-based transient magnitude factor and \ac{ebt} classifier using fixed window
\newline \textbf{Result:} Accuracy \qty{99.5}\,\%; response time \qty{10}\,ms
\\ \addlinespace

Devi et al.\newline
\cite{devi_detection_2023}
\newline
(2023)
& \textbf{Focus:} \ac{fd} and \ac{fc} in transmission lines
\newline
\textbf{Data:} Simulink simulation of a \qty{55}\,Km short transmission line; \num{2400} samples; \(\mathrm{f}_\mathrm{s}\) unspecified; signal type: waveform \(I\)
\newline
\textbf{Method:} \ac{lstm}
\newline
\textbf{Result:} \ac{fd} accuracy of \qty{99.58}\,\%, \ac{fc} accuracy of \qty{94.17}\,\%
\\ \addlinespace

Dhingra et al.\newline
\cite{dhingra_machine_2023}
\newline
(2023)
& \textbf{Focus:} \ac{fd} and \ac{fc} in transmission lines
\newline
\textbf{Data:} Simulink simulation of a power system with 4 generators, transformers, and faults at the midpoint of the transmission line; sample size \& \(\mathrm{f}_\mathrm{s}\) unspecified; signal type: phasor domain \(V\)/\(I\)
\newline
\textbf{Method:} \ac{qda}
\newline
\textbf{Result:} \ac{fd} accuracy of \qty{99.15}\,\%, AUC score of \qty{99.95}\,\%
\\ \addlinespace

Fahim et al.\newline
\cite{fahim_self_2020}\newline
(2020)
& \textbf{Focus:} \ac{fd} and \ac{fc} for HV transmission lines
\newline
\textbf{Data:} Simulink simulation on IEEE 39-bus system; \num{16000} samples; \( \mathrm{f}_\mathrm{s} = \qty{10}\,\text{kHz}\); signal type: waveform \(V\)/\(I\)
\newline
\textbf{Method:} Self-attention \ac{cnn} using GAF-based time-series imaging and DWT preprocessing
\newline
\textbf{Result:} Nearly \qty{100}\,\% accuracy across all fault types
\\ \addlinespace

Guama et al. \newline
\cite{guama_simple_2021}
\newline
(2021)
& \textbf{Focus:} \ac{fc} for transmission line protection
\newline
\textbf{Data:} \num{8000} samples from PowerFactory and real \qty{230}\,kV substation faults; \(\mathrm{f}_\mathrm{s}\) unspecified; signal type: phasor domain \(V\)/\(I\)
\newline
\textbf{Method:} Two-stage Self-Organizing Map
\newline
\textbf{Result:} Accuracy \qty{99}\,\% (stage 1), \qty{96}\,\% (stage 2)
\\ \addlinespace

Han et al.\newline
\cite{han_faulted-phase_2021}
\newline
(2021)
& \textbf{Focus:} Faulted-phase classification in HV transmission systems
\newline
\textbf{Data:} PSCAD simulation on IEEE 39-bus system; \( \mathrm{f}_\mathrm{s} = \qty{10}\,\text{kHz}\); sample size unspecified; signal type: phasor domain \(V\)/\(I\)
\newline
\textbf{Method:} Gradient similarity-based classification
\newline
\textbf{Result:} Outperforms ANN and symmetrical component-based methods under noise and high-impedance faults
\\ \addlinespace

Haq et al. \newline
\cite{haq_improved_2021}
\newline
(2021)
& \textbf{Focus:} \ac{fd} and \ac{fc} for transmission line protection
\newline
\textbf{Data:} Simulink simulation with \num{31680} three-phase current signals; custom topology; \(\mathrm{f}_\mathrm{s} = \qty{20}\,\text{kHz}\); signal type: waveform \(V\)/\(I\)
\newline
\textbf{Method:} Double-channel Extreme Learning Machine
\newline
\textbf{Result:} Accuracy >\qty{99}\,\%; \ac{fd} time \qty{0.12}\,s
\\ \addlinespace

Harish et al. \newline
\cite{harish_comparative_2023}
\newline
(2023)
& \textbf{Focus:} \ac{fd} and \ac{fc} for transmission line protection
\newline
\textbf{Data:} PowerWorld simulation of IEEE-39 bus (\qty{345}\,kV); \num{24654} samples; \(\mathrm{f}_\mathrm{s}\) unspecified; signal type: phasor domain \(V\)/\(I\)
\newline
\textbf{Method:} \Ac{sae}, \ac{mlelm}
\newline
\textbf{Result:} ML-ELM: \qty{100}\,\% \ac{fd}, \qty{99.92}\,\% \ac{fc} in \qty{29}\,ms; SAE: \qty{100}\,\% \ac{fd}, \qty{97.99}\,\% \ac{fc} in \qty{91}\,ms
\\ \addlinespace

Jones et al. \newline
\cite{jones_machine_2021}
\newline
(2021)
& \textbf{Focus:} \ac{fc} and \ac{fl} in distribution network relays
\newline
\textbf{Data:} Simulink simulation on IEEE 123-bus system; \num{20000} samples; \qty{20}\,kHz sampling rate; signal type: unspecified \(V\)/\(I\)
\newline
\textbf{Method:} \ac{svm}
\newline
\textbf{Result:} Accuracy \qty{100}\,\% (\ac{fc}), \qty{89}\,\% (location)
\\ \addlinespace

Kabeel et al.\newline
\cite{kabeel_centralized_2022}
\newline
(2022)
& \textbf{Focus:} Centralized protection of \acp{mg} using \ac{fd} and \ac{fli}
\newline
\textbf{Data:} Simulink simulation of IEEE 9-bus system; \num{3e6} samples across MG modes and topologies; \(\mathrm{f}_\mathrm{s}\) unspecified; signal type: phasor domain \(V\)/\(I\)
\newline
\textbf{Method:} \ac{ann}
\newline
\textbf{Result:} \qty{99.0}\,\% accuracy, consistent across different \ac{res} levels
\\ \addlinespace

Khan et al.\newline
\cite{khan_machine_2023}
\newline
(2023)
& \textbf{Focus:} ML-based coordination of IDMT overcurrent and earth fault relays
\newline
\textbf{Data:} Synthetic dataset with 425 samples from a 132/33/11\,kV substation scenario; 7 features incl. fault type, current, voltage, CT ratio, pickup setting; \(\mathrm{f}_\mathrm{s}\) unspecified; signal type: phasor domain \(V\)/\(I\)
\newline
\textbf{Method:} Regression models (LR, DT, RF, SVR, GB); Gradient Boosting best
\newline
\textbf{Result:} \(R^2 = 97\%\), CV score: \qty{86.2}\,\% (GB); lowest MAE and MSE
\\ \addlinespace

Kumar et al.\newline
\cite{kumar_deep_2022}
\newline
(2022)
& \textbf{Focus:} \ac{fc} in transmission lines
\newline
\textbf{Data:} Public Kaggle dataset; \num{7861} samples; no simulation platform used \& \ \(\mathrm{f}_\mathrm{s}\) unspecified; signal type: waveform \(V\)/\(I\)
\newline
\textbf{Method:} \ac{cnn}, \ac{ann}
\newline
\textbf{Result:} Accuracy for both models approximately \qty{80}\,\%
\\ \addlinespace

Li et al. \newline
\cite{li_data-driven_2024}
\newline
(2024)
& \textbf{Focus:} Fault analysis and \ac{fd} for UHVDC transmission lines
\newline \textbf{Data:} Dataset of alarm information set of the Sunan ±\qty{800}\,kV UHV project in Jinping; \num{110000} samples; no simulation platform used \& \ \(\mathrm{f}_\mathrm{s}\) unspecified; signal type: alarm text
\newline \textbf{Method:} Con-Att-BiLSTM with adversarial training
\newline \textbf{Result:} F1 score of \qty{96.81}\,\%
\\ \addlinespace

Liang et al.\newline
\cite{liang_current_2023}
\newline
(2023)
& \textbf{Focus:} \ac{fc} in HVAC/\ac{hvdc} systems with \ac{mmc} stations
\newline
\textbf{Data:} PSCAD (IEEE 39-bus) and RTDS data from \qty{100}\,Km \ac{mmc}-\ac{hvdc} line; \num{3272} samples; \(\mathrm{f}_\mathrm{s}\) unspecified; signal type: waveform \(V\)/\(I\)
\newline
\textbf{Method:} CA-\ac{cnn} on current trajectory images with data augmentation
\newline
\textbf{Result:} \qty{100}\,\% accuracy; robust to \qty{0.5}\,ms delay and CT saturation
\\ \addlinespace

Mana et al.\newline
\cite{mana_detection_2023}
\newline
(2023)
& \textbf{Focus:} \ac{fd} and \ac{fc} of SLG faults in unbalanced islanded microgrids
\newline
\textbf{Data:} Simulink simulation of modified IEEE 13-bus microgrid with diesel and inverter-based DERs; \num{499} scenarios; \(\mathrm{f}_\mathrm{s} = \qty{1}\,\text{MHz}\); \(\mathrm{f}_\mathrm{s}\) unspecified; signal type: waveform \(I\)
\newline
\textbf{Method:} Multi-resolution analysis with 8-level \ac{dwt}; feature extraction using Shannon entropy and energy ratio; polynomial-kernel \ac{svm}
\newline
\textbf{Result:} Accuracy of \qty{99}\,\% (combined features); \qty{96}\,\% (entropy only)
\\ \addlinespace

Massaoudi et al.\newline
\cite{massaoudi_short-term_2023}
\newline
(2023)
& \textbf{Focus:} \Ac{stvs} detection and \ac{fc} in transmission networks
\newline
\textbf{Data:} PSSE simulation of IEEE 39-bus system with CLOD model; \num{3753} samples; \(\mathrm{f}_\mathrm{s}\) unspecified; signal type: phasor domain \(V\)/\(P\)/\(Q\)
\newline
\textbf{Method:} Feature selection and optimization with \ac{mlp}, \ac{knn}, \ac{svm}, \ac{nb}
\newline
\textbf{Result:} Accuracy \qty{98.0}\,\% (\ac{stvs}), \qty{91.5}\,\% (\ac{fc}); precision, recall, and \(F1\) reported
\\ \addlinespace

Mehdi et al.\newline
\cite{mehdi_squaring_2023}
\newline
(2023)
& \textbf{Focus:} \ac{fc} of AC faults in hybrid HVAC/\ac{hvdc} transmission systems
\newline
\textbf{Data:} Simulink simulation on \qty{200}\,km hybrid AC/DC two-line system; \num{1210} samples; \(\mathrm{f}_\mathrm{s}\) unspecified; signal type: waveform \(V\)/\(I\)
\newline
\textbf{Method:} Squaring, lowpass filtering, and transient energy extraction; \ac{ann}-based classifier
\newline
\textbf{Result:} Accuracy \qty{99.3}\,\% under noise-free conditions; \qty{93.6}\,\%–\qty{96.9}\,\% across \qtyrange{10}{40}\,dB SNR
\\ \addlinespace

Merlin et al.\newline
\cite{merlin_frequency_2022}
\newline
(2022)
& \textbf{Focus:} \ac{fd} and \ac{fc} in VSC- and CSC-based \ac{hvdc} transmission systems
\newline
\textbf{Data:} Simulink simulation of HVDC network under various fault types and locations; \num{3000} samples; \(\mathrm{f}_\mathrm{s}\) unspecified; signal type: waveform \(V\)/\(I\) and DC \(V\)/\(I\)
\newline
\textbf{Method:} Frequency-domain voltage signal analysis with \ac{ann}-based detection and classification
\newline
\textbf{Result:} \qty{100}\,\% \ac{fd} accuracy in \qty{4.9}\,ms; \qty{97.2}\,\% \ac{fc} accuracy in \qty{9.76}\,ms
\\ \addlinespace

Mohanty et al.\newline
\cite{mohanty_decision_2020}
\newline
(2020)
& \textbf{Focus:} \ac{fd} and \ac{fc} in series-compensated transmission lines
\newline
\textbf{Data:} PSCAD simulation of \qty{400}\,kV, \qty{300}\,Km two-area system with TCSC; \num{3600} samples; \(\mathrm{f}_\mathrm{s} = \qty{1}\,\text{kHz}\); signal type: waveform \(V\)/\(I\)
\newline
\textbf{Method:} \ac{dt} trained on one-cycle post-fault current and voltage
\newline
\textbf{Result:} \ac{fc} accuracy \qtyrange{97.5}{98.7}\,\% across fault types, locations, and resistances
\\ \addlinespace

Moloi et al.\newline
\cite{moloi_wavelet-neural_2020}
\newline
(2020)
& \textbf{Focus:} \ac{fd} and \ac{fc} in transmission lines of a South African utility network
\newline
\textbf{Data:} Simulink simulation of \qty{132}\,kV network with 3 lines and 4 loads; \num{42000} samples; \(\mathrm{f}_\mathrm{s} = \qty{10}\,\text{kHz}\); signal type: waveform s\(I\)
\newline
\textbf{Method:} \ac{dwt}-based energy and entropy features; BP-\ac{ann} optimized with \ac{pso}
\newline
\textbf{Result:} \ac{fc} accuracy \qty{99.0}\,\% (with \ac{pso}), \qty{92.6}\,\% (without); precision and recall \(>\qty{96}\,\%\) for all types of faults
\\ \addlinespace

Moloi and Davidson\newline
\cite{moloi_high_2022}
\newline
(2022)
& \textbf{Focus:} \ac{hif} detection in radial distribution networks
\newline
\textbf{Data:} PowerFactory simulation of a \qty{132}\,kV/\qty{22}\,kV Eskom feeder network with 3 feeders; \num{4000} samples; \(\mathrm{f}_\mathrm{s} = \qty{12.5}\,\text{kHz}\); signal type: waveform \(V\)/\(I\)
\newline
\textbf{Method:} \ac{dwt} feature extraction (db4), statistical energy/entropy features, GA-optimized \ac{svm}; \ac{dt} and \ac{ann} for comparison
\newline
\textbf{Result:} \ac{svm} achieved \qty{97.6}\,\% accuracy (simulated), \qty{87.0}\,\% (experimental); lowest classification time: \qty{0.90}\,ms
\\ \addlinespace

Muntasir et al.\newline
\cite{muntasir_fast_2023}
\newline
(2023)
& \textbf{Focus:} \ac{fd} and \ac{fc} in HVAC transmission lines
\newline
\textbf{Data:} Simulink simulation of a 132\,kV, 1\,km line using real data from a 50\,MW plant in Bangladesh; \num{6804} samples; \(\mathrm{f}_\mathrm{s}\) unspecified; signal type: unspecified \(V\)/\(I\)
\newline
\textbf{Method:} Hybrid sampling, label encoding, and MinMax scaling; \ac{dt}, \ac{rf}, \ac{svm}, \ac{knn}, XGBoost, CatBoost, \ac{mlp}, and \ac{ann}
\newline
\textbf{Result:} 100\,\% accuracy for tree/boosting models; execution time per instance: 11.2\,ms (DT), 126.7\,ms (XGBoost)
\\ \addlinespace

Muzzammel et al.\newline
\cite{muzzammel_low_2022}
\newline
(2022)
& \textbf{Focus:} \ac{fd} and \ac{fc} in two-terminal \ac{vsc}-\ac{hvdc} transmission systems
\newline
\textbf{Data:} Simulink simulation of 100\,km bipolar \ac{vsc}-\ac{hvdc} system; pre- and post-fault TW voltage data; sample size \& \(\mathrm{f}_\mathrm{s}\) unspecified; signal type: DC \(V\)/\(I\)
\newline
\textbf{Method:} \ac{bml} for frequency-domain features; logistic regression for binary classification
\newline
\textbf{Result:} Fault classification in as low as 2.17\,ms; no accuracy, F1-score, or related metrics reported
\\ \addlinespace

Myint et al.\newline
\cite{myint_artificial_2024}
\newline
(2024)
& \textbf{Focus:} \ac{fc} and phase identification in transmission systems with railway traction loads
\newline
\textbf{Data:} Simulink simulation on modified IEEE 9-bus system with traction load; \num{1188} scenarios; \(\mathrm{f}_\mathrm{s}\) unspecified; signal type: waveform \(V\)/\(I\)
\newline
\textbf{Method:} \ac{dwt} (db6, 3-level) for modal current TW features; \ac{ann} trained via Bayesian backpropagation
\newline
\textbf{Result:} Accuracy of 96.1\,\% in \ac{fc}, including phase identification
\\ \addlinespace

Ojetola et al.\newline
\cite{ojetola_testing_2022}
\newline
(2022)
& \textbf{Focus:} \ac{fd} and \ac{fc} in low-voltage DC microgrids
\newline
\textbf{Data:} PSCAD simulation on \qty{750}\,V microgrid (\num{2400000} points); real data from ETL-KAFB testbed (\num{4000000} points); downsampled to \qty{10}\,µs; signal type: DC \(V\)/\(I\)
\newline
\textbf{Method:} \ac{svm}, \ac{nb}, \ac{dt}, \ac{nc}, \ac{mlp}; binary \ac{fd} and 5-class \ac{fc}; trained on simulation, tested on real data
\newline
\textbf{Result:} \ac{svm}, \ac{nc}, \ac{mlp} reached \qty{99}\,\% in \ac{fc}; only \ac{mlp} achieved \qty{100}\,\% in fault resistance classification
\\ \addlinespace

Patnaik et al.\newline
\cite{patnaik_modwt-xgboost_2021}
\newline
(2021)
& \textbf{Focus:} \ac{fd}, fault type, and resistance classification in low-voltage DC microgrids
\newline
\textbf{Data:} PSCAD simulation of \qty{750}\,V DC microgrid; real fault data from Kirtland AFB; \num{1585} samples; \(\mathrm{f}_\mathrm{s}\) unspecified; signal type: waveform \(V\)/\(I\)
\newline
\textbf{Method:} \ac{svm}, \ac{nc}, \ac{mlp}, \ac{dt}, \ac{nb}; binary \ac{fd}, multi-class \ac{fc} (type and resistance)
\newline
\textbf{Result:} \ac{svm}, \ac{nc}, \ac{mlp} achieved \qty{99}\,\% in fault type classification; only \ac{mlp} reached \qty{99}\,\% in fault resistance classification
\\ \addlinespace

Pragati et al. \newline
\cite{pragati_bayesian_2023}
\newline
(2023)
& \textbf{Focus:} \ac{fd} and multi-stage \ac{fc} in MT-\ac{vsc}-\ac{hvdc} systems
\newline
\textbf{Data:} Simulink simulation of four-terminal \qty{230}\,kV \ac{hvdc} grid; \num{17056} samples; \(\mathrm{f}_\mathrm{s}\) unspecified; signal type: DC \(V\)/\(I\)
\newline
\textbf{Method:} Bayesian-optimized \ac{dt} ensembles (AdaBoost, \ac{rusboost}); 3-class \ac{fd}, followed by 4-class \ac{fc} for internal DC faults
\newline
\textbf{Result:} \ac{fd} accuracy \qty{99.9}\,\% (internal vs. external faults); internal DC fault \ac{fc} accuracy \qty{99.3}\,\% across 4 fault types
\\ \addlinespace

Pragati et al.
\newline
\cite{pragati_fault_2024}
\newline
(2024)
& \textbf{Focus:} \ac{fd} and \ac{fc} in MT-\ac{vsc}-\ac{hvdc} systems; \ac{fc} aims to identify the faulted pole(s) (positive, negative, both, or none); signal type: DC \(V\)/\(I\)
\newline
\textbf{Data:} Simulink simulation of 4-terminal \qty{100}\,kV symmetrical monopolar \ac{hvdc} grid; \num{3744} fault and \num{13312} non-fault samples; \(\mathrm{f}_\mathrm{s} = \qty{3.2}\,\text{kHz}\)
\newline
\textbf{Method:} Stockwell transform + bidirectional GRU-\ac{rnn} for 3-class \ac{fd}; TKEO-based 4-class \ac{fc} to identify faulted pole(s)
\newline
\textbf{Result:} \ac{fd} accuracy \qty{99.92}\,\% (clean), \qty{99.02}\,\% (20 dB); \ac{fc} (phase ID) achieved \qty{100}\,\% accuracy across all noise levels
\\ \addlinespace

Prasad et al.\newline
\cite{prasad_tiwari_enhancing_2021}
\newline
(2021)
& \textbf{Focus:} \ac{fd}, \ac{fc}, and faulty section identification in reconfigurable DC microgrids with \ac{res}
\newline
\textbf{Data:} Simulink simulation of \ac{pv} and wind-based \acp{der}; \num{8930} samples; \(\mathrm{f}_\mathrm{s}\) unspecified; signal type: \(V\)/\(I\)
\newline
\textbf{Method:} Communication-free \ac{cnn}-based protection scheme
\newline
\textbf{Result:} \ac{fc} accuracy 99.26\,\%; 98.8\,\%–98.91\,\% for DER disconnection; response time <2\,ms
\\ \addlinespace

Ramadoss et al.\newline
\cite{ramadoss_machine_2023}
\newline
(2023)
& \textbf{Focus:} \ac{fd} and \ac{fc} of \ac{loe} in synchronous generators under \ac{psc}
\newline
\textbf{Data:} Simulink simulation of SMIB system; \num{98378} samples across CLOE, PLOE, PSC, and normal conditions; \(\mathrm{f}_\mathrm{s}\) unspecified; signal type: phasor domain \(V\)
\newline
\textbf{Method:} Hotelling’s T\textsuperscript{2} for \ac{fd}; binary \ac{svm} for \ac{fc} of \ac{loe} vs. non-\ac{loe} (normal/PSC)
\newline
\textbf{Result:} \ac{fd} time \qty{5.31}\,s; \ac{fc} accuracy \qty{96.6}\,\%, dependability \qty{96.2}\,\%, AUC = 0.985
\\ \addlinespace

Ramana et al.\newline
\cite{ramana_intelligent_2023}
\newline
(2023)
& \textbf{Focus:} \ac{fd} and \ac{fc} in TCSC-compensated transmission lines
\newline
\textbf{Data:} Simulink simulation of \qty{400}\,kV, \qty{320}\,km line with TCSC; \num{15000} fault cases; \(\mathrm{f}_\mathrm{s} = \qty{1}\,\text{kHz}\); signal type: waveform \(I\)
\newline
\textbf{Method:} \ac{fdst} feature extraction from 5-cycle current windows; ensemble bagging for binary \ac{fd} and 10-class \ac{fc}
\newline
\textbf{Result:} \ac{fd} accuracy \qty{100}\,\%, \ac{fc} accuracy \qty{99.99}\,\%; response time \qtyrange{3}{8}\,ms
\\ \addlinespace

Sadiq et al.\newline
\cite{sadiq_machine_2024}
\newline
(2024)
& \textbf{Focus:} \ac{fc} of parametric faults in passive components of Zeta converters for predictive maintenance in PV-integrated smart grids
\newline
\textbf{Data:} Simulink simulation of a \qty{48}\,V DC microgrid with PV input and Zeta converter; \num{1500} samples under 3 operating conditions; \(\mathrm{f}_\mathrm{s}\) unspecified; signal type: DC \(V\)/\(I\)
\newline
\textbf{Method:} Multi-class \ac{svm} (OvR), \ac{knn}, and \ac{dt}; classification into 5 classes (healthy, L1, L2, C1, C2 degradation)
\newline
\textbf{Result:} Quadratic \ac{svm} achieved highest accuracy \qty{99.2}\,\%, F1 score \qty{95.5}\,\%, outperforming other ML methods
\\ \addlinespace

Swetapadma et al.\newline
\cite{swetapadma_novel_2022}
\newline
(2022)
& \textbf{Focus:} \ac{fd} and \ac{fc} of internal vs. external faults near converters in bipolar LCC-\ac{hvdc} systems
\newline
\textbf{Data:} Simulink simulation of bipolar LCC-\ac{hvdc} link with filters and reactors; \num{10000} samples; \(\mathrm{f}_\mathrm{s} = \qty{5}\,\text{kHz}\); signal type: DC \(V\)/\(I\)
\newline
\textbf{Method:} Voltage-difference-based features; binary \ac{fc} using \ac{lstm}, \ac{rf}, and \ac{ann}
\newline
\textbf{Result:} \ac{lstm} achieved \qty{100}\,\% accuracy with response time <\qty{0.5}\,ms; robust under varying system conditions
\\ \addlinespace

Tsotsopoulou et al.\newline
\cite{tsotsopoulou_protection_2023}
\newline
(2023)
& \textbf{Focus:} Centralized \ac{fd} and \ac{fc} in MT-\ac{hvdc} with superconducting cables
\newline
\textbf{Data:} Simulink simulation of MT-\ac{hvdc} grid with MMCs and SCs; \num{1800} samples; \(\mathrm{f}_\mathrm{s} = \qty{20}\,\text{kHz}\); signal type: DC \(V\)/\(I\)
\newline
\textbf{Method:} XGBoost on local voltage/current; no communication needed
\newline
\textbf{Result:} \(F_1 = \qty{98}\,\%\); operation time \qtyrange{0.08}{1.23}\,ms; noise robust
\\ \addlinespace

Uddin et al.\newline
\cite{uddin_protection_2022}
\newline
(2022)
& \textbf{Focus:} \ac{fd} and \ac{fc} in transmission lines
\newline
\textbf{Data:} Simulink simulation on benchmark TL model (IEC standard); \num{4950} samples; \(\mathrm{f}_\mathrm{s} = \qty{20}\,\text{kHz}\); signal type: waveform \(V\)/\(I\)
\newline
\textbf{Method:} \ac{dwt} feature extraction; \ac{svm} tuned via GridSearchCV
\newline
\textbf{Result:} Accuracy \qty{99.46}\,\%; response time 1.99\,ms; \qty{96.11}\,\% accuracy at 2\,kHz; robust under varying fault and source conditions
\\ \addlinespace

Wang et al.\newline
\cite{wang_use_2020}
\newline
(2020)
& \textbf{Focus:} \ac{fd} and \ac{fc} of \acp{hif} in radial distribution systems for enhanced electrical safety
\newline
\textbf{Data:} Simulink simulation with improved \ac{hif} model on radial system; \num{60000} samples; \(\mathrm{f}_\mathrm{s} = \qty{7680}\,\text{Hz}\); signal type: waveform \(V\)/\(I\)
\newline
\textbf{Method:} PCQ-\ac{wt} feature extraction with modified Gabor wavelets; compact \ac{cnn}-based classifier
\newline
\textbf{Result:} Accuracy \qty{99.95}\,\%; detection delay \qty{33}\,ms + processing time \qty{6.3}\,ms
\\ \addlinespace

Wang et al.\newline
\cite{wang_detection_2024}
\newline
(2024)
& \textbf{Focus:} \ac{fd} of abnormal transformer-user topologies in three-phase distribution systems
\newline
\textbf{Data:} Real voltage data from 3 substations; \num{14000} samples; \qty{50}\,Hz sampling; \num{128}-point window; no simulation platform used; signal type: phasor domain \(V\)
\newline
\textbf{Method:} ResNet-18 with data augmentation; binary \ac{fc}
\newline
\textbf{Result:} Accuracy \qty{96.8}\,\%, F1-score \qty{96.1}\,\%, AUC \qty{98.0}\,\%
\\ \addlinespace

Wong et al.\newline
\cite{wong_hybrid_2024}
\newline
(2024)
& \textbf{Focus:} \ac{fd} and \ac{fc} in hybrid HVAC-\ac{hvdc} transmission systems using TW-based signals
\newline
\textbf{Data:} Simulink simulation of bipolar hybrid HVAC-HVDC grid with rectifying VSCs; \num{116} fault scenarios; \(\mathrm{f}_\mathrm{s}\) unspecified; signal type: waveform \(V\)/\(I\)
\newline
\textbf{Method:} TW feature extraction (voltage, current, real and reactive power); pattern-recognition \ac{ann} for binary \ac{fd} and \ac{fc}
\newline
\textbf{Result:} Accuracy \qty{96.25}\,\%; recall \qty{96.29}\,\%, precision \qty{92.86}\,\%
\\ \addlinespace

% ANN-based robust DC fault protection algorithm for MMC high-voltage direct current grids
Xiang et al.\newline
\cite{xiang_ann-based_2020}
\newline
(2020)
& \textbf{Focus:} \ac{fd} and pole-specific \ac{fc} for DC line and bus faults in MMC-based \ac{hvdc} grids
\newline
\textbf{Data:} EMTDC simulation on 4-terminal symmetrical monopole MMC-\ac{hvdc} grid; \num{816} samples; \(\mathrm{f}_\mathrm{s} = \qty{10}\,\text{kHz}\); signal type: DC \(V\)/\(I\)
\newline
\textbf{Method:} 1-level \ac{dwt} on transient voltages; 5-output \ac{ann} for \ac{fc} (PTP, P-PTG, N-PTG, DC line, DC bus)
\newline
\textbf{Result:} Accuracy \qty{100}\,\% up to \qty{350}\,ohm fault resistance; \qty{94}\,\% at \qty{400}\,ohm; \qty{78}\,\% at \qty{500}\,ohm; detection delay <\qty{2.5}\,ms
\\ \addlinespace

% A novel dc fault protection scheme based on intelligent network for meshed dc grids
Yousaf et al.\newline
\cite{yousaf_novel_2023}
\newline
(2023)
& \textbf{Focus:} \ac{fd} and \ac{fc} in \ac{mt-hvdc} grids using single-ended \ac{ann}-based main and backup protection with \ac{bo}-optimized hyperparameters
\newline
\textbf{Data:} EMTDC simulation on 4-terminal \qty{\pm320}\,kV MMC-\ac{mt-hvdc} mesh; \num{1260} training and \num{486} test samples; fault resistances up to \qty{500}\,ohm, 20–\qty{45}\,dB noise; \(\mathrm{f}_\mathrm{s} = \qty{50}\,\text{kHz}\); signal type: DC \(V\)/\(I\)
\newline
\textbf{Method:} \ac{dwt}-based MRA and Parseval’s theorem for feature extraction; 3-layer \ac{ffnn}; \ac{bo} with GP and Expected Improvement to optimize learning rate, neurons, epochs
\newline
\textbf{Result:} Accuracy \qty{99.38}\,\% (main) and \qty{100.0}\,\% (backup); robust to \qty{465}\,ohm faults and \qty{20}\,dB noise; 5× faster training than grid/random search
\\ \addlinespace

% Deep Learning-Based Robust DC Fault Protection Scheme for Meshed HVDC Grids
Yousaf et al.\newline
\cite{yousaf_deep_2023}
\newline
(2023)
& \textbf{Focus:} \ac{dl}-based \ac{fd} and \ac{fc} for DC transmission lines in \ac{mt-hvdc} grids
\newline
\textbf{Data:} EMTDC simulation of custom \qty{320}\,kV 4-terminal \ac{mt-hvdc} mesh; \num{1892} samples; \(\mathrm{f}_\mathrm{s} = \qty{50}\,\text{kHz}\); signal type: DC \(V\)/\(I\)
\newline
\textbf{Method:} \ac{lstm} with \ac{dwt}; communication-free design
\newline
\textbf{Result:} Accuracy of \qty{98.6}\,\%, robust to noise
\\ \addlinespace

\end{longtable}

%% file: tables/tab_fault_location.tex
\begin{longtable}{@{}p{2.4cm}p{14.4cm}@{}}
\caption{Comparison of Recently Proposed ML-Based Approaches for Fault Localization.} \label{tab:fault_location} \\
\toprule
\textbf{Author} & \textbf{Description} \\
\midrule
\endfirsthead

\toprule
\textbf{Author} & \textbf{Description} \\
\midrule
\endhead

\bottomrule
\endfoot

Afrasiabi et al.
\newline
\cite{afrasiabi_fault_2022}
\newline
(2022)
& \textbf{Focus:} \ac{fl} and \ac{fli} in transmission systems using \ac{wams} under noisy conditions
\newline
\textbf{Data:} Simulink simulations on IEEE 68-bus system (\num{44000} samples) and 145-bus system (\num{61000} samples); \(\mathrm{f}_\mathrm{s}\) unspecified; signal type: phasor domain \(V\)/\(I\)
\newline
\textbf{Method:} Improved gated recurrent neural network (IGRNN)
\newline
\textbf{Result:} Accuracy of \qty{98}\,\% (68-bus) and \qty{96}\,\% (145-bus); average runtime of \qty{2.64}\,ms
\\ \addlinespace

Alhanaf et al.
\newline
\cite{alhanaf_fault_2024}
\newline
(2024)
& \textbf{Focus:} \ac{fc}, \ac{fli}, and \ac{fl} in ring power systems with varying \ac{dg} penetration and topology
\newline
\textbf{Data:} Simulink simulations on IEEE 6-bus (\num{115220} samples) and 9-bus (\num{51840}) systems with diverse fault types, locations, and resistances; \(\mathrm{f}_\mathrm{s}\) unspecified; signal type: phasor domain \(V\)/\(I\)
\newline
\textbf{Method:} \ac{cnn}, \ac{lstm}, and hybrid \ac{cnn}-\ac{lstm}
\newline
\textbf{Result:} Hybrid model achieved up to \(99.97\%\) (\ac{fc}), \(99.71\%\) (\ac{fli}), and \(97.25\%\) (\ac{fl}) accuracy on 6-bus; slightly lower on 9-bus
\\ \addlinespace

Alhanaf et al.\newline
\cite{alhanaf_intelligent_2023}
\newline
(2022)
& \textbf{Focus:} \ac{fd}, \ac{fc}, and \ac{fl} in smart grids using deep learning
\newline
\textbf{Data:} Simulink on IEEE 6-bus system; \num{249030} voltage and current samples; \(\mathrm{f}_\mathrm{s} = \qty{2.5}\,\text{kHz}\); signal type: waveform \(V\)/\(I\)
\newline
\textbf{Method:} \ac{ann} and 1D-\ac{cnn}
\newline
\textbf{Result:} Up to \qty{99.99}\,\% for \ac{fd} and \ac{fc}; \ac{fl} accuracy lower (min \qty{96.85}\,\%)
\\ \addlinespace

Alilouch et al.\newline
\cite{alilouch_intelligent_2022}
\newline
(2022)
& \textbf{Focus:} \ac{fd}, \ac{fc}, and \ac{fl} for a single \qty{400}\,kV, \qty{100}\,km transmission line
\newline
\textbf{Data:} Simulink simulations with various types, resistances, and locations; sample size and \(\mathrm{f}_\mathrm{s}\) not reported; signal type: waveform \(V\)/\(I\)
\newline
\textbf{Method:} \ac{ann}
\newline
\textbf{Result:} No evaluation metrics reported
\\ \addlinespace

Alrashdan et al.\newline
\cite{alrashdan_artificial_2023}
\newline
(2023)
& \textbf{Focus:} \ac{fd}, \ac{fc}, and \ac{fl} in \ac{dfig}-based \qty{220}\,kV transmission systems
\newline
\textbf{Data:} Simulink simulation; \num{18360} samples; \(\mathrm{f}_\mathrm{s} = \qty{1.92}\,\text{kHz}\); signal type: waveform \(V\)/\(I\)
\newline
\textbf{Method:} \ac{ann} and \ac{dwt}
\newline
\textbf{Result:} \qty{99.6}\,\% accuracy for \ac{fc}, \qty{98.3}\,\% for fault area classification
\\ \addlinespace

Banihashemi and Cuzner
\newline
\cite{banihashemi_novel_2023}
\newline
(2023)
& \textbf{Focus:} Fault area classification and zone discrimination in \ac{mg} ring networks
\newline
\textbf{Data:} Typhoon HIL simulation of 13.8\,kV 2-bus \ac{mg} with synchronous generator and PV (\num{240} samples); \(\mathrm{f}_\mathrm{s}\) = \qty{6}{\kilo\hertz}; multi-class dataset across load and PV scenarios; signal type: phasor domain \(V\)/\(I\)
\newline
\textbf{Method:} XGBoost classifier with 4 protection zones; evaluated with and without communication
\newline
\textbf{Result:} Accuracy \qty{93.82}\,\% (2 cycles); robust primary/backup zone discrimination and reduced fault clearance time
\\ \addlinespace

Bera et al.
\newline
\cite{bera_hybrid_2024}
\newline
(2024)
& \textbf{Focus:} \ac{fd}, \ac{fc}, and \ac{fl} in single-ended \qty{230}\,kV transmission line in the zone between the grid and the PV farm
\newline \textbf{Data:} PSCAD simulations on IEEE 9-bus system with PV farm at bus-9 (\num{28800} samples); \(\mathrm{f}_\mathrm{s} = \qty{7680}\,\text{Hz}\); signal type: waveform \(V\)/\(I\)
\newline \textbf{Method:} \ac{dt}, \ac{rf}, \ac{svm}, \ac{knn}, and \ac{nb}
\newline \textbf{Result:} Using \ac{rf} balanced accuracy of \qty{98.2}\,\% for \ac{fd}, \qty{92.3}\,\% for \ac{fl}, \qty{97.2}\,\% for phase selection 
\\ \addlinespace

Cholla et al.\newline
\cite{cholla_end--end_2023}
\newline
(2023)
& \textbf{Focus:} \ac{fd} and \ac{fl} in transmission lines using lightweight learning models
\newline
\textbf{Data:} Single transmission line; no details provided on simulation software, sample size, or sampling rate; signal type: unspecified \(V\)/\(I\)
\newline
\textbf{Method:} \ac{elm} and summation wavelet extreme learning machine
\newline
\textbf{Result:} Accuracy of \qty{99.4}\,\%
\\ \addlinespace

Farshad et al.\newline
\cite{farshad_intelligent_2021} \newline
(2021)
& \textbf{Focus:} \ac{fd} and \ac{fc} of internal and external \ac{sc} faults in \ac{csc}-\ac{hvdc} systems
\newline
\textbf{Data:} EMTDC and Simulink simulations based on CIGRE bipolar \ac{hvdc} model; \num{7863} samples; \(\mathrm{f}_\mathrm{s}\) unspecified; signal type: DC \(V\)/\(I\)
\newline
\textbf{Method:} Multiclass \ac{svm} using features from moving average and max coordinate difference
\newline
\textbf{Result:} Average \ac{fd} time up to \qty{10.7}\,ms; external fault detection accuracy \qty{99.375}\,\%
\\ \addlinespace

Garika et al.\newline
\cite{garika_condition_2022} \newline
(2022)
& \textbf{Focus:} Fault analysis and condition monitoring in IoT-enabled wind–\ac{svm} integrated \ac{hv} networks
\newline
\textbf{Data:} \num{55} training and \num{9} testing samples per zone; \(\mathrm{f}_\mathrm{s} = \qty{264}\,\text{kHz}\); no simulation tool specified; signal type: waveform \(V\)/\(I\)
\newline
\textbf{Method:} \ac{ann} using wavelet features (bior1.5)
\newline
\textbf{Result:} No quantitative performance metrics reported
\\ \addlinespace

Gnanamalar et al.\newline
\cite{gnanamalar_cnnsvm_2023} \newline
(2023)
& \textbf{Focus:} \ac{fd}, \ac{fc}, and \ac{fl} in multi-terminal \ac{vsc}-\ac{hvdc} systems for offshore wind integration
\newline
\textbf{Data:} Simulink simulation; \num{335} samples; \(\mathrm{f}_\mathrm{s}\) unspecified; signal type: DC \(V\)/\(I\)
\newline
\textbf{Method:} Hybrid \ac{cnn}-\ac{svm} model with Hilbert–Huang transform feature extraction; compared with \ac{svm}, \ac{ann}, and \ac{rnn}
\newline
\textbf{Result:} Accuracy \qty{99.87}\,\%; \ac{fd}, \ac{fc}, and \ac{fl} completed within \qty{2}\,ms
\\ \addlinespace

Goli et al.\newline
\cite{goli_neuro-wavelet_2022}
\newline
(2022)
& \textbf{Focus:} \ac{fd} and \ac{fl} in a 6-terminal hybrid \ac{pv}-Wind \ac{mg} using neuro-wavelet protection
\newline
\textbf{Data:} \num{21600} samples; \(\mathrm{f}_\mathrm{s} = \qty{216}\,\text{kHz}\); simulation software not reported; signal type: waveform \(V\)/\(I\)
\newline
\textbf{Method:} \ac{ann} with bior1.5 wavelet feature extraction
\newline
\textbf{Result:} Fault location error within \qtyrange{-3.97}{3.68}\,\%
\\ \addlinespace

Hassan et al.\newline
\cite{hassan_fault_2024}\newline
(2024)
& \textbf{Focus:} \ac{fc} and \ac{fl} in transmission lines
\newline
\textbf{Data:} Simulink simulation on IEEE 9-bus system; \num{20760} samples; sampling rate not reported; signal type: phasor domain \(V\)/\(I\)
\newline
\textbf{Method:} Optimizable Weighted Extreme Learning Machine
\newline
\textbf{Result:} \qty{99.83}\,\% classification accuracy; \qty{95.8}\,\% localization accuracy
\\ \addlinespace

Imani et al.\newline
\cite{imani_novel_2024} \newline
(2024)
& \textbf{Focus:} \ac{fd}, \ac{fc}, and \ac{fl} in \ac{mtdc} transmission systems
\newline
\textbf{Data:} Simulink simulations on a 4-terminal \ac{mmc}–\ac{hvdc} grid; \num{1212} samples; \(\mathrm{f}_\mathrm{s}\) unspecified; signal type: DC \(V\)/\(I\)
\newline
\textbf{Method:} Maximal overlap discrete \ac{wt}-based feature extraction; classification using rate of change of voltage and general regression \ac{nn}
\newline
\textbf{Result:} \ac{fd} up to \qty{200}\,ohm; \ac{fl} error of \qty{0.319}\,\%; \ac{fc} accuracy \(\geq \qty{94}\,\%\)
\\ \addlinespace

Jayasinghe et al.\newline
\cite{jayasinghe_classification_2024}
\newline
(2024)
& \textbf{Focus:} \ac{fc} and \ac{fl} in low-voltage microgrids
\newline
\textbf{Data:} Simulink and EMTDC simulations on \qty{400}\,V AC microgrid (3-line system, University of Moratuwa); sample size \& \(\mathrm{f}_\mathrm{s}\) unspecified; signal type: waveform \(V\)/\(I\)
\newline
\textbf{Method:} \ac{dwt} feature extraction; \ac{ann} models for \ac{fc} and \ac{fl}
\newline
\textbf{Result:} \ac{fc} accuracy \qty{99.39}\,\%; decision time < \qty{72.5}\,ms; \ac{fl} \(R^2 = 0.866\)
\\ \addlinespace

Kandasamy et al.\newline
\cite{kandasamy_intelligent_2024} \newline
(2024)
& \textbf{Focus:} \ac{fd}, \ac{fc}, and \ac{fl} in active distribution networks
\newline
\textbf{Data:} EMTP-RV/Simulink simulations on modified IEEE 13- and 34-bus \ac{adn}s; \num{51158} samples; \(\mathrm{f}_\mathrm{s}\) unspecified; signal type: phasor domain \(V\)/\(I\)
\newline
\textbf{Method:} \ac{dost}-based preprocessing; \ac{afd}-\ac{cnn} optimized via multi-objective firefly algorithm; comparison with \ac{mlp} baseline
\newline
\textbf{Result:} Accuracy of \qty{99.74}\,\% (\ac{fd}), \qty{99.69}\,\% (\ac{fc}), and \qty{99.58}\,\% (\ac{fl}); fault location error between \qtyrange{0.6}{2}\,\% at \qty{1000}\,ohm
\\ \addlinespace

Kumari et al.
\newline
\cite{kumari_enhancing_2024}
\newline
(2024)
& \textbf{Focus:} \ac{fd}, \ac{fc}, and \ac{fl} in AC microgrids under grid-connected and islanded operation modes
\newline
\textbf{Data:} Simulink simulations on 10-bus AC microgrid with DG, PV, and variable fault resistances (\num{2442000} samples); \(\mathrm{f}_\mathrm{s}\) unspecified; signal type: phasor domain \(V\)/\(I\)
\newline
\textbf{Method:} \ac{ann}
\newline
\textbf{Result:} \ac{fc} accuracy \qty{98.82}\,\%; \ac{fl} \qty{97.4}\,\%; feeder isolation \qty{92.3}\,\%; robust to \qty{300}\,ohm faults and \qty{20}\,dB noise 
\\ \addlinespace

Mampilly et al.\newline
\cite{mampilly_empirical_2024}
\newline
(2024)
& \textbf{Focus:} \ac{fd}, \ac{fc}, and \ac{fl} in hybrid \ac{res}-based microgrids
\newline
\textbf{Data:} Simulink simulation of IEEE 13-bus system; \num{44000} samples; \(\mathrm{f}_\mathrm{s}\) unspecified; signal type: waveform \(V\)/\(I\)
\newline
\textbf{Method:} \ac{ewt} feature extraction; POA-optimized hybrid HC-\ac{rnn}
\newline
\textbf{Result:} Accuracy \qty{99.97}\,\%; outperformed 13 state-of-the-art methods; +\qty{1.56}\,\% over SVM
\\ \addlinespace

Meyer et al.\newline
\cite{meyer_system-based_2020} \newline
(2020)
& \textbf{Focus:} \ac{fl} in transmission networks with incomplete \ac{dp} coverage
\newline
\textbf{Data:} PSS NETOMAC simulations on modified IEEE 9-bus system; \num{9800} three-phase fault scenarios with varying fault resistances; \(\mathrm{f}_\mathrm{s}\) unspecified; signal type: phasor domain \(V\)/\(I\)
\newline
\textbf{Method:} Feedforward \ac{ann} trained on patterns from 14 \ac{dp} relays
\newline
\textbf{Result:} \ac{fl} accuracy \(\geq \qty{95.1}\,\%\) across all lines and resistances; \qty{98.6}\,\% for unmonitored line sections
\\ \addlinespace

Miraj et al.\newline
\cite{miraj_improved_2023} \newline
(2023)
& \textbf{Focus:} \ac{fc} and \ac{fl} in meshed AC microgrids
\newline
\textbf{Data:} Simulink simulations on 9-bus AC microgrid with 6 line sections (\num{6480} samples); \(\mathrm{f}_\mathrm{s}\) unspecified; signal type: phasor domain \(V\)/\(I\)
\newline
\textbf{Method:} \ac{svm} for \ac{fc} and \ac{svm} for \ac{fl}
\newline
\textbf{Result:} \ac{fc} accuracy \qty{98.99}\,\%; \ac{fl} \(RMSE=0.0081\); training time under \qty{94}\,s
\\ \addlinespace

Muzzammel et al.\newline
\cite{muzzammel_support_2020} \newline
(2020)
& \textbf{Focus:} \ac{fd}, \ac{fc}, and \ac{fl} in \ac{mt-hvdc} transmission systems
\newline
\textbf{Data:} Simulink simulations on 4-terminal \ac{mt-hvdc} grid; \num{81001} samples; \(\mathrm{f}_\mathrm{s}\) unspecified; signal type: DC \(V\)/\(I\)
\newline
\textbf{Method:} \ac{pca}-based feature reduction followed by \ac{svm} for \ac{fd}, \ac{fc}, and \ac{fl}
\newline
\textbf{Result:} Accurate \ac{fd}, \ac{fc}, and \ac{fl} within \qty{0.15}\,ms; \ac{fl} classification accuracy \qty{100}\,\%
\\ \addlinespace

Muzzammel et al.\newline
\cite{muzzammel_wavelet_2023} \newline
(2023)
& \textbf{Focus:} \ac{fd}, \ac{fc}, and \ac{fl} in \ac{mt-hvdc} transmission systems
\newline
\textbf{Data:} Simulink simulations on 4-terminal \ac{mt-hvdc} grid; sample size and \(\mathrm{f}_\mathrm{s}\) unspecified; signal type: DC \(V\)/\(I\)
\newline
\textbf{Method:} Traveling wave and wavelet-based feature extraction with multiscale \ac{pca}; classification and localization via \ac{svm}
\newline
\textbf{Result:} \ac{fc} accuracy \qty{99.3}\,\%; \ac{fl} accuracy \qty{98.4}\,\%; total protection time \qty{4}\,ms
\\ \addlinespace

N et al.\newline
\cite{n_transmission_2021} \newline
(2021)
& \textbf{Focus:} \ac{fd}, \ac{fc}, and \ac{fl} in transmission lines
\newline
\textbf{Data:} Simulink simulation of a single \qty{132}\,kV, \qty{100}\,km transmission line; \num{1200} samples; \(\mathrm{f}_\mathrm{s}\) unspecified; signal type: waveform \(V\)/\(I\)
\newline
\textbf{Method:} \ac{mlp} trained on raw current and voltage signals for all tasks
\newline
\textbf{Result:} \ac{fd} accuracy \qty{100}\,\%; \ac{fc} accuracy \qty{53.3}\,\%; \ac{fl} regression \(R \approx 0.99\)
\\ \addlinespace

Najafzadeh et al.\newline
\cite{najafzadeh_fault_2024} \newline
(2024)
& \textbf{Focus:} \ac{fd}, \ac{fc}, and \ac{fl} in \ac{hv} transmission systems using \ac{pmu}-based measurements
\newline
\textbf{Data:} Simulink simulations on a \qty{735}\,kV transmission line; \num{1100} samples; \(\mathrm{f}_\mathrm{s}\) unspecified; signal type: waveform \(V\)/\(I\)
\newline
\textbf{Method:} WHO-optimized \ac{rf}, \ac{dt}, and (ANFIS); fuzzy logic for thresholding
\newline
\textbf{Result:} \ac{fc} accuracy \qty{100}\,\% (\ac{rf}), \qty{98.1}\,\% (\ac{dt}); \ac{fl} error < \qty{153.6}\,m; robust under \qty{10}\,dB noise
\\ \addlinespace

Ponukumati et al.\newline
\cite{ponukumati_pattern_2021} \newline
(2021)
& \textbf{Focus:} \ac{fd}, \ac{fc}, and \ac{fl} in multi-\ac{mg} systems with \ac{res} integration
\newline
\textbf{Data:} Simulations on multi-\ac{mg} testbed with \ac{pv}, wind, and battery; modified IEEE 33-bus system; \num{1536} samples; simulation tool and \(\mathrm{f}_\mathrm{s}\) unspecified; signal type: phasor domain \(V\)/\(I\)
\newline
\textbf{Method:} Backpropagation \ac{ann} using DWT and \ac{pll}-based features
\newline
\textbf{Result:} Accuracy \(> \qty{90}\,\%\) for \ac{fc} and \ac{fl} across all fault types
\\ \addlinespace

Poudel et al.\newline
\cite{poudel_zonal_2022}
\newline
(2022)
& \textbf{Focus:} Fault area classification in distribution systems with \acp{der}
\newline
\textbf{Data:} PSS®CAPE, OpenDSS simulations on IEEE 123-bus system (4.16\,kV) with 9 \acp{ibr}, multiple circuit topologies, and diverse fault scenarios; sample size and \(\mathrm{f}_\mathrm{s}\) unspecified; signal type: phasor domain \(V\)/\(I\)
\newline
\textbf{Method:} \ac{svm} classifiers for circuit topology, fault type, and zone identification
\newline
\textbf{Result:} Topology estimation, fault type, and zone classification all above \qty{99}\,\% accuracy; response time \(\sim\)2.6\,ms after RMS calculation; robust operation without communication
\\ \addlinespace

Prabhakar\newline
\cite{prabhakar_distance_2023} \newline
(2023)
& \textbf{Focus:} \ac{fc} and \ac{fl} in wind-integrated transmission systems for adaptive distance protection
\newline
\textbf{Data:} DIgSILENT PowerFactory simulations on IEEE 9-bus system with wind farm at bus 8; sample size and \(\mathrm{f}_\mathrm{s}\) unspecified; signal type: phasor domain \(V\)/\(I\)
\newline
\textbf{Method:} Hybrid \ac{svm}-\ac{ann} model; \ac{svm} for \ac{fc}, \ac{ann} for \ac{fl}
\newline
\textbf{Result:} \ac{fc} accuracy \qty{88.5}\,\% with partial data, \qty{100}\,\% with full-bus measurements; \ac{fl} with negligible error
\\ \addlinespace

Qusayer et al.\newline
\cite{qusayer_communication_2024} \newline
(2024)
& \textbf{Focus:} \ac{fd} and \ac{fl} in multi-\ac{mg} systems under grid-connected, islanded, and hybrid modes with communication-assisted coordination
\newline
\textbf{Data:} Simulink simulations on modified IEEE 14-bus multi-\ac{mg} system; \num{478980} samples; \(\mathrm{f}_\mathrm{s}\) unspecified; signal type: waveform \(V\)/\(I\)
\newline
\textbf{Method:} Centralized \ac{ann}-based protection using Levenberg–Marquardt backpropagation
\newline
\textbf{Result:} Testing accuracy up to \qty{100}\,\% across configurations; all models \(\geq \qty{99.21}\,\%\); includes end-to-end evaluation of \ac{ann} and communication delays
\\ \addlinespace

Ren et al.\newline
\cite{ren_fault_2024} \newline
(2024)
& \textbf{Focus:} \ac{fl} in multi-terminal \ac{hvdc} grids using single-ended voltage measurements
\newline
\textbf{Data:} EMTDC simulations on a 4-terminal \ac{hvdc} grid; \num{4752} samples; \(\mathrm{f}_\mathrm{s}\) unspecified; signal type: DC \(V\)/\(I\)
\newline
\textbf{Method:} Panoramic voltage feature extraction via \ac{gaf}, \ac{rp}, \ac{mtf}, and \ac{cwt}; regression using lightweight \ac{vit} variants (MobileViT, \ac{emo}, \ac{edgevit})
\newline
\textbf{Result:} \ac{fl} error \qty{0.441}\,km; robust to faults up to \qty{400}\,ohm, noise up to \qty{30}\,dB, and sampling rates as low as \qty{30}\,\text{kHz}
\\ \addlinespace

Rizk-Allah et al.\newline
\cite{rizk-allah_dynamic_2023} \newline
(2023)
& \textbf{Focus:} \ac{fl} in long \ac{hv} transmission lines using two-ended measurements
\newline
\textbf{Data:} ATP and Simulink simulations on a \qty{220}\,km transmission line; \num{1050} samples; \(\mathrm{f}_\mathrm{s}\) unspecified; signal type: phasor domain \(V\)/\(I\)
\newline
\textbf{Method:} Dynamically shaped feedforward \ac{ann} optimized via \ac{pso}; compared against \ac{rf} and cross-validated \ac{ann}
\newline
\textbf{Result:} Proposed model achieved \ac{fl} error \(\leq \qty{0.40}\,km\); baseline models showed errors up to \qty{3.01}\,km
\\ \addlinespace

Rong et al.\newline
\cite{rong_convolutional_2023} \newline
(2023)
& \textbf{Focus:} Faulted feeder identification in low-current resonant-earthed distribution systems
\newline
\textbf{Data:} EMTP/ATP and RelaySimTest simulations on 6-feeder system; \num{1800000} samples; \(\mathrm{f}_\mathrm{s} = \qty{2400}\,\text{Hz}\); signal type: waveform \(V\)/\(I\)
\newline
\textbf{Method:} \ac{cnn} trained on transient superimposed zero-sequence currents within first post-fault cycle
\newline
\textbf{Result:} Accuracy \(\geq \qty{99.5}\,\%\) in simulation; \qty{100}\,\% in lab tests up to \qty{5000}\,ohm fault resistance
\\ \addlinespace

Sapountzoglou et al.\newline
\cite{sapountzoglou_fault_2020}\newline
(2020)
& \textbf{Focus:} \ac{fd}, \ac{fc}, and \ac{fl} in low-voltage smart grids
\newline
\textbf{Data:} Simulink simulation of a semi-rural LV grid in Portugal; \num{230688} samples; \( \mathrm{f}_\mathrm{s} = \qty{20}\,\text{Hz} \); signal type: phasor domain \(V\)/\(I\)
\newline
\textbf{Method:} \ac{gbt} via XGBoost
\newline
\textbf{Result:} \qty{100}\,\% feeder ID accuracy; \qtyrange{86.7}{99.7}\,\% fault classification; \qtyrange{86.2}{95.8}\,\% localization accuracy
\\ \addlinespace

Srivastava et al.\newline
\cite{srivastava_random_2023} \newline
(2023)
& \textbf{Focus:} \ac{fd}, \ac{fc}, and \ac{fl} in AC microgrids with variable \ac{dg} configurations
\newline
\textbf{Data:} Simulink simulations on 7-bus AC microgrid with \ac{dg} and \ac{pv} sources; \num{34020} samples; \(\mathrm{f}_\mathrm{s}\) unspecified; signal type: phasor domain \(V\)/\(I\)
\newline
\textbf{Method:} \ac{rf} with task-wise data filtering; compared with \ac{dt} and \ac{ann}
\newline
\textbf{Result:} \ac{fc} accuracy \qty{99.73}\,\%; line identification \qty{97.08}\,\%; \ac{fl} error MSE reduced to \num{0.25} via task-specific segregation
\\ \addlinespace

Taheri et al.\newline
\cite{taheri_fault-location_2021} \newline
(2021)
& \textbf{Focus:} \ac{fl} in wind-integrated parallel transmission systems using distributed parameter models
\newline
\textbf{Data:} PSCAD simulations on two-circuit \qty{345}\,kV transmission network with wind and synchronous generation; sample size and \(\mathrm{f}_\mathrm{s}\) unspecified; signal type: waveform \(V\)/\(I\)
\newline
\textbf{Method:} \ac{lstm}-based estimation of network impedance and fault location
\newline
\textbf{Result:} \ac{fl} error \(< \qty{0.86}\,\%\); robust to faults up to \qty{150}\,ohm and noise up to \qty{20}\,dB
\\ \addlinespace

Taheri et al.\newline
\cite{taheri_novel_2022} \newline
(2022)
& \textbf{Focus:} \ac{fl} in wind-integrated, shunt-compensated parallel transmission lines
\newline
\textbf{Data:} PSCAD simulations on double-circuit \qty{345}\,kV transmission line with wind farm and \ac{svm}; sample size and \(\mathrm{f}_\mathrm{s}\) unspecified; signal type: waveform \(V\)/\(I\)
\newline
\textbf{Method:} \ac{lstm}-based estimation of source impedance and fault distance
\newline
\textbf{Result:} \ac{fl} error \(< \qty{0.98}\,\%\); robust to \qty{150}\,ohm faults and noise at \qty{20}\,dB SNR
\\ \addlinespace

Tiwari\newline
\cite{tiwari_deep_2023} \newline
(2023)
& \textbf{Focus:} \ac{fd}, \ac{fc}, and \ac{fl} in wind-integrated \ac{hvdc} transmission systems under variable fault and operating conditions
\newline
\textbf{Data:} Simulink simulations on \ac{hvdc} system integrated with \ac{dfig}-based wind array; \num{22450} samples; \(\mathrm{f}_\mathrm{s}\) unspecified; signal type: DC \(V\)/\(I\)
\newline
\textbf{Method:} \ac{cnn}; compared against \ac{svm} and \ac{dt}
\newline
\textbf{Result:} \ac{fc} accuracy \qty{99.58}\,\%; \ac{fl} error down to \qty{0.104}\,\%; validated via dependability and security metrics
\\ \addlinespace

Tong et al.\newline
\cite{tong_semi_2022} \newline
(2022)
& \textbf{Focus:} \ac{fd}, \ac{fc}, and \ac{fl} in \ac{mmc}-based \ac{mtdc} systems using local measurements
\newline
\textbf{Data:} EMTDC simulations on 4-terminal \qty{500}\,kV \ac{mmc}-\ac{mtdc} grid; \num{3100} samples; \(\mathrm{f}_\mathrm{s} = \qty{50}\,\text{kHz}\); signal type: DC \(V\)/\(I\)
\newline
\textbf{Method:} Semi-\ac{ai} protection scheme with startup logic and \ac{cnn}-based internal fault identification; \qty{1}\,ms data window
\newline
\textbf{Result:} \qty{100}\,\% accuracy up to \qty{500}\,ohm faults; \qty{94}\,\% at \qty{800}\,ohm; robust for SNR > \qty{30}\,dB
\\ \addlinespace

Tran \newline
\cite{tran_integration_2023}
\newline
(2023)
& \textbf{Focus:} \ac{fl} enhancement for distance relays using \ac{ml} based on local voltage and current signals
\newline
\textbf{Data:} ATP/EMTP simulation; \num{2136} samples; \(\mathrm{f}_\mathrm{s}\) unspecified; signal type: waveform \(V\)/\(I\)
\newline
\textbf{Method:} \ac{mlp} integrated with Siemens 7SA522 relay via Omicron CMC-356
\newline
\textbf{Result:} Average \ac{fl} error reduced from \qty{0.92}\,\% to \qty{0.42}\,\%; max error reduced from \qty{8.23}\,\% to \qty{2.49}\,\%
\\ \addlinespace

Uddin et al.\newline
\cite{uddin_hybrid_2022}
\newline
(2022)
& \textbf{Focus:} \ac{fd}, \ac{fc}, and fault area classification in grid-connected \ac{dfig}-based wind farms
\newline
\textbf{Data:} PSCAD, ETAP, and Simulink simulations; sample size and \(\mathrm{f}_\mathrm{s}\) unspecified; signal type: waveform \(V\)/\(I\)
\newline
\textbf{Method:} \ac{dt} for \ac{fc}, \ac{svm} with \ac{pca} for \ac{fl}, PI-based control for wind energy conversion system
\newline
\textbf{Result:} Accuracy of \qty{97.9}\,\% (\ac{dt}) and \qty{94.5}\,\% (\ac{svm}); control scheme maintained stable operation under faults and wind variation
\\ \addlinespace

Yousaf et al.\newline
\cite{yousaf_intelligent_2022}
\newline
(2022)
& \textbf{Focus:} \ac{fl} in \ac{mt-hvdc} systems using local terminal data under varying fault resistance and noise
\newline
\textbf{Data:} EMTDC simulation of 4-terminal MMC–\ac{hvdc} system with offshore wind; \num{714} samples; \(\mathrm{f}_\mathrm{s}\) unspecified; signal type: DC \(V\)/\(I\)
\newline
\textbf{Method:} \ac{dwt}-based features and \ac{ffnn} optimized via Bayesian optimization and Levenberg–Marquardt backpropagation
\newline
\textbf{Result:} \qty{0.5144}\,\% overall error across \qtyrange{10}{485}\,ohm; robust to noise and fault variation
\\ \addlinespace

Zhang et al.\newline
\cite{zhang_fault_2022}
\newline
(2022)
& \textbf{Focus:} \ac{fc} and \ac{fl} in small-current grounding distribution networks
\newline
\textbf{Data:} Simulated \qty{10}\,kV 6-line system using relay protection equipment; \num{3000} samples; \(\mathrm{f}_\mathrm{s} = \qty{1}\,\text{kHz}\); signal type: phasor domain \(V\)/\(I\)
\newline
\textbf{Method:} Hybrid \ac{vmd}-\ac{cnn} with Hilbert–Huang transform-based features from positive-sequence current
\newline
\textbf{Result:} \qty{100}\,\% accuracy in \ac{fc} and \ac{fl}; robust to fault resistance and distance
\\ \addlinespace

Zhang et al.\newline
\cite{zhang_fault_2023}
\newline
(2023)
& \textbf{Focus:} Spatio-temporal \ac{fd} and protection in \ac{adn} using multi-agent \ac{rl}
\newline
\textbf{Data:} EMTDC simulation of IEEE 34-bus \ac{adn} with 10 \ac{pmu}s; sample size and \(\mathrm{f}_\mathrm{s}\) unspecified; signal type: phasor domain \(V\)/\(I\)
\newline
\textbf{Method:} Augmented-state extended kalman filter phase lock loop, supervised multi-residual generation learning, and model-free spatio-temporal multi-agent reinforcement learning
\newline
\textbf{Result:} Accuracy \qty{97.125}\,\%; response time \qty{1.274}\,s; outperformed SOGI, wavelet neural network, \ac{sarsa}, and deep deterministic policy gradient
\\ \addlinespace

\end{longtable}

%% file: tables/tab_checklist.tex
% TABLE PRISMA-SCR CHECKLIST
\begin{longtable}{p{3cm} p{0.8cm} p{10cm} p{1.5cm}}
\caption{PRISMA-ScR checklist \cite{tricco_prisma_2018}} \label{tab:checklist} \\

\toprule
\textbf{Section} & \textbf{Item} & \textbf{PRISMA-ScR checklist item} & \textbf{Section} \\
\midrule
\endfirsthead

\toprule
\textbf{Section} & \textbf{Item} & \textbf{PRISMA-ScR checklist item} & \textbf{Section} \\
\midrule
\endhead

\bottomrule
\endfoot

\textbf{Title} &  &  & \\
\addlinespace
\midrule
Title & 1 & Identify the report as a scoping review. & Title \\
\addlinespace
\midrule
\textbf{Abstract} &  &  & \\
\addlinespace
Structured summary & 2 & Provide a structured summary that includes (as applicable): background, objectives, eligibility criteria, sources of evidence, charting methods, results, and conclusions that relate to the review questions and objectives. & Abstract \\
\addlinespace
\hline
\addlinespace
\textbf{Introduction} &  &  & \\
\addlinespace
\hline
\addlinespace
Rationale & 3 & Describe the rationale for the review in the context of what is already known. Explain why the review questions/objectives lend themselves to a scoping review approach. & 1 \\
\addlinespace
Objectives & 4 & Provide an explicit statement of the questions and objectives being addressed with reference to their key elements (e.g., population or participants, concepts, and context) or other relevant key elements used to conceptualize the review questions and/or objectives. & 1 \\
\addlinespace
\hline
\addlinespace
\textbf{Methods} &  &  & \\
\addlinespace
\hline
\addlinespace
Protocol and registration & 5 & Indicate whether a review protocol exists; state if and where it can be accessed (e.g., a Web address); and if available, provide registration information, including the registration number. & 2 \\
\addlinespace
Eligibility criteria & 6 & Specify characteristics of the sources of evidence used as eligibility criteria (e.g., years considered, language, and publication status), and provide a rationale. & 2 \\
\addlinespace
Information sources & 7 & Describe all information sources in the search (e.g., databases with dates of coverage and contact with authors to identify additional sources), as well as the date the most recent search was executed. & 2 \\
\addlinespace
Search & 8 & Present the full electronic search strategy for at least 1 database, including any limits used, such that it could be repeated. & A.2 \\
\addlinespace
Selection of sources of evidence & 9 & State the process for selecting sources of evidence (i.e., screening and eligibility) included in the scoping review. & 2 \\
\addlinespace
Data charting process & 10 & Describe the methods of charting data from the included sources of evidence (e.g., calibrated forms or forms that have been tested by the team before their use, and whether data charting was done independently or in duplicate) and any processes for obtaining and confirming data from investigators. & 2 \\
\addlinespace
Data items & 11 & List and define all variables for which data were sought and any assumptions and simplifications made. & 2 \\
\addlinespace
Critical appraisal of individual sources of evidence & 12 & If done, provide a rationale for conducting a critical appraisal of included sources of evidence; describe the methods used and how this information was used in any data synthesis (if appropriate). & N/A \\
\addlinespace
Synthesis of results & 13 & Describe the methods of handling and summarizing the data that were charted. & 2 \\
\addlinespace
\hline
\addlinespace
\textbf{Results} &  &  & \\
\addlinespace
\hline
\addlinespace
Selection of sources of evidence & 14 & Give numbers of sources of evidence screened, assessed for eligibility, and included in the review, with reasons for exclusions at each stage, ideally using a flow diagram. & 2 \\
\addlinespace
Characteristics of sources of evidence & 15 & For each source of evidence, present characteristics for which data were charted and provide the citations & 4, 5, 6 \\
\addlinespace
Critical appraisal within sources of evidence & 16 & If done, present data on critical appraisal of included sources of evidence (see item 12). & N/A \\
\addlinespace
Results of individual sources of evidence & 17 & For each included source of evidence, present the relevant data that were charted that relate to the review questions and objectives. & 4, 5, 6 \\
\addlinespace
Synthesis of results & 18 & Summarize and/or present the charting results as they relate to the review questions and objectives & 4, 5, 6 \\
\addlinespace
\hline
\addlinespace
\textbf{Discussion} &  &  & \\
\addlinespace
\hline
\addlinespace
Summary of evidence & 19 & Summarize the main results (including an overview of concepts, themes, and types of evidence available), link to the review questions and objectives, and consider the relevance to key groups. & 7 \\
\addlinespace
Limitations & 20 & Discuss the limitations of the scoping review process & 7 \\
\addlinespace
Conclusions & 21 & Provide a general interpretation of the results with respect to the review questions and objectives, as well as potential implications and/or next steps. & 7 \\
\addlinespace
\hline
\addlinespace
\textbf{Funding} &  &  & \\
\addlinespace
\hline
\addlinespace
Funding & 19 & Describe sources of funding for the included sources of evidence, as well as sources of funding for the scoping review. Describe the role of the funders of the scoping review. & Additional information \\
\addlinespace

\bottomrule
\end{longtable}

%% file: tables/tab_database_query.tex
\begin{longtable}{@{}p{2.5cm}p{1.5cm}p{12cm}@{}}
\caption{Search results from different databases}\label{tab:database_queries} \\
\toprule
\textbf{Database} & \textbf{Results} & \textbf{Search string} \\
\midrule
\endfirsthead

\toprule
\textbf{Database} & \textbf{Results} & \textbf{Search string} \\
\midrule
\endhead

\bottomrule
\endfoot

Scopus & 349 & TITLE-ABS-KEY ( "relay protection" OR "system integrity protection schemes" OR "protection schemes" OR "grid protection" OR "power system protection" OR "protection coordination" ) AND ( "machine learning" OR "deep learning" OR "neural networks" OR "reinforcement learning" OR "artificial intelligence" OR "data-driven techniques" ) AND ( "power grid" OR "electrical grid" OR "power network" OR "smart grid" OR "HVDC" OR "AC-DC power system" OR "high-voltage grid" OR "energy grid" ) \\

IEEE Xplore & 254 & ("All Metadata":"relay protection" OR "All Metadata":"system integrity protection schemes" OR "All Metadata":"protection schemes" OR "All Metadata":"grid protection" OR "All Metadata":"power system protection" OR "All Metadata":"protection coordination") AND ("All Metadata":"machine learning" OR "All Metadata":"deep learning" OR "All Metadata":"neural networks" OR "All Metadata":"reinforcement learning" OR "All Metadata":"artificial intelligence" OR "All Metadata":"data-driven techniques") AND ("All Metadata":"power grid" OR "All Metadata":"electrical grid" OR "All Metadata":"power network" OR "All Metadata":"smart grid" OR "All Metadata":"HVDC" OR "All Metadata":"AC-DC power system" OR "All Metadata":"high-voltage grid" OR "All Metadata":"energy grid") \\

\end{longtable}